\documentclass[10pt,journal,compsoc]{IEEEtran}

\ifCLASSOPTIONcompsoc
  \usepackage[nocompress]{cite}
\else
  \usepackage{cite}
\fi

\ifCLASSINFOpdf
\else
\fi

\usepackage{times}
\usepackage{epsfig}
\usepackage{graphicx}
\usepackage{amsmath}
\usepackage{amssymb}

\usepackage{multirow}
\usepackage{siunitx}
\newcommand{\tabincell}[2]{\begin{tabular}{@{}#1@{}}#2\end{tabular}}

\usepackage{mathtools}

\DeclarePairedDelimiter\floor{\lfloor}{\rfloor}

\usepackage{url}
\usepackage{xspace}
\usepackage{xcolor}

\makeatletter
\DeclareRobustCommand\onedot{\futurelet\@let@token\@onedot}
\def\@onedot{\ifx\@let@token.\else.\null\fi\xspace}

\def\eg{\emph{e.g}\onedot} 
\def\ie{\emph{i.e}\onedot}

\def\etal{\emph{et al}\onedot}
\makeatother

\begin{document}

\title{From Points to Parts: 3D Object Detection from Point Cloud with Part-aware and Part-aggregation Network}

\author{Shaoshuai Shi, 
	    Zhe Wang, Jianping Shi,
	    Xiaogang Wang, 
	    Hongsheng Li
        % <-this % stops a space
\IEEEcompsocitemizethanks{\IEEEcompsocthanksitem S. Shi, X. Wang and H. Li are with the Department
of Electrical Engineering, The Chinese University of Hong Kong, Hong Kong, China.

E-mail: \{ssshi, xgwang, hsli\}@ee.cuhk.edu.hk

\IEEEcompsocthanksitem Z. Wang and J. Shi are with SenseTime Research.

E-mail: \{wangzhe, shijianping\}@sensetime.com
} 

}

\IEEEtitleabstractindextext{%
\begin{abstract}
3D object detection from LiDAR point cloud is a challenging problem in 3D scene understanding and has many practical applications.
\textcolor{black}{In this paper, we extend our preliminary work PointRCNN to a novel and strong point-cloud-based 3D object detection framework, the part-aware and aggregation neural network (Part-$A^2$ net)}. 
The whole framework consists of the part-aware stage and the part-aggregation stage. 
Firstly, the part-aware stage for the first time fully utilizes free-of-charge part supervisions derived from 3D ground-truth boxes to simultaneously predict high quality 3D proposals and accurate intra-object part locations.
The predicted intra-object part locations within the same proposal are grouped by our new-designed RoI-aware point cloud pooling module, which results in an effective representation to encode the geometry-specific features of each 3D proposal. 
Then the part-aggregation stage learns to re-score the box and refine the box location by exploring the spatial relationship of the pooled intra-object part locations.
Extensive experiments are conducted to demonstrate the performance improvements from each component of our proposed framework. 
Our Part-$A^2$ net outperforms all existing 3D detection methods and achieves new state-of-the-art on KITTI 3D object detection dataset by utilizing only the LiDAR point cloud data. Code is available at \url{https://github.com/sshaoshuai/PointCloudDet3D}.
\end{abstract}

}

\maketitle

\IEEEdisplaynontitleabstractindextext

% \fi
\IEEEpeerreviewmaketitle

\IEEEraisesectionheading{\section{Introduction}\label{sec:introduction}}

\IEEEPARstart{W}{ith} the surging demand from autonomous driving and robotics, increasing attention has been paid to 3D object detection 
%from point clouds obtained by the depth sensors like LiDAR sensors 
\cite{Chen2017CVPR, song2014sliding, song2016deep, ku2018joint, Liang2018ECCV, qi2017frustum, yan2018second, 8461232, lang2018pointpillars, Yang2018CoRL, yang2018pixor, luo2018fast, simony2018complex}. 
Though significant achievements have been made in 2D object detection from images \cite{girshick2015fast, ren2015faster, liu2016ssd, redmon2016you, redmon2017yolo9000, lin2017feature, dai2017deformable, lin2018focal, law2018cornernet, he2017mask, cai2018cascade}, directly extending these 2D detection methods to 3D detection 
%from point clouds 
might lead to inferior performance, since the point cloud data of 3D scenes has irregular data format and 
3D detection with point clouds faces great challenges from the irregular data format and large search space of 6 Degrees-of-Freedom (DoF) of 3D objects.

\begin{figure}[t]
%	\vspace{-0.5cm}
	\begin{center}
		\includegraphics[width=1.0\linewidth]{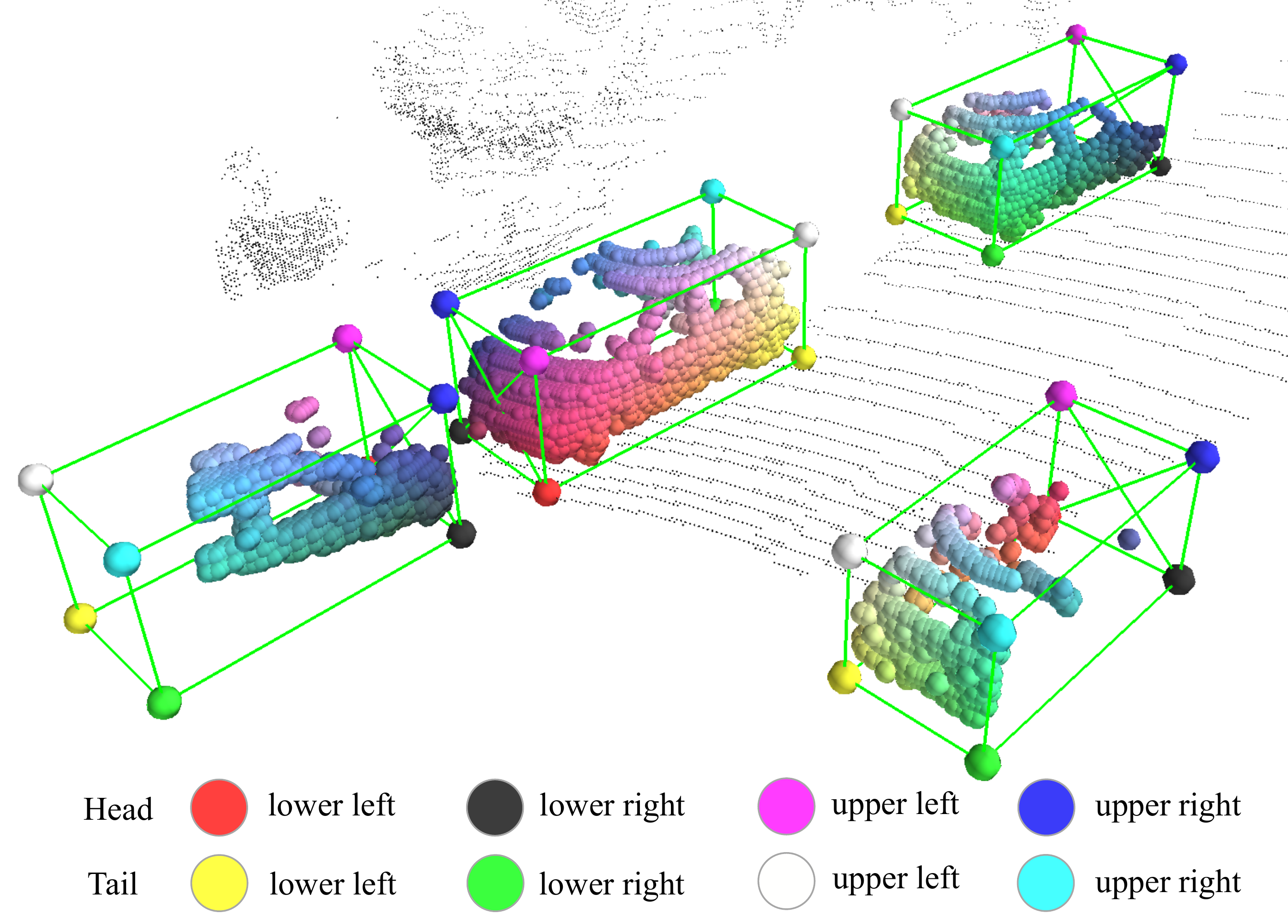}
	\end{center}
	\vspace{-0.1cm}
	\caption{
	Our proposed part-aware and aggregation network can accurately predict intra-object part locations even when objects are partially occluded. Such part locations can assist accurate 3D object detection. The predicted intra-object part locations by our proposed method are visualized by interpolated colors of eight corners. Best viewed in colors.}
	\label{fig:demo}
	\vspace{-0.2cm}
\end{figure}

Existing 3D object detection methods have explored several ways to tackle these challenges. 
Some works \cite{qi2017frustum, xu2018pointfusion, wang2019frustum} utilize 2D detectors to detect 2D boxes from the image, and then adopt PointNet~\cite{qi2017pointnet, qi2017pointnet++} to the cropped point cloud to directly regress the parameters of 3D boxes from raw point cloud. However, these methods heavily depend on the performance of 2D object detectors and cannot take the advantages of 3D information for generating robust bounding box proposals. 
Some other works \cite{Chen2017CVPR, ku2018joint, Yang2018CoRL, yang2018pixor, Liang2018ECCV} project the point cloud from the bird view to create a 2D bird-view point density map and apply 2D Convolutional Neural Networks (CNN) to these feature maps for 3D object detection, but the hand-crafted features cannot fully exploit the 3D information of raw point cloud and may not be optimal. 
There are also some one-stage 3D object detectors \cite{zhou2018voxelnet, yan2018second, lang2018pointpillars} that divide the 3D space into regular 3D voxels and apply 3D CNN or 3D sparse convolution \cite{SubmanifoldSparseConvNet, 3DSemanticSegmentationWithSubmanifoldSparseConvNet} 
to extract 3D features and finally compress to bird-view feature map for 3D  object detection. 
These works do not fully exploit all available information from 3D box annotations for improving the performance of 3D detection. For instance, the 3D box annotations also imply the point distributions within each 3D objects, which are beneficial for learning more discriminative features to improve the performance of 3D object detection.  
Also, these works are all one-stage detection frameworks which cannot utilize the RoI-pooling scheme to pool specific features of each proposal for the box refinement in a second stage.

In contrast, we propose a novel two-stage 3D object detection framework, the part-aware and aggregation neural network (\ie Part-$A^2$ net), which directly operates on 3D point cloud and achieves state-of-the-art 3D detection performance by fully exploring the informative 3D box annotations from the training data. 
Our key observation is that,
unlike object detection from 2D images, 3D objects in autonomous driving scenes are naturally and well separated by annotated 3D bounding boxes, which means the training data with 3D box annotations automatically provides free-of-charge semantic masks and even the relative location of each foreground point within the 3D ground truth bounding boxes (see Fig. \ref{fig:demo} for illustration). In the remaining parts of this paper, the relative location of each foreground point w.r.t. the object box that it belongs to is denoted as \textit{the intra-object part locations}.
This is totally different from the box annotations in 2D images, since some parts of objects in the 2D images may be occluded. Using the ground-truth 2D bounding boxes would generate inaccurate and noisy intra-object part locations for each pixel within objects. 
These 3D intra-object part locations imply the 3D point distributions of 3D objects. 
Such 3D intra-object part locations are informative and can be obtained for free, but \textit{were never explored in 3D object detection}.

Motivated by this observation, our proposed 
Part-$A^2$ net is designed as a novel two-stage 3D detection framework, which consists of the part-aware stage (Stage-I) for predicting accurate intra-object part locations and learning point-wise features, and the part-aggregation stage (Stage-II) for aggregating the part information to improve the quality of predicted boxes. 
\textcolor{black}{
Our approach produces 3D bounding boxes parameterized with $(x, y, z, h, w, l, \theta)$, where $(x, y, z)$ are the box center coordinates, $(h, w, l)$ are the height, width and length of each box respectively, and $\theta$ is the orientation angle of each box from the bird's eye view.}

Specifically, in the part-aware stage-I, the network learns to segment the foreground points and estimate the intra-object part locations for all the foreground points (see Fig.~\ref{fig:demo}), where the segmentation masks and ground-truth part location annotations are directly generated from the ground-truth 3D box annotations.% 
In addition, it also generates 3D proposals from the raw point cloud simultaneously with foreground segmentation and part estimation. 
We investigate two strategies, i.e., anchor-free v.s. anchor-based strategies, for 3D proposal generation to handle to different scenarios. The anchor-free strategy is relatively light-weight and is more memory efficient, while the anchor-based strategy achieves higher recall rates with more memory and calculation costs.
For the anchor-free strategy, we propose to directly generate 3D bounding box proposals in a bottom-up scheme by segmenting foreground points and generating the 3D proposals from the predicted foreground points simultaneously. 
Since it avoids using the large number of 3D anchor boxes in the whole 3D space as previous methods \cite{zhou2018voxelnet, ku2018joint} do, it saves much memory.
For the anchor-based strategy, it generates 3D proposals from downsampled bird-view feature maps with pre-defined 3D anchor boxes at each spatial location.  Since it needs to place multiple 3D anchors with different orientations and classes at each location, it needs more memory but can achieve higher object recall.

In the second stage of existing two-stage detection methods, information within 3D proposals needs to be aggregated by certain pooling operations for the following box re-scoring and location refinement.
However, the previous point cloud pooling strategy (as used in our preliminary PointRCNN \cite{shi2019pointrcnn}) result in ambiguous representations, since different proposals might end up pooling the same group of points, which lose the abilities to encode the geometric information of the proposals.
To tackle this problem, we propose a novel differentiable RoI-aware point cloud pooling operation, which keeps all information from both non-empty and empty voxels within the proposals, to eliminate the ambiguity of previous point cloud pooling strategy. 
This is vital to obtain an effective representation for box scoring and location refinement, as the empty voxels also encode the box's geometry information.

The stage-II aims to aggregate the pooled part features from stage-I by the proposed RoI-aware pooling for improving the quality of the proposals. Our stage-II network adopts the sparse convolution and sparse pooling operations to gradually aggregate the pooled part features of each 3D proposal for accurate confidence prediction and box refinement. 
The experiments show that the aggregated part features could improve the quality of the proposals remarkably and our overall framework achieves state-of-the-art performance on KITTI 3D detection benchmark.

Our primary contributions could be summarized into four-fold. 
(1) We proposed the Part-$A^2$ net framework for 3D object detection from point cloud, which boosts the 3D detection performance by using the free-of-charge intra-object part information 
to learning discriminative 3D features and by effectively aggregating the part features with RoI-aware pooling and sparse convolutions.
(2) We present two strategies for 3D proposal generation to handle different scenarios. The anchor-free strategy is more memory efficient while the anchor-based strategy results in higher object recall.
(3) We propose a differentiable RoI-aware point cloud region pooling operation to eliminate the ambiguity in existing point cloud region pooling operations. The experiments show that the pooled feature representation benefits box refinement stage significantly.
(4) Our proposed Part-$A^2$ net outperforms all published methods with remarkable margins 
and ranks $1^{st}$ with 14 FPS inference speed on the challenging KITTI 3D detection benchmark \cite{Geiger2012CVPR} as of August 15, 2019, which demonstrates the effectiveness of our method.

\begin{figure*}
	\begin{center}
		\includegraphics[width=1.0\linewidth]{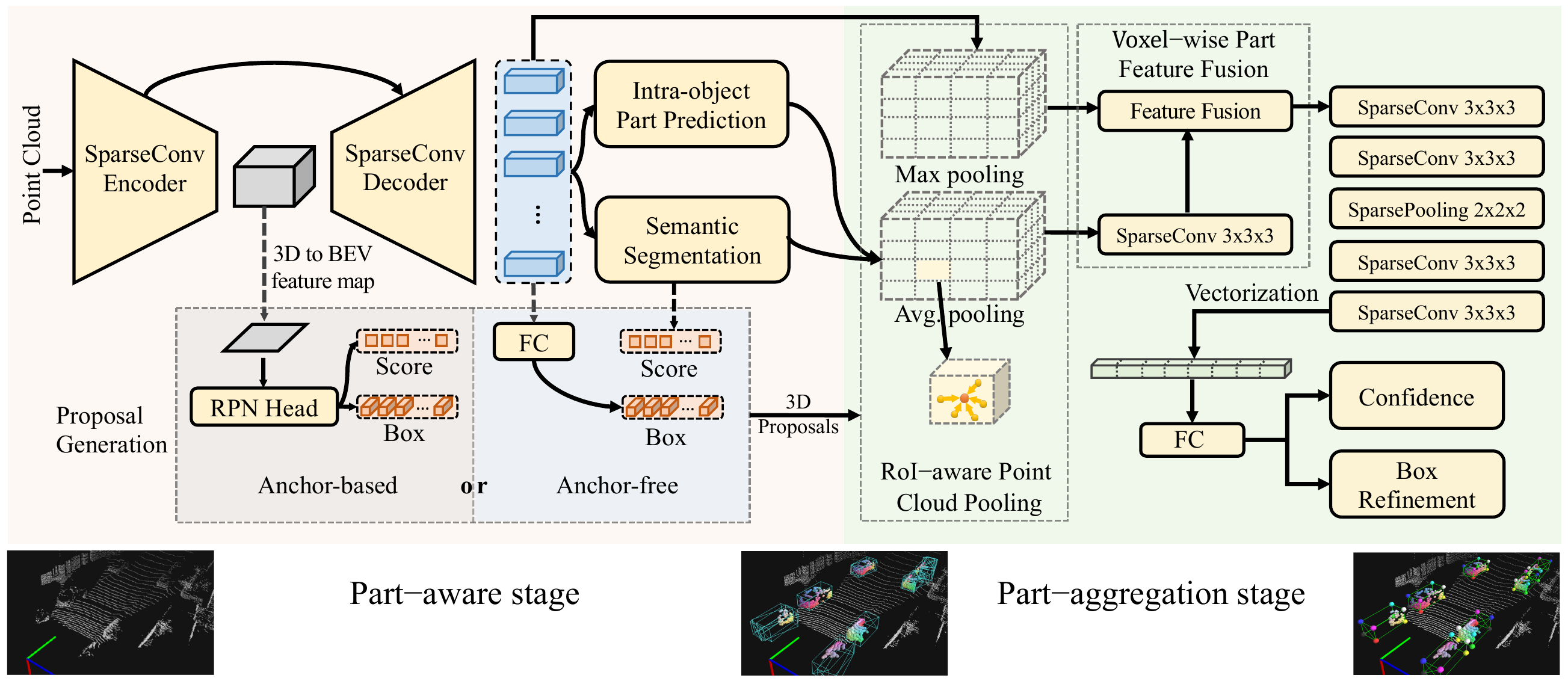}
	\end{center}
	\vspace{-0.4cm}
	\caption{The overall framework of our part-aware and aggregation neural network for 3D object detection. It consists of two stages: (a) the part-aware stage-I for the first time predicts intra-object part locations and generates 3D proposals by feeding the point cloud to our encoder-decoder network. (b) The part-aggregation stage-II conducts the proposed RoI-aware point cloud pooling operation to aggregate the part information from each 3D proposal, then the part-aggregation network is utilized to score boxes and refine locations based on the part features and information from stage-I.}
	\label{fig:framework_total}
%	\vspace{-0.3cm}
\end{figure*}

\section{Related Work}
\noindent
\textbf{3D object detection from 2D images.}~
There are several existing works on estimating the 3D bounding box from images. \cite{mousavian20173d,li2019gs3d} leveraged the geometry constraints between 3D and 2D bounding box to recover the 3D object pose. \cite{chabot2017deep, zhu2014single, mottaghi2015coarse, manhardt2018roi10d} exploited the similarity between 3D objects and the CAD models. Chen \etal \cite{chen2016monocular, chen20153d} formulated the 3D geometric information of objects as an energy function to score the predefined 3D boxes. Ku \etal \cite{ku2019monopsr} proposed the aggregate losses to improve the 3D localization accuracy from monocular image. 
Recently \cite{licvpr2019, wangcvpr2019} explored the stereo pair of images to improve the 3D detection performance from stereo cameras. 
These works can only generate coarse 3D detection results due to the lack of accurate depth information and can be substantially affected by appearance variations.

\noindent
{\bf 3D object detection from multiple sensors.} ~
Several existing methods have worked on fusing the information from multiple sensors (\eg, LiDAR and camera) to help 3D object detection. 
\cite{Chen2017CVPR, ku2018joint} projected the point cloud to the bird view and extracted features from bird-view maps and images separately, which are then cropped and fused by projecting 3D proposals to the corresponding 2D feature maps for 3D object detection. \cite{Liang2018ECCV} further explored the feature fusion strategy by proposing continuous fusion layer to fuse image feature to bird-view features.  Different from projecting point cloud to bird-view map, \cite{qi2017frustum, xu2018pointfusion} utilized off-the-shelf 2D object detectors to detect 2D boxes first for cropping the point cloud and then applied PointNet~\cite{qi2017pointnet, qi2017pointnet++} to extract features from the cropped point clouds for 3D box estimation. These methods may suffers from the time synchronization problem of multiple sensors in the practical applications. 
Unlike these sensor fusion methods, our proposed 3D detection frameworks Part-$A^2$ net could achieve comparable or even better 3D detection results by using only point cloud as input.

\medskip
\noindent
{\bf 3D object detection from point clouds only.} ~
Zhou \etal~\cite{zhou2018voxelnet} for the first time proposed VoxelNet architecture to learn discriminative features from point cloud and detect 3D object with only point cloud. \cite{yan2018second} improved VoxelNet by introducing sparse convolution~\cite{SubmanifoldSparseConvNet, 3DSemanticSegmentationWithSubmanifoldSparseConvNet} for efficient voxel feature extraction. \cite{yang2018pixor, Yang2018CoRL, lang2018pointpillars} projected the point cloud to bird-view maps and applied 2D CNN on these maps for 3D detection. 
These methods do not fully exploit all available information from the informative \textbf{}3D box annotations and are all one-stage 3D detection methods. 
In contrast, our proposed two-stage 3D detection framework Part-$A^2$ net explores the abundant information provided by 3D box annotations and learns to predict accurate intra-object part locations to learn the point distribution of 3D objects, the predicted intra-object part locations are aggregated in the second stage for refining the 3D proposals,
%with the pooled proposal-specific features, 
which significantly improves the performance of 3D object detection. 

\medskip
\noindent
{\bf Point cloud feature learning for 3D object detection.}~
There are generally three ways of learning features from point cloud for 3D detection. (1) \cite{Chen2017CVPR,ku2018joint,yang2018pixor,Yang2018CoRL,Liang2018ECCV} projected point cloud to bird-view map and utilized 2D CNN for feature extraction. 
(2) \cite{qi2017frustum, xu2018pointfusion} conducted PointNet~\cite{qi2017pointnet, qi2017pointnet++} to learn the point cloud features directly from raw point cloud. 
(3) \cite{zhou2018voxelnet} proposed VoxelNet and \cite{yan2018second} applied sparse convolution \cite{SubmanifoldSparseConvNet, 3DSemanticSegmentationWithSubmanifoldSparseConvNet} to speed up the VoxelNet for feature learning. 
Only the second and third methods have the potential to extract point-wise features for segmenting the foreground points and predicting the intra-object part locations in our framework. 
Here we design an encoder-decoder point cloud backbone network similarly with UNet \cite{ronneberger2015u} to extract discriminative point-wise features, which is based on the 3D sparse convolution and 3D sparse deconvolution operations since they are more efficient and effective than the point-based backbone like PointNet++ \cite{qi2017pointnet++}. The point-based backbone and voxel-based backbone are experimented and discussed in Sec.~\ref{ab:backbone}. 

\vspace{1mm}
\noindent
\textcolor{black}{
	\textbf{3D/2D instance segmentation.}  These approaches are often based on point-cloud-based 3D detection methods. 
	Several approaches of 3D instance segmentation are based on the 3D detection bounding boxes with an extra mask branch for predicting the object mask. Yi \etal~\cite{yi2019gspn} proposed an analysis-by-synthesis strategy to generate 3D proposals for 3D instance segmentation. Hou \etal~\cite{hou20193d} combined the multi-view RGB images and 3D point cloud to better generate proposals and predict object instance masks in and end-to-end manner.}

	\textcolor{black}{
	Some other approaches first estimate the semantic segmentation labels and then group the points into instances based on the learned point-wise embeddings. Wang \etal~\cite{wang2018sgpn} calculated the similarities between points for grouping foreground points of each instance. Wang \etal~\cite{wang2019associatively} proposed a semantic-aware point-level instance embedding strategy to learn better features for both the semantic and instance point cloud segmentation. Lahoud \etal~\cite{lahoud20193d} proposed a mask-task learning framework to learn the feature embedding and the directional information of the instance's center for better clustering the points into instances. However, they did not utilize the free-of-charge intra-object part locations as extra supervisions as our proposed method does.}
	
	\textcolor{black}{
	There are also some anchor-free approaches for 2D instance segmentation by clustering the pixels into instances.
	Brabandere \etal~\cite{bert2017} adopted a discriminative loss function to cluster the pixels of the same instance in a feature space while Bai \etal~\cite{bai2017deep} proposed to estimate a modified watershed energy landscape to separate the pixels of different instances. However, those methods only group foreground pixels/points into different instances and did not estimate the 3D bounding boxes. 
	Different with the above methods, our proposed anchor-free approach estimates intra-object part locations and directly generates 3D bounding box proposals from individual 3D points for achieving 3D object detection.
}

\noindent 
\textcolor{black}{
\textbf{Part models for object detection.}
Deformable Part-based Models (DPM) \cite{felzenszwalb2009object} achieved great success on 2D object detection before the deep learning models are utilized. \cite{fidler20123d,pepik20123d,lim2014fpm} extended the DPM to 3D world to reason the parts in 3D and estimate the object poses, where \cite{fidler20123d} modeled the object as a 3D cuboid with both deformable faces and deformable parts, \cite{pepik20123d} proposed a 3D DPM that generates a full 3D object model with continuous appearance representation, and \cite{lim2014fpm} presented the notion of 3D part sharing with 3D CAD models to estimate the fine poses. These DPM based approaches generally adopt several part templates trained with hand-crafted features to localize the objects and estimate the object pose. In contrast, we formulate the object parts as point-wise intra-object part locations in the context of point cloud, where the training labels of part locations could be directly generated from 3D box annotations and they implicitly encode the part distribution of 3D objects. Moreover, both the estimation and aggregation of intra-object part locations are learned by the more robust deep learning networks instead of the previous hand-crafted schemes. 
}

\section{Part-$A^2$ Net: 3D Part-Aware and Aggregation for 3D Detection from Point Cloud}\label{sec:method}
A preliminary version of this work was presented in \cite{shi2019pointrcnn}, where we proposed PointRCNN for 3D object detection from raw point cloud. 
To make the framework more general and effective, in this paper, we extend PointRCNN to a new end-to-end 3D detection framework,  the part-aware and aggregation neural network, \ie Part-$A^2$ net, to further boost the performance of 3D object detection from point cloud. 

The key observation is that, the ground-truth boxes of 3D object detection not only automatically provide accurate segmentation mask because of the fact that 3D objects are naturally separated in 3D scenes, but also imply the relative locations for each foreground 3D point within the ground truth boxes.
This is very different from 2D object detection, where 2D object boxes might only contain portion of an object due to occlusion and thus cannot provide accurate relative location for each 2D pixel. 
These relative locations of foreground points encode valuable information of foreground 3D points and is beneficial for 3D object detection.
This is because foreground objects of the same class (like car class) generally have similar 3D shapes and point distributions. The relative locations of foreground points provide strong cues for box scoring and localization.
We name the relative locations of the 3D foreground points w.r.t. to their corresponding boxes \textit{the intra-object part locations}.

\textcolor{black}{
Those intra-object part locations provide rich information for learning discriminative 3D features from point cloud but were never explored in previous 3D object detection methods.
}
With such rich supervisions, we propose a novel part-aware and aggregation 3D object detector, Part-$A^2$ net, for 3D object detection from point cloud. 
Specifically, we propose to use the free-of-charge 3D intra-object part location labels and segmentation labels as extra supervisions to learn better 3D features in the first stage.  The predicted 3D intra-object part locations and point-wise 3D features within each 3D proposal are then aggregated in the second stage to score the boxes and refine their locations. 
The overall framework is illustrated in Fig.~\ref{fig:framework_total}.

\subsection{Stage-I: Part-aware 3D proposal generation}\label{sec:rpn_v2}
The part-aware network aims to extract discriminative features from the point cloud by learning to estimate the intra-object part locations of foreground points, since these part locations implicitly encode the 3D object's shapes by indicating the relative locations of surface points of 3D objects. Also, 
the part-aware stage learns to estimate the intra-object part locations of foreground points and to generate 3D proposals simultaneously. 
Two strategies for 3D proposal generation from point clouds, anchor-free and anchor-based schemes, are proposed to handle different scenarios.

\begin{figure}
	\begin{center}
		\includegraphics[width=1.0\linewidth, height=3.3cm]{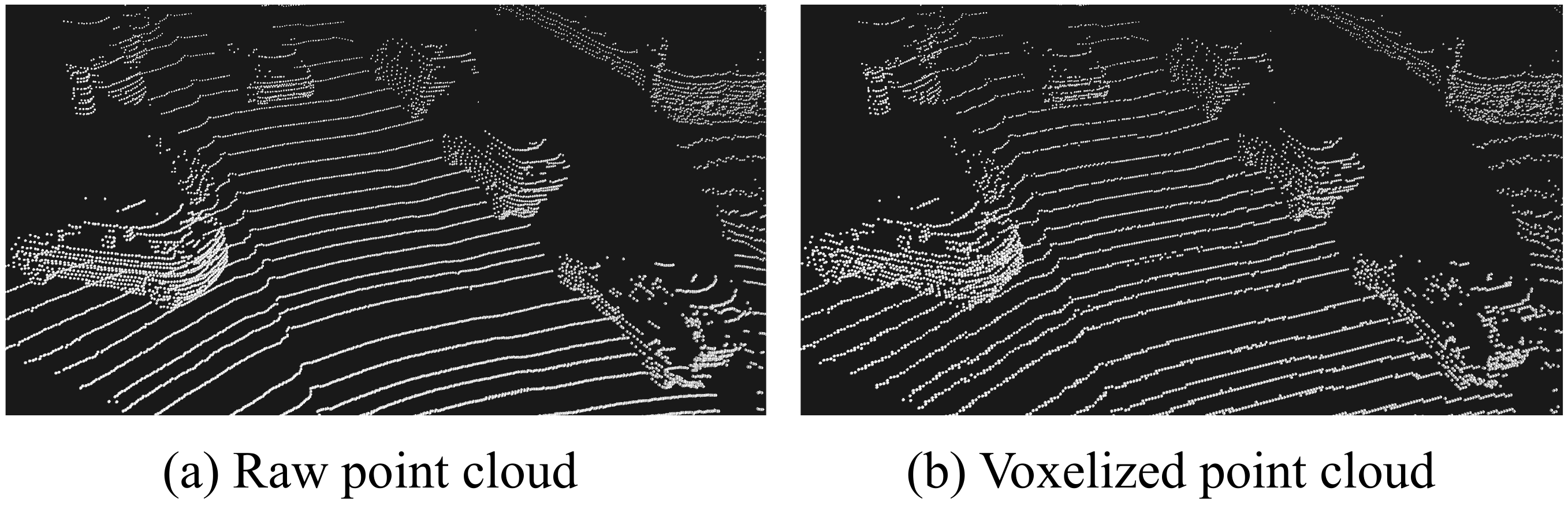}
	\end{center}
	\vspace{-0.3cm}
	\caption{Comparison of voxelized point cloud and raw point cloud in autonomous driving scenarios. The center of each non-empty voxel is considered as a point to form the voxelized point cloud. 
		The voxelized point cloud is approximately equivalent to the raw point cloud and 3D shapes of 3D objects are well kept for 3D object detection.}
	\label{fig:voxel_compare}
\end{figure}

\subsubsection{Point-wise feature learning via sparse convolution}\label{sec:backbone}
For segmenting the foreground points and estimating their intra-object part locations, we first need to learn discriminative point-wise features for describing the raw point clouds. Instead of using point-based methods like \cite{qi2017pointnet, qi2017pointnet++, huang2018recurrent, li2018pointcnn, wang2018deep, zhao2019pointweb,wu2019pointconv} for extracting point-wise features from the point cloud, as show in the left part of Fig.~\ref{fig:framework_total}, we propose to utilize an encoder-decoder network with sparse convolution and deconvolution \cite{SubmanifoldSparseConvNet, 3DSemanticSegmentationWithSubmanifoldSparseConvNet} to learn discriminative point-wise features for foreground point segmentation and intra-object part location estimation, which is more efficient and effective than the previous PointNet++ backbone as used in our preliminary work \cite{shi2019pointrcnn}.

Specifically, we voxelize the 3D space into regular voxels and extract the voxel-wise features of each non-empty voxel by stacking sparse convolutions and sparse deconvolutions, where the initial feature of each voxel is simply calculated as the mean values of point coordinates within each voxel in the LiDAR coordinate system. 
The center of each non-empty voxel is considered as a point to form a new point cloud with the point-wise features (\ie the voxel-wise features), 
which is approximately equivalent to the raw point cloud as shown in Fig.~\ref{fig:voxel_compare}, since the voxel size is much smaller (e.g., 5cm$\times$5cm$\times$10cm in our method) compared to the  whole 3D space ($\sim$70m$\times$80m$\times$4m). For each 3D scene in the KITTI dataset \cite{Geiger2012CVPR}, there are generally about 16,000 non-empty voxels in the 3D space. The voxelized point cloud could not only be processed by the more efficient sparse convolution based backbone, but it also keeps approximately equivalent to the raw point cloud for 3D object detection.

Our sparse convolution based backbone is designed based on the encoder-decoder architecture. The spatial resolution of input feature volumes is 8 times downsampled by a series of sparse convolution layers with stride 2, and is then gradually upsampled to the original resolution by the sparse deconvolutions for the voxel-wise feature learning. The detailed network structure is illustrated in Sec.~\ref{sec:imp_v2} and Fig.~\ref{fig:part_stage1}.
Our newly designed 3D sparse convolution based-backbone results in better 3D box recall than the PointNet++ based backbone in our preliminary PointRCNN framework \cite{shi2019pointrcnn} (as shown by experimental results in Table~\ref{tab:exp_recall_v2}), which demonstrates the effectiveness of this new backbone for the point-wise feature learning. 

\begin{figure}[t]
	\begin{center}
		\includegraphics[width=1.0\linewidth]{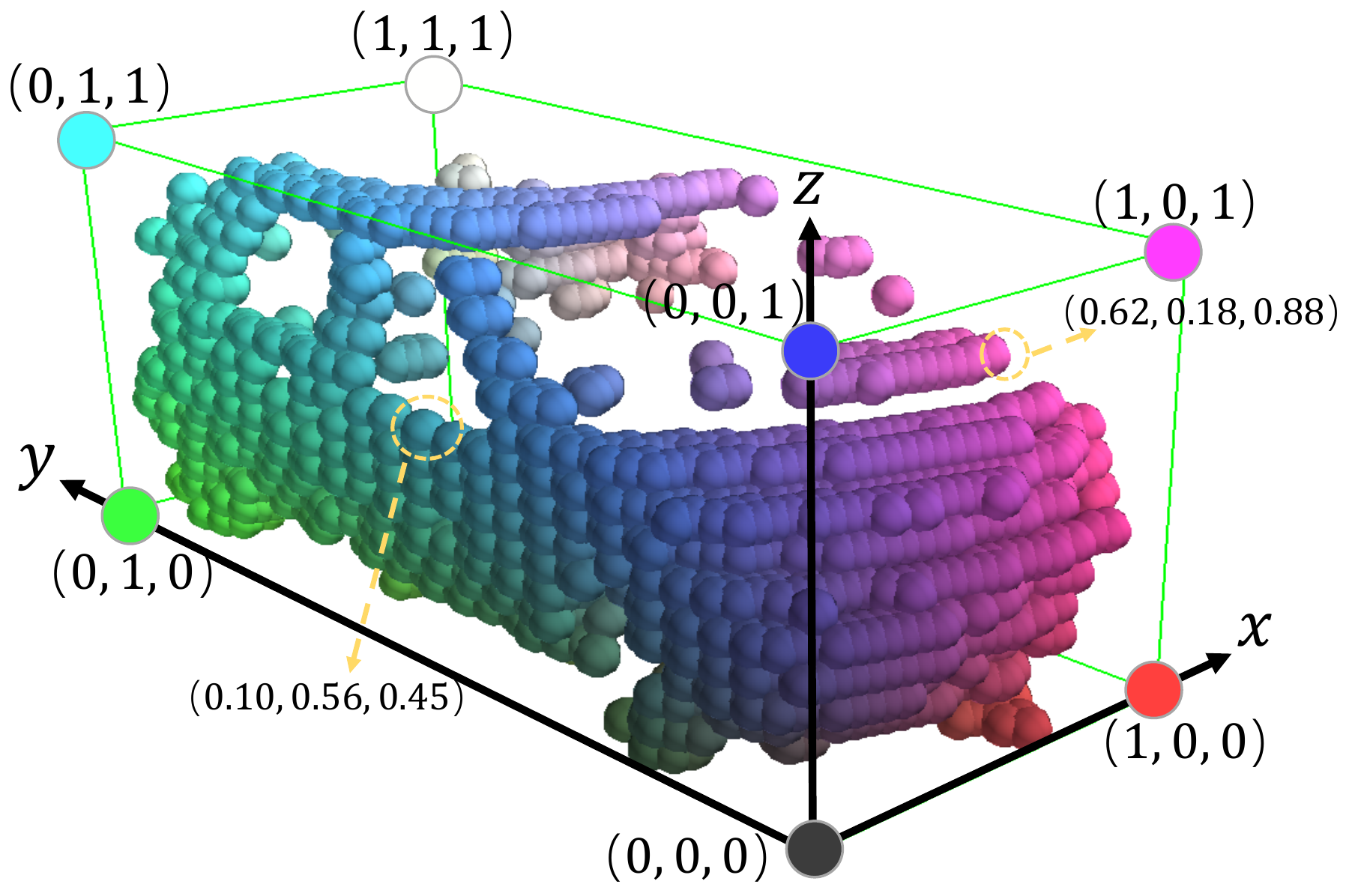}
	\end{center}
	\caption{Illustration of intra-object part locations for foreground points. Here we use interpolated colors to indicate the intra-object part location of each point. Best viewed in colors. }
	\label{fig:part_coords}
	\vspace{-0.5cm}
\end{figure}

\subsubsection{Estimation of foreground points and intra-object part locations}\label{sec:part_formul}
The segmentation masks help the network to distinguish foreground points and background, while the intra-object part locations provide rich information for the neural network to recognize and detect 3D objects. For instance, the side of a vehicle is usually a plane parallel to a side of its corresponding bounding box. By learning to estimate not only the foreground segmentation mask but also the intra-object part location of each point, the neural network develops the ability of inferring the shape and pose of the objects, which is crucial for 3D object detection.

\medskip
\noindent
{\bf Formulation of intra-object part location.}~
As shown in Fig.~\ref{fig:part_coords}, we formulate intra-object part location of each foreground point as its relative location in the 3D ground-truth bounding box that it belongs to.
We denote three continuous values $(x^{(part)}, y^{(part)}, z^{(part)})$ as the target intra-object part location of the foreground point $(x^{(p)}, y^{(p)}, z^{(p)})$, which can be calculated as follows
\begin{align}\label{part_label}
\begin{bmatrix}
x^{(t)}& y^{(t)}
\end{bmatrix}
&=
\begin{bmatrix}
x^{(p)} - x^{(c)} & y^{(p)} - y^{(c)}
\end{bmatrix}
\begin{bmatrix}
\cos(\theta) & \sin(\theta) \nonumber \\
-\sin(\theta) & \cos(\theta)
\end{bmatrix},\\
x^{(part)} &= \frac{x^{(t)}}{w}+0.5,~~y^{(part)} = \frac{y^{(t)}}{l}+0.5,\nonumber \\ z^{(part)} &= \frac{z^{(p)} - z^{(c)}}{h}+0.5,
\end{align}
\noindent
where $(x^{(c)}, y^{(c)}, z^{(c)})$ is the box center, $(h, w, l)$ is the box size (height, width, length), and $\theta$ is the box orientation in bird-view. 
The relative part location of a foreground point $x^{(part)}, y^{(part)}, z^{(part)} \in [0, 1]$ , and the part location of the object center is therefore $(0.5, 0.5, 0.5)$. 
Note that intra-object location coordinate system follows the similar definition of KITTI's global coordinate system, where the direction of $z$ is perpendicular to the ground, and $x$ and $y$ are parallel to the horizontal plane.

\medskip
\noindent
{\bf Learning foreground segmentation and intra-object part location estimation.}~
As shown in Fig.~\ref{fig:framework_total}, given the above sparse convolution based backbone, two branches 
are appended to the output features of the encoder-decoder backbone for segmenting the foreground points and predicting their intra-object part locations. Both branches utilize the sigmoid function as last non-linearity function for generating outputs.
The segmentation scores of foreground points indicate the confidence of the predicted intra-object part locations since the intra-object part locations are defined and learned on foreground points only in the training stage. 
Since the number of foreground points is generally much smaller than that of the background points in large-scale outdoor scenes, we adopt the focal loss \cite{lin2018focal} for calculating point segmentation loss $\mathcal{L}_{\text{seg}}$ to handle the class imbalance issue
\begin{align}\label{eq:fl}
\mathcal{L}_{\textrm{seg}}(p_t)  &= -\alpha_t(1 - p_t)^\gamma \log(p_t),\\
\text{where } p_t &= 
\begin{cases}
p  &\text{for forground points},\\ 
1 - p & \text{otherwise}.
\end{cases} \nonumber
\end{align}
where $p$ is the predicted foreground probability of a single 3D point and we keep $\alpha_t=0.25$ and $\gamma=2$ as the original paper. All points inside the ground-truth boxes are utilized as positive points and others are considered as negative points for training. 

For estimating the intra-object part location (denoted as $(x^{(part)}, y^{(part)}, z^{(part)})$) of each foreground point, 
since they are bounded between $[0,1]$, we apply the binary cross entropy losses to each foreground point as follows
\begin{align}\label{loss:bce}
\mathcal{L}_{\textrm{part}}(u^{(part)}) = &-u^{(part)}\log(\tilde{u}^{(part)}) \nonumber \\
&- (1 - u^{(part)})\log(1 - \tilde{u}^{(part)})\nonumber\\
&\text{for } u \in \{x, y, z\},
\end{align}
where $\tilde{u}^{(part)}$ is the predicted intra-object part location from the network output, and $u^{(part)}$ is the corresponding ground-truth intra-object part location. Note that the part location estimation is only conducted for foreground points.

\medskip
\noindent
\subsubsection{3D proposal generation from point cloud}\label{sec:proposal_gen}
To aggregate the predicted intra-object part locations and the learned point-wise 3D features for improving the performance of 3D object detection in the second stage, we need to generate 3D proposals to group the foreground points that belong to the same object. 
Here we investigate two strategies for 3D proposal generation from point cloud, the anchor-free scheme and anchor-based scheme, to handle different scenarios. The anchor-free strategy is more memory efficient while the anchor-based strategy achieves higher recall with more memory cost.

\medskip
\noindent
{\bf Anchor-free 3D proposal generation.}~
Our model with this strategy is denoted as Part-$A^2$-free. We propose a novel scheme similarly to our preliminary PointRCNN \cite{shi2019pointrcnn} to generate 3D proposals in a bottom-up manner. 
As shown in the left part of Fig.~\ref{fig:framework_total}, we append an extra branch to the decoder of our sparse convolution backbone to generate 3D proposal from each point that is predicted as foreground.

However, if the algorithm directly estimates object's center locations from each foreground point, the regression targets would vary in a large range. For instance, for a foreground point at the corner of an object, its relative offsets to the object center is much larger than those of a foreground point on the side of an object. If directly predicting relative offsets w.r.t. each foreground point with conventional regression losses (e.g., $L1$ or $L2$ losses), the loss would be dominated by errors of corner foreground points. 

\begin{figure}[t]
	\begin{center}
		\includegraphics[width=0.99\linewidth,height=4.5cm]{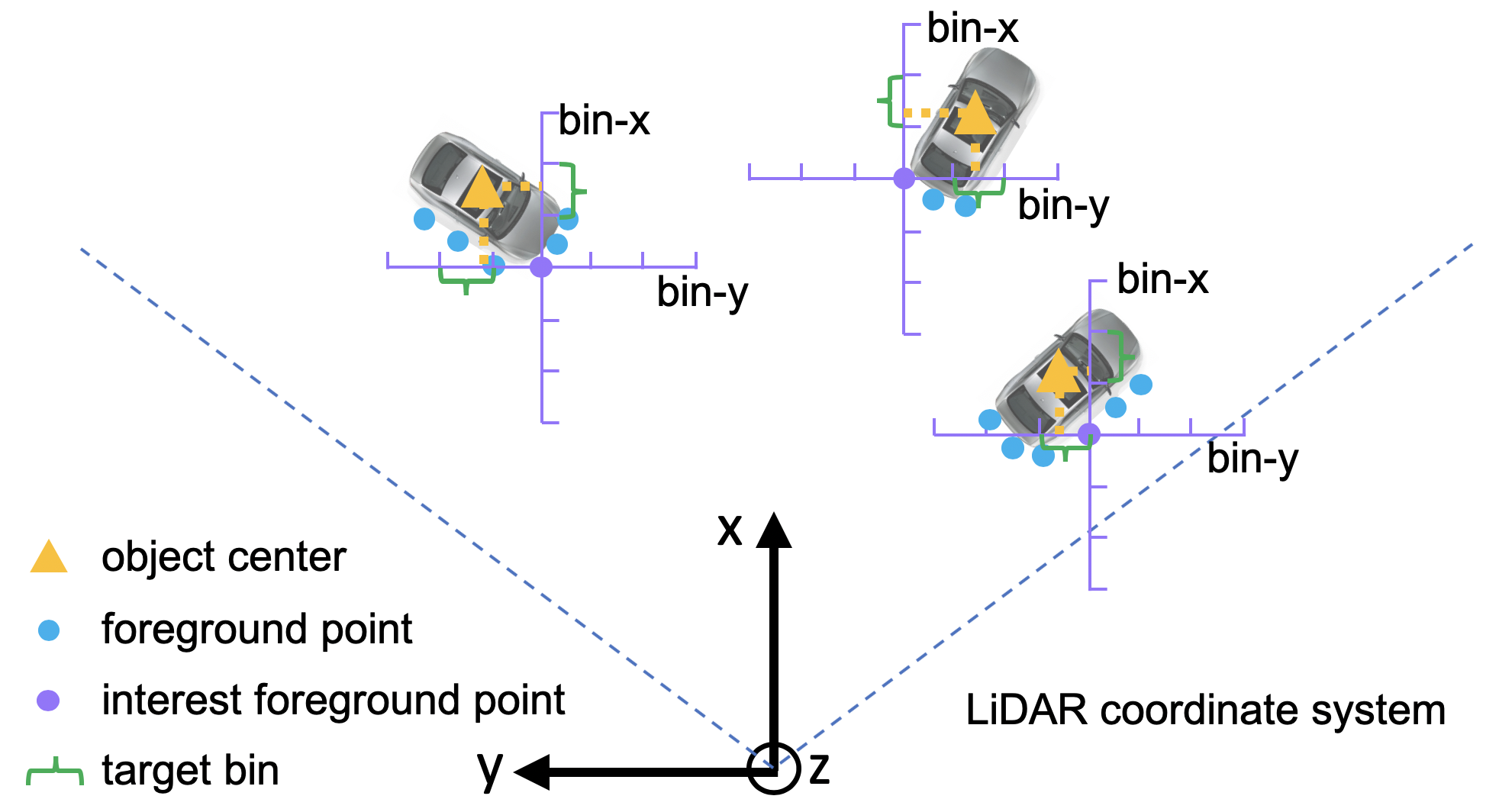}
	\end{center}
	\caption{Illustration of bin-based center localization. The surrounding area along $X$ and $Y$ axes of each foreground point is split into a series of bins to locate the object center.}
	\label{fig:loc}
%	\vspace{-0.3cm}
\end{figure}

To solve the issue of large varying ranges of the regression targets, we propose the bin-based center regression loss.
As shown in Fig.~\ref{fig:loc}, we split the surrounding bird-view area of each foreground point into a series of discrete bins along the $X$ and $Y$ axes by dividing a search range $\mathcal{S}$ of each axis into bins of uniform length $\delta$, which represents different object centers $(x, y)$ on the $X$-$Y$ plane. We observe that conducting bin-based classification with cross-entropy loss for the $X$ and $Y$ axes instead of direct regression with smooth-$L1$ loss \cite{girshick2015fast} results in more accurate and robust center localization. To refine small localization after the assignment into each $X$-$Y$ bin, a small residual is also estimated.
The overall regression loss for the $X$ or $Y$ axes therefore consists of bin classification loss and the residual regression loss within the classified bin.
For the center location $z$ along the vertical $Z$ axis,
we directly utilize smooth-$L1$ loss for the regression since most objects' $z$ values are generally within a very small range. 
The object center regression targets could therefore be formulated as
\begin{align}\label{eqn:bin-based}
%\mathop{\text{bin}_u^{(p)}}\limits_{u \in \{x, z\}}  &= \floor*{\frac{u^p - u^{(p)} + \mathcal{S}}{\delta}},
\text{bin}_x^{(p)}  &= \floor*{\frac{x^{(c)} - x^{(p)} + \mathcal{S}}{\delta}},~
\text{bin}_y^{(p)}  = \floor*{\frac{y^{(c)} - y^{(p)} + \mathcal{S}}{\delta}},\nonumber
\\
\mathop{\text{res}_u^{(p)}}\limits_{u \in \{x, y\}}  &= \frac{1}{\delta}\left( u^{(c)} - u^{(p)} + \mathcal{S} - \left(\text{bin}_u^{(p)} \cdot \delta + \frac{\delta}{2}\right)\right),
\\
\text{res}_z^{(p)} &= z^{(c)} - z^{(p)}, \nonumber
\end{align}
\noindent
where $\delta$ is the bin size, $\mathcal{S}$ is the search range, 
$(x^{(p)}, y^{(p)}, z^{(p)})$ is the coordinates of a foreground point of interest,  $(x^{(c)}, y^{(c)}, z^{(c)})$ is the center coordinates of its corresponding object,
$\text{bin}^{(p)}_x$ and $\text{bin}^{(p)}_y$ are ground-truth bin assignments and $\text{res}^{(p)}_x$ and $\text{res}^{(p)}_y$ are the ground-truth residual for further location refinement within the assigned bin.

Since our proposed bottom-up proposal generation strategy is anchor-free, it does not have an initial value for box orientation. 
Hence 
we directly divide the orientation $2\pi$ into discrete bins with bin size $w$, and calculate the bin classification target $\text{bin}_\theta^{(p)}$ and residual regression target $\text{res}_\theta^{(p)}$ as 
\begin{align}\label{ort_loss}
\text{bin}_{\theta}^{(p)}  &= \floor*{\frac{(\theta + \frac{w}{2})~\text{mod}~ 2\pi}{\omega}},
\\
\text{res}_{\theta}^{(p)}  &= \frac{2}{\omega}\left( ((\theta + \frac{w}{2})~\text{mod}~ 2\pi) - \left(\text{bin}^{(p)}_{\theta} \cdot \omega + \frac{\omega}{2}\right)\right), \nonumber
\end{align}

Thus the overall 3D bounding box regression loss $\mathcal{L}_{\text{box}}$ could be formulated as 
\begin{align}
\mathcal{L}_\text{bin}^{(p)} &= \sum_{u \in \{x, y, \theta\}} ( \mathcal{L}_\text{ce}(\widehat{\text{bin}}_u^{(p)}, \text{bin}_u^{(p)})
+ \mathcal{L}_{\text{smooth-L1}}(\widehat{\text{res}}_u^{(p)}, \text{res}_u^{(p)}) ), \nonumber \\
\mathcal{L}_\text{res}^{(p)} &= \sum_{v \in \{z, h, w, l\}} \mathcal{L}_{\text{smooth-L1}}(\widehat{\text{res}}_v^{(p)}, \text{res}^{(p)}_v),  \label{reg_loss}\\
\mathcal{L}_\text{box} &= \mathcal{L}_\text{bin}^{(p)} + \mathcal{L}_\text{res}^{(p)}, \nonumber
\end{align}
\noindent
where $\widehat{\text{bin}}^{(p)}_u$ and $\widehat{\text{bin}}^{(p)}_u$ are the predicted bin assignments and residuals of the foreground point $p$, $\text{bin}^{(p)}_u$ and $\text{res}^{(p)}_u$ are the ground-truth targets calculated as above,
\textcolor{black}{
$\widehat{\text{res}}^{(p)}_v$ is the predicted center residual in vertical axis or the size residual with respect to the average object size of each class in the entire training set, $\text{res}^{(p)}_v$ is the corresponding ground-truth target of $\widehat{\text{res}}^{(p)}_v$,}
$\mathcal{L}_\text{ce}$ denotes the cross-entropy classification loss, and $\mathcal{L}_{\text{smooth-L1}}$ denotes the smooth-$L1$ loss. Note that the box regression loss $\mathcal{L}_{\text{box}}$ is only applied to the foreground points.

In the inference stage, regressed $x$, $y$ and $\theta$ are obtained by first choosing the bin center with the highest predicted confidence and then adding the predicted residuals. 

Based on this anchor-free strategy, our method not only fully explores the 3D information from point cloud for 3D proposal generation, but also avoids using a large set of predefined 3D anchor boxes in the 3D space by constraining the 3D proposals to be only generated by foreground points. %

\medskip
\noindent
{\bf Anchor-based 3D proposal generation.}~
%For the anchor-based strategy (denoted as Part-$A^2$-anchor), 
Our model with this strategy is denoted as Part-$A^2$-anchor.
The stage-I is illustrated in Fig.~\ref{fig:framework_total}, 
the sparse convolution based encoder takes a voxelized point cloud with shape $M\times N\times H$, and produces an 8-time $X$- and $Y$-axially downsampled $\frac{M}{8}\times \frac{N}{8}$ 2D bird-view feature map with $\frac{H}{16} \times D$ feature channels, where $\frac{H}{16}$ denotes the feature volume being 16-time downsampled along the $Z$ axis, $D$ is the feature dimension of each encoded feature voxel, and ``$\times D$'' denotes concatenating the features at each $x-y$ bird-view location of different heights to obtain a 1D feature vector.
We then append a Region Proposal Network (RPN) head similar with \cite{yan2018second} to the above bird-view feature map for 3D proposal generation with predefined 3D anchors. Each class has $2\times \frac{M}{8}\times \frac{N}{8}$ predefined anchors with the specified anchor size for each class, where each pixel on the bird-view feature map has one anchor parallel to the $X$ axis and one anchor parallel to the $Y$ axis.
Each class will have its own predefined anchors since the object sizes of different classes vary significantly. For instance, we use $(l=3.9, w=1.6, h=1.56)$ meters for cars, $(l=0.8, w=0.6, h=1.7)$ meters for pedestrians and $(l=1.7, w=0.6, h=1.7)$ meters for cyclists on the KITTI dataset. 

The anchors are associated with the ground-truth boxes by calculating the 2D bird-view Intersection-over-Union (IoU), where the positive IoU thresholds are empirically set as 0.6, 0.5, 0.5, and the negative IoU thresholds are 0.45, 0.35, 0.35 for cars, pedestrians and cyclists, respectively. 
We append two convolution layers with kernel size $1\times1\times1$ to the bird-view feature map for proposal classification and box regression. We use focal loss similarly to Eq.~\eqref{eq:fl} for anchor scoring, and directly use the residual-based regression loss for the positive anchors. Here we directly adopt the commonly used smooth-$L1$ loss for regression since the center distances between anchors and their corresponding ground-truth boxes are generally within a smaller range than those of the anchor-free strategy due to the IoU thresholds. 
With the candidate anchor $(x^{(a)}, y^{(a)}, z^{(a)}, h^{(a)}, w^{(a)}, l^{(a)}, \theta^{(a)})$ and the target ground truth $(x^{(gt)}, y^{(gt)}, z^{(gt)}, h^{(gt)}, w^{(gt)}, l^{(gt)}, \theta^{(gt)})$, the residual-based box regression targets for center, angle and size are defined as 
\begin{align}\label{eqn:reg_targets}
&\scriptstyle \Delta x^{(a)} = \frac{x^{(gt)} - x^{(a)}}{d^{(a)}} , ~\Delta y^{(a)} = \frac{y^{(gt)} - y^{(a)}}{d^{(a)}} , 
~\Delta z^{(a)} = \frac{z^{(gt)} - z^{(a)}}{h^{(a)}} , \nonumber \\
&\scriptstyle \Delta l^{(a)} = \log\left(\frac{l^{(gt)}}{l^{(a)}}\right), ~\Delta h^{(a)} = \log\left(\frac{h^{(gt)}}{h^{(a)}}\right), ~\Delta w^{(a)} = \log\left(\frac{w^{(gt)}}{w^{(a)}}\right), \\
&\scriptstyle \Delta \theta^{(a)} = \sin\left(\theta^{(gt)} - \theta^{(a)}\right), ~\text{where} ~d^{(a)}=\sqrt{\left(l^{(a)})^2+(w^{(a)}\right)^2}, \nonumber
\end{align}
where the orientation target is encoded as $\sin(\theta^{(gt)} - \theta^{(a)})$ 
to eliminate the ambiguity of cyclic values of orientation. 
However, this method encodes two opposite directions to the same value, so we adopt an extra convolution layer with kernel size $1\times1\times1$ to the bird-view feature map as in \cite{yan2018second} for classifying two opposite directions of orientation, where the direction target is calculated by the following approach: if $\theta^{(gt)}$ is positive, the direction target is one, otherwise the direction target is zero (Note that $\theta^{(gt)} \in [-\pi, \pi)$). We use cross entropy loss similarly to Eq.~\eqref{loss:bce} for the binary classification of orientation direction, which is denoted as term $\mathcal{L}_{\text{dir}}$. 
Then the overall 3D bounding box regression loss $\mathcal{L}_{\text{box}}$ could be formulated as 
\begin{align}
\mathcal{L}_{\textrm{box}} = \sum_{\text{res} \in \{x, y, z, l, h, w, \theta\}} \mathcal{L}_{\text{smooth-L1}}(\widehat{\Delta \text{res}^{(a)}}, \Delta \text{res}^{(a)}) + \beta\mathcal{L}_{\text{dir}},
\end{align}
where $\widehat{\Delta \text{res}^{(a)}}$ is the predicted residual for the candidate anchor, $\Delta \text{res}^{(a)}$ is the corresponding ground-truth target calculated as Eq.~\eqref{eqn:reg_targets}, and the loss weight $\beta=0.1$. Note that the box regression loss $\mathcal{L}_{\text{box}}$ is only applied to the positive anchors.

\medskip
\noindent
{\bf Discussion of the two 3D proposal generation strategies.}~
Both of these two 3D proposal generation strategies have their advantages and limitations. 
The proposed anchor-free strategy is generally light-weight and memory efficient because it does not requires evaluating a large number of anchors at each spatial location in the 3D space. The efficiency is more obvious for multi-class object detection since different classes in 3D object detection generally require different anchor boxes, while the anchor-free scheme can share the point-wise feature for generating proposals for multiple classes. The second anchor-based proposal generation strategy achieves slightly higher recall by covering the whole bird-view feature map with its predefined anchors for each class, but 
has more parameters and 
requires more GPU memory. The detailed experiments and comparison are discussed in Sec.~\ref{sec:ab_bottom_up}.

\begin{figure}[t]
	\begin{center}
		\includegraphics[width=1.0\linewidth]{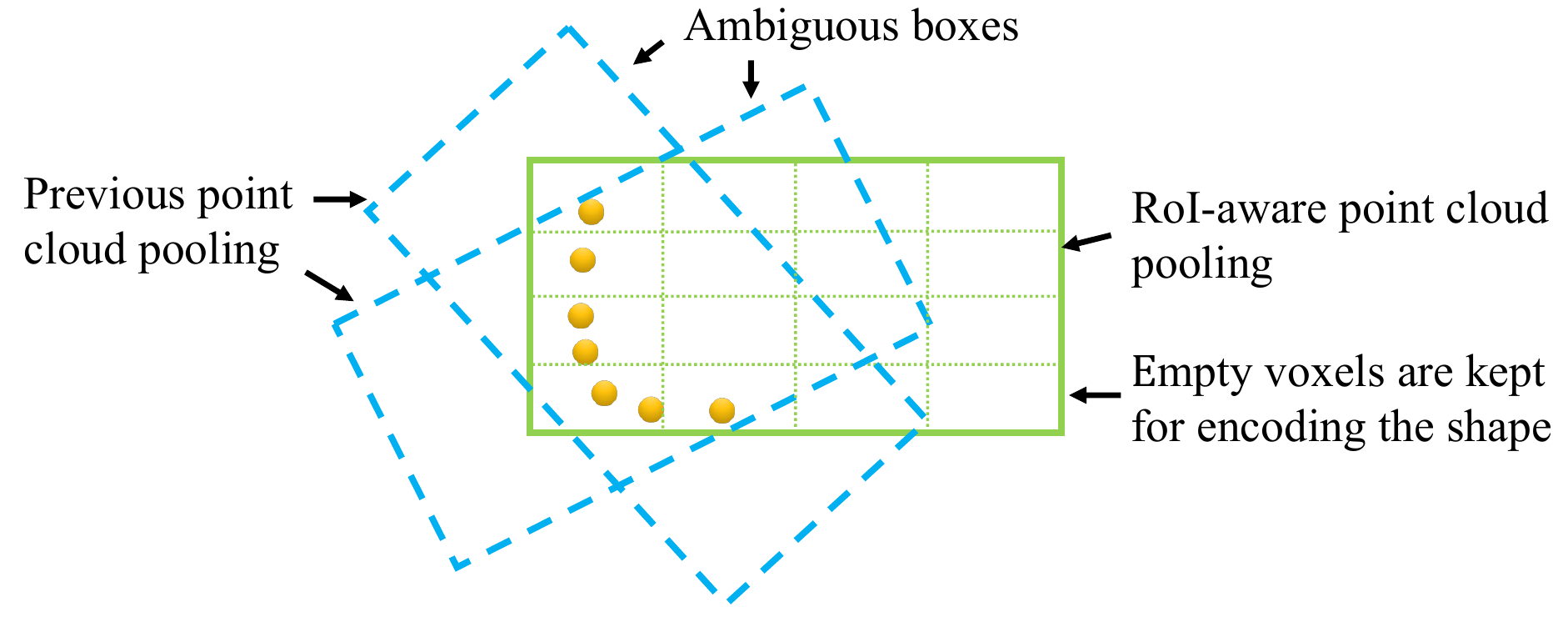}
	\end{center}
	\caption{Illustration of the proposed RoI-aware point cloud feature pooling. The previous point cloud pooling approach could not effectively encode the proposal's geometric information (blue dashed boxes). Our proposed RoI-aware point cloud pooling method could encode the box's geometric information (the green box) by keeping the empty voxels, which could be efficiently processed by following sparse convolution.}
	\label{fig:roipool}
\end{figure}

\subsection{RoI-aware point cloud feature pooling}\label{sec:roi}
Given the predicted intra-object part locations and the 3D proposals, we aim to conduct box scoring and proposal refinement by aggregating the part information and learned point-wise features of all the points within the same proposal. 
In this subsection, we first introduce the canonical transformation to reduce the effects from the rotation and location variations of different 3D proposals, then we propose the RoI-aware point cloud feature pooling module to eliminate the ambiguity of previous point cloud pooling operation and to encode the position-specific features of 3D proposals for box refinement. 

\smallskip
\noindent
{\bf Canonical transformation.}~
We observe that if the box refinement targets are normalized in a canonical coordinate system, it can be better estimated by the following box refinement stage. We transform the pooled points belonging to each proposal to individual canonical coordinate systems of the corresponding 3D proposals. The canonical coordinate system for one 3D proposal denotes that (1) the origin is located at the center of the box proposal; (2) the local $X'$ and $Y'$ axes are approximately parallel to the ground plane with $X'$ pointing towards the head direction of proposal and the other $Y'$ axis perpendicular to $X'$; (3) the $Z'$ axis remains the same as that of the global coordinate system. All pooled points' coordinates $p$ of the box proposal should be transformed to the canonical coordinate system as $\tilde{p}$ by proper rotation and translation. The positive 3D proposals and their corresponding ground-truth 3D boxes are also transformed to the canonical coordinate system to calculate the residual regression targets for box refinement. 
The proposed canonical coordinate system substantially eliminates much rotation and location variations of different 3D proposals and improves the efficiency of feature learning for later box location refinement.

\smallskip
\noindent
{\bf RoI-aware point cloud feature pooling.}~
The point cloud pooling operation in our preliminary work PointRCNN \cite{shi2019pointrcnn} simply pools the point-wise features from the 3D proposals whose corresponding point locations are inside the 3D proposal. All the inside points' features are aggregated by the PointNet++ encoder for refining the proposal in the second stage. 
However, we observe that this operation loses much 3D geometric information and introduces ambiguity between different 3D proposals. This phenomenon is
illustrated in Fig.~\ref{fig:roipool}, where different proposals result in the same pooled points. The same pooled features introduces adverse effects to the following refinement stage.

Therefore, we propose the RoI-aware point cloud pooling module to evenly divide each 3D proposal into regular voxels with a fixed spatial shape $(L_x \times L_y \times L_z)$, where $L_x, L_y, L_z$ are the integer hyperparameters of the pooling resolution in each dimension of the 3D proposals (\eg, $14\times 14\times 14$ is adopted in our framework) and independent of different 3D proposal sizes. 

Specifically, let $\mathcal{F}=\{f_i \in \mathbb{R}^{C_0}, i \in 1, \cdots, n\}$ denote the point-wise features of all the inside points in a 3D proposal $\mathbf{b}$, and they are scattered in the divided voxels of the 3D proposal according to their local canonical coordinates  $\mathcal{X}=\{(x_i^{(ct)}, y_i^{(ct)}, z_i^{(ct)}) \in \mathbb{R}^3, i \in 1, \cdots, n\}$, where $n$ is the number of inside points. 
Then the RoI-aware voxel-wise max pooling and average pooling operations could be denoted as 
\begin{align}\label{eq_roiaware_pool}
Q&=\text{RoIAwareMaxPool}(\mathcal{X},~\mathcal{F},~\mathbf{b}),~~Q \in \mathbb{R}^{L_x \times L_y \times L_z\times C_0},\\
Q&=\text{RoIAwareAvgPool}(\mathcal{X},~\mathcal{F},~\mathbf{b}),~~Q \in \mathbb{R}^{L_x \times L_y \times L_z\times C_0},
\end{align}
where $Q$ is the pooled 3D feature volumes of proposal $\mathbf{b}$. 
Specifically, the feature vector at the $k^{th}$ voxel $Q_k$ of the voxel-wise max pooling and average pooling could be computed as
\begin{align}\label{eq_roiaware}
Q_k &=
\begin{cases}
\max~\{f_i \in \mathcal{N}_k\} ~&\text{if}~ |\mathcal{N}_k| > 0, \\
0 &\text{if}~|\mathcal{N}_k|=0,\\
\end{cases} \nonumber \\\
&~\text{or} \\
Q_k &=
\begin{cases}
\frac{1}{|\mathcal{N}_k|}\sum_i f_i,~f_i \in \mathcal{N}_k ~&\text{if}~ |\mathcal{N}_k| > 0, \\
0 &\text{if}~|\mathcal{N}_k|=0,\\
\end{cases} \nonumber
\end{align}
where $\mathcal{N}_k$ is the set of points belonging to the $k^{th}$ voxel and $k \in \{1, \cdots, L_x\times L_y \times L_z\}$.
Note that here the features of empty voxels ($|\mathcal{N}_k|=0$) would be set to zeros and marked as empty for the following sparse convolution based feature aggregation.

The proposed RoI-aware point cloud pooling module encodes different 3D proposals with the same local spatial coordinates, where each voxel encodes the features of a corresponding fixed grid in the 3D proposal box. This position-specific feature pooling better captures the geometry of the 3D proposal and results in an effective representation for the follow-up box scoring and location refinement. 
Moreover, the proposed RoI-aware pooling module is 
differentiable, which enables the whole framework to be end-to-end trainable.

\subsection{Stage-II: Part location aggregation for confidence prediction and 3D box refinement}\label{sec:refine}
By considering the spatial distribution of the predicted intra-object part locations and the learned point-wise part features in a 3D box propsoal from stage-I, it is reasonable to aggregate all the information within a proposal for box proposal scoring and refinement. 
Based on the pooled 3D features, we train a sub-network to robustly aggregate information  to score box proposals and refine their locations.

\medskip
\noindent
{\bf Fusion of predicted part locations and semantic part features.} ~
As shown in the right part of Fig.~\ref{fig:framework_total}, we adopt the proposed RoI-aware point cloud pooling module to obtain discriminative features of each 3D proposal. Let $\mathbf{b}$ denote a single 3D proposal, and for all of its inside points (with canonical coordinate $\mathcal{X}=\{(x_i^{(ct)}, y_i^{(ct)}, z_i^{(ct)}) \in \mathbb{R}^3, i \in 1, \cdots, n\}$), we denote 
$\mathcal{F}_1=\{(x_i^{(part)}, y_i^{(part)}, z_i^{(part)}, s_i) \in \mathbb{R}^4, i \in 1, \cdots, n\}$ as their predicted point-wise part locations and semantic scores from stage-I, and denote $\mathcal{F}_2=\{f_i^{(sem)} \in \mathbb{R}^C, i \in 1, \cdots, n\}$ as their point-wise semantic part features learned by backbone network. 
Here $n$ is the total number of inside points of proposal $\mathbf{b}$. Then the part feature encoding of proposal $\mathbf{b}$ could be formulated as follows
\begin{align}
Q^{(part)} &= \text{RoIAwareAvgPool}\left(\mathcal{X}, ~\mathcal{F}_1,~ \mathbf{b}\right),\nonumber\\
Q^{(sem)} &= \text{RoIAwareMaxPool}\left(\mathcal{X}, ~\mathcal{F}_2,~ \mathbf{b}\right),\\
Q^{(roi)}_k &= \left[G(Q_k^{(part)}),~  Q_k^{(sem)}\right], 
k \in \{1, \cdots, L_x\times L_y \times L_z\},
\nonumber
\end{align}
where $G$ denotes a submanifold sparse convolution layer to transform the pooled part locations to the same feature dimensions $C$ to match $Q^{(sem)}$, and $[\cdot, \cdot]$ denotes feature concatenation. 
Here $Q^{(part)}$, $Q^{(sem)}$ and $Q^{(roi)}$ have the same spatial shape $(14\times14\times14$ by default). 
The fused features $Q^{(roi)}$ encode both geometric and semantic information of the box proposals by the backbone network.
Note that here we use the average pooling for pooling the predicted intra-object part locations $\mathcal{F}_1$ to obtain representative predicted part location of each voxel of the proposal, while we use the max pooling for pooling the semantic part features $\mathcal{F}_2$.

\medskip
\noindent
{\bf Sparse convolution for part information aggregation.} ~
For each 3D proposal, we need to aggregate fused features $Q^{(roi)}$ from all inside spatial locations of this proposal for robust box scoring and refinement. As shown in the right part of Fig.~\ref{fig:framework_total}, we stack several 3D sparse convolutional layers with kernel size $3\times 3 \times 3$ to aggregate all part features of a proposal as the receptive filed increases. Here we also insert a sparse max-pooling with kernel size $2 \times 2\times 2$ and stride $2$ between the sparse convolutional layers to down-sample the feature volume to $7\times7\times7$ for saving the computation cost and parameters. Finally we vectorize it to a feature vector (empty voxels are kept as zeros) and feed it into two branches for box scoring and location refinement.  

Compared with the naive method to directly vectorize the pooled 3D feature volume to a feature vector, our sparse convolution based part aggregation strategy could learn the spatial distribution of the predicted part locations effectively by aggregating features from local to global scales. The sparse convolution strategy also enables a larger $14\times 14 \times 14$ pooling size by saving much computations, parameters and GPU memory. 

\medskip
\noindent
{\bf 3D IoU guided box scoring.}
For the box scoring branch of stage-II, inspired by \cite{li2019gs3d, jiang2018acquisition}, we normalize the 3D Intersectoin-over-Union (IoU) between 3D proposal and its corresponding ground truth box as the soft label for proposal quality evaluation. The proposal quality $q^{(a)}$ is defined as
\begin{align}\label{iou_cls}
q^{(a)} &= 
\begin{cases}
1~~~&$if$~~\text{IoU} > 0.75,\\ 
0~~~&$if$~~\text{IoU} < 0.25, \\
2\text{IoU} - 0.5 & \text{otherwise},
\end{cases} 
\end{align}
which is also supervised by a binary cross entropy loss $\mathcal{L}_{\text{score}}$ defined similarly to Eq.~\eqref{loss:bce}.
Our experiments in Sec.~\ref{sec:ab_iou} show that comparing with the traditional classification based box scoring, the IoU guided box scoring leads to slightly better performance.

\subsection{Overall loss}\label{sec:train}
Our whole network is end-to-end trainable and the overall loss function is consist of the part-aware loss and part-aggregation loss. 

\medskip
\noindent
{\bf Losses of part-aware stage-I.}~
For the part-aware stage-I, the loss function consists of three terms with equal loss weights, including focal loss for foreground point segmentation, binary cross entropy loss for the regression of part locations and smooth-$L1$ loss for 3D proposal generation,
\begin{align}
\mathcal{L}_{\textrm{aware}} = \mathcal{L}_{\textrm{seg}} + \frac{1}{N_{\text{pos}}}\mathcal{L}_{\textrm{part}} + \lambda\frac{1}{M_{\text{pos}}}\mathcal{L}_{\textrm{box}}
\end{align}
where loss weight $\lambda=2.0$, $N_{\text{pos}}$ is the total number of foreground points, $M_{\text{pos}} = N_{\text{pos}}$ for Part-$A^2$-free model and $M_{pos}$ is the total number of positive anchors for Part-$A^2$-anchor model. 
For $\mathcal{L}_{\textrm{box}}$ loss, as mentioned in Sec.~\ref{sec:proposal_gen}, we adopt the bin-based box generation loss for Part-$A^2$-free, and adopt
the residual-based box regression loss for Part-$A^2$-anchor model. 

\medskip
\noindent
{\bf Losses of part-aggregation stage-II.}~
For the part-aggregation stage-II, the loss function includes a binary cross entropy loss term for box quality regression and a smooth-$L1$ loss term for 3D box proposal refinement,
\begin{align}
\mathcal{L}_{\textrm{aggregation}} = \mathcal{L}_{\textrm{score}} + \frac{1}{T_{\text{pos}}}\mathcal{L}_{\textrm{box\_refine}},
\end{align}
where $T_{\text{pos}}$ is the number of positive proposals, and we conducted the residual-based box regression loss for $\mathcal{L}_{\textrm{box\_refine}}$ as used in Eq.~\eqref{eqn:reg_targets}, which includes 
box center refinement loss, size refinement loss and angle refinement loss. 
Besides that, we also add the corner regularization loss $\mathcal{L}_{\text{corner}}$ as used in \cite{qi2017frustum}, and the final box refinement loss is as follows
\begin{align}
\scriptsize \mathcal{L}_{\textrm{box\_refine}} = \sum_{\text{res} \in \{x, y, z, l, h, w, \theta\}} \mathcal{L}_{\text{smooth-L1}}(\widehat{\Delta \text{res}^{(r)}}, \Delta \text{res}^{(r)}) + \mathcal{L}_{\text{corner}}
\end{align}
where $\widehat{\Delta \text{res}^{(r)}}$ is the predicted residual for the 3D proposal, $\Delta \text{res}^{(r)}$ is the corresponding ground-truth target calculated similarly to Eq.~\eqref{eqn:reg_targets}, 
and all losses here have the same loss weights. Note that here the angle refinement target is directly encoded as $\Delta\theta^{(r)}=(\theta^{(gt)} - \theta^{(r)})$ since the angle difference between proposals and their corresponding ground-truth boxes are within a small range due to the IoU constraint for positive proposals. 

\medskip
\noindent
{\bf Overall loss.}~
Hence the overall loss function of our Part-$A^2$ net for end-to-end training is calculated as
\begin{align}
\mathcal{L}_{\textrm{total}} = \mathcal{L}_{\textrm{aware}} + \mathcal{L}_{\textrm{aggregation}}
\end{align}
where the losses of these two stages have equal loss weights.

\subsection{Implementation details}\label{sec:imp_v2}

\begin{figure}
	\begin{center}
		%		\fbox{\rule{0pt}{2in} \rule{0.9\linewidth}{0pt}}
		\includegraphics[width=1.0\linewidth]{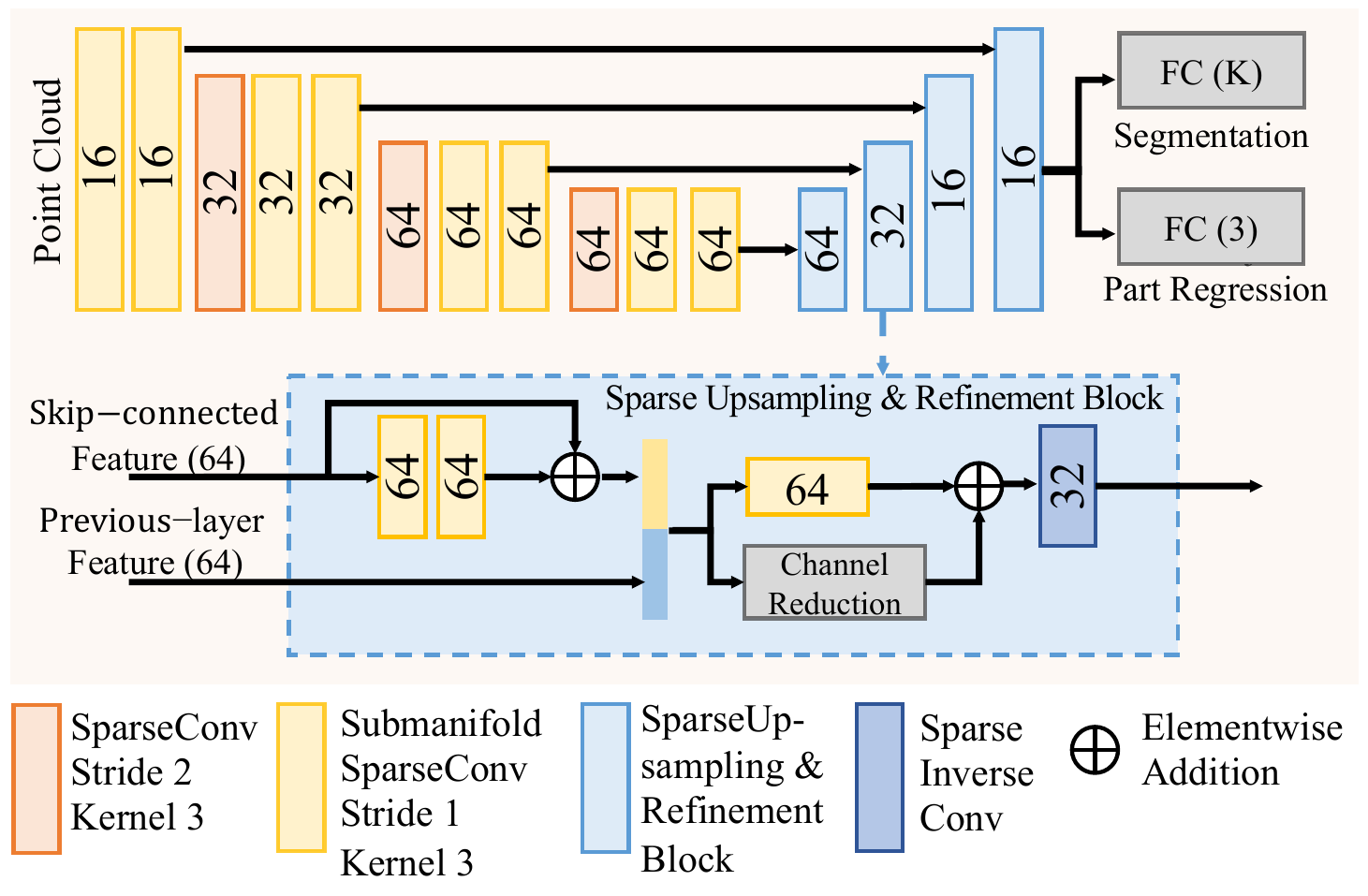}
	\end{center}
	\caption{The sparse convolution based encoder-decoder backbone network of part-aware stage-I of the Part-$A^2$-anchor model.}
	\label{fig:part_stage1}
\end{figure}

We design a UNet-like architecture~\cite{ronneberger2015u} for learning point-wise feature representations with 3D sparse convolution and 3D sparse deconvolution on the obtained sparse voxels. The spatial resolution is downsampled 8 times by three sparse convolutions of  stride 2, each of which is followed by several submanifold sparse convolutions. As illustrated in Fig.~\ref{fig:part_stage1}, we also design a similar up-sampling block as that in \cite{sun2018fishnet} based on sparse operations for refining the fused features. 

\noindent
{\bf Network details.} 
As shown in Fig.~\ref{fig:part_stage1}, for the part-aware stage-I of Part-$A^2$-anchor model, the spatial feature volumes have four scales with feature dimensions 16-32-64-64, and we use three 3D sparse convolution layers with kernel size $3\times 3 \times 3$ and stride $2$ to downsample the spatial resolution by 8 times. We stack two submanifold convolution layers in each level with kernel size $3\times 3\times 3$ and stride $1$. 
There are four sparse up-sampling blocks (see Fig.~\ref{fig:part_stage1} for network details) in the decoder to gradually increase feature dimensions as 64-64-32-16. Note that the stride of the last up-sampling block is 1 and the stride of other three up-sampling blocks is 2. For the Part-$A^2$-free net, we increase the feature dimensions of decoder to 128 in each scale and use simple concatenation to fuse the features from the same level of encoder and the previous layer of decoder,
since the learned point-wise features of decoder should encode more discriminative features for the bottom-up 3D proposal generation. 

For the part-aggregation stage, as shown in Fig.~\ref{fig:framework_total}, the pooling size of RoI-aware point cloud pooling module is $14\times 14\times 14$, which is downsampled to $7\times 7\times 7$ after processed by the sparse convolutions and max-pooling with feature dimensions 128. 
We vectorize the downsampled feature volumes to a single feature vector for the final box scoring and location refinement.

\medskip
\noindent
{\bf Training and inference details.}
We train the entire network end-to-end with the ADAM optimizer and a batch size 6 for 50 epochs. 
The cosine annealing learning rate strategy is used with an initial learning rate $0.001$. 
We randomly select 128 proposals from each scene for training stage-II with $1:1$ positive and negative proposals, where the positive proposals for box refinement have 3D IoU with their corresponding ground truth box of at least 0.55 for all classes, otherwise they are negative proposals, and the scoring target is defined as Eq.~\eqref{iou_cls} for the confidence prediction. 

We conduct common data augmentation during training, including random flipping, global scaling with scaling factor uniformly sampled from $[0.95, 1.05]$, global rotation around the vertical axis by an angle uniformly sampled from $[-\frac{\pi}{4}, \frac{\pi}{4}]$. In order to simulate objects with various environment as \cite{yan2018second}, we also randomly ``copy'' several ground-truth boxes with their inside points from other scenes and ``paste'' them to current training scenes. 
The whole training process of our proposed part-aware and part-aggregation networks takes about 17 hours on a single 
%NVIDIA TITAN Xp 
NVIDIA Tesla V100 GPU.

For inference, only 100 proposals are kept from part-aware stage-I with NMS threshold 0.85 for Part-$A^2$-free and 0.7 for Part-$A^2$-anchor, which are then scored and refined by the following part-aggregation stage-II. We finally apply the rotated NMS with threshold 0.01 to remove redundant boxes and generate the final 3D detection results. The overall inference time is about 70ms on a single Tesla V100 GPU card.
\textcolor{black}{\subsection{Pros and cons.}}
\textcolor{black}{
Our proposed 3D object detection framework has some advantages and disadvantages under different situations. }

\textcolor{black}{
Compared with the previous 3D object detection methods \cite{Chen2017CVPR,qi2017frustum,ku2018joint,zhou2018voxelnet,yan2018second,Liang2018ECCV,Liang2019CVPR,wang2019frustum}, 
(1) our proposed method for the first time introduces the learning of intra-object part locations to improve the performance of 3D object detection from point cloud, where the predicted intra-object part locations effectively encode the point distribution of 3D objects to benefit the 3D object detection. (2) The proposed RoI-aware feature pooling module eliminates the ambiguity of previous point cloud pooling operations and transforms the sparse point-wise features to the regular voxel features to encode the position-specific geometry  and semantic features of 3D proposals, which effectively bridges the proposal generation network and the proposal refinement network, and results in high detection accuracy. (4) Besides, the learning process of part locations could also be adopted to other tasks for learning more discriminative point-wise features, such as the instance segmentation of point cloud. The proposed RoI-aware pooling module could also be flexibly utilized on transforming the point-wise features from point-based networks (such as PointNet++) to the sparse voxel-wise features, that could be processed by more efficient sparse convolution networks. }

\textcolor{black}{
On the other hand, our method also has some limitations. 
Since our method aims at high performing 3D object detection in autonomous driving scenarios, some parts of our method could not be well applied for the 3D object detection of indoor scenes.
This is because the 3D bounding boxes in indoor scenes may overlap with each other (such as chairs under the table), therefore the 3D bounding box annotations of indoor scenes could not provide the accurate point-wise segmentation labels. Also, there are some categories whose orientation is not well-defined (such as the round tables), hence we could not generate accurate labels of the proposed intra-object part locations. }

\textcolor{black}{
However, our proposed anchor-free proposal generation strategy still shows great potential on the 3D proposal generation of indoor scenes since the indoor objects do not always stay on the ground and our anchor-free strategy avoids to set 3D anchors in the whole 3D space.
}
\\

\section{Experiments}
In this section, we evaluate our proposed method with extensive experiments on the challenging 3D detection benchmark of KITTI~\cite{Geiger2012CVPR} dataset. 
In Sec.~\ref{sec:ab_part}, we present extensive ablation studies and analysis to investigate individual components of our models. 
In Sec.~\ref{sec:exp_kitti}, we demonstrate the main results of our methods by comparing with state-of-the-art 3D detection methods. 
Finally we visualize some qualitative results of our proposed 3D detection model in Sec.~\ref{sec:exp4}.

\medskip
\noindent
{\bf Dataset.}
There are 7481 training samples and 7518 test samples in the dataset of KITTI 3D detection benchmark. The training samples are divided into the \textit{train} split (3712 samples) and \textit{val} split (3769 samples) as the frequently used partition of KITTI dataset. 
All models are only trained on the \textit{train} split, and evaluated on the \textit{val} and \textit{test} splits.

\medskip
\noindent
{\bf Models.}
There are three main models in our experiments, \ie, Part-$A^2$-free model, Part-$A^2$-anchor model and our preliminary PointRCNN model~\cite{shi2019pointrcnn}. 
The network details of Part-$A^2$-free and Part-$A^2$-anchor models have been demonstrated in Sec.~\ref{sec:method}, and the whole framework is illustrated in Fig.~\ref{fig:framework_total}. As discussed in Sec.~\ref{sec:proposal_gen}, the key differences between these two versions of Part-$A^2$ models is that Part-$A^2$-free generates 3D proposals in the bottom-up (anchor-free) manner, while the Part-$A^2$-anchor net generates 3D proposals with the proposed anchor-based scheme. 
PointRCNN is in the preliminary version of this work \cite{shi2019pointrcnn}. It utilizes PointNet++~\cite{qi2017pointnet++} to extract point-wise features, which are used to generate 3D proposals in a bottom-up manner via segmenting the foreground points as demonstrated in Sec.~\ref{sec:proposal_gen}. Furthermore, in the stage-II of PointRCNN, we pool the inside points and their point-wise features for each 3D proposals, which are then fed to a second PointNet++ encoder to extract features of each 3D proposal for proposal confidence prediction and 3D box proposal refinement.

\subsection{From points to parts: ablation studies for Part-$A^2$ net}\label{sec:ab_part}
In this section, we provide extensive ablation experiments and analysis to investigate the individual components of our proposed Part-$A^2$ net models.

\begin{table}
	\begin{center}
		\scalebox{1.0}{
			\begin{tabular}{c|cc|c}
				\hline
				Method & Backbone & 
				\tabincell{c}{Proposal\\Scheme} & Recall \\
				\hline
				PointRCNN & PointNet++ & Anchor-free & 74.81 \\
				Part-$A^2$-free & SparseConvUNet & Anchor-free & 81.54 \\
				Part-$A^2$-anchor & SparseConvUNet & Anchors-based & 85.58 \\
				\hline
			\end{tabular}
		}
	\end{center}
	\caption{Recall (with 100 proposals) of the proposal generation stage by different backbone network and different proposal generation strategy. The experiments are conducted on the car class at moderate difficulty of the \textit{val} split of KITTI dataset, and the evaluation metric is the 3D rotated IoU with threshold 0.7.
	}
	\label{tab:exp_recall_v2}
\end{table}

\subsubsection{SparseConvUNet v.s. PointNet++ backbones for 3D point-wise feature learning}\label{ab:backbone}
As mentioned in Sec.~\ref{sec:rpn_v2}, instead of utilizing PointNet++ as the backbone network, we design a sparse convolution based UNet (denoted as SparseConvUNet) for point-wise feature learning, and the network details are illustrated in Fig.~\ref{fig:part_stage1}. 
We first compare the PointRCNN with PointNet++ backbone and Part-$A^2$-free with SparseConvUNet backbone with the same loss functions to test these two different backbone networks.

Table~\ref{tab:exp_recall_v2} shows that our SparseConvUNet based Part-$A^2$-free ($2^{nd}$ row) achieves 81.54\% recall with 3D IoU threshold 0.7, which is 6.73\% higher than the recall of PointNet++ based PointRCNN ($1^{st}$ row), and it demonstrates that our new designed SparseConvUNet could learn more discriminative point-wise features from the point cloud for the 3D proposal generation. 
As shown in Table~\ref{tab:recall}, we also provide the recall values of different number of proposals for these two backbones. We could find the recall of the sparse convolution based backbone consistently outperforms the recall of the PointNet++ based backbone, which further validates that the sparse convolution based backbone is better than the PointNet++ based backbone for point-wise feature learning and 3D proposal generation. 

Table~\ref{tab:exp_recall_v2} and Table~\ref{tab:recall} also show that our Part-$A^2$-anchor model achieves higher recall than the Part-$A^2$-free model. 
Therefore, in our remaining experimental sections, we mainly adopt the Part-$A^2$-anchor model for ablation studies and experimental comparison unless specified otherwise.

\subsubsection{Ablation studies for RoI-aware point cloud pooling}
In this section, we designed ablation experiments to validate the effectiveness of our proposed RoI-aware point cloud pooling module with the Part-$A^2$-anchor model, and we also explored more pooling sizes to investigate the trend of performance when increasing the RoI pooling size. 

\medskip
\noindent
{\bf Effects of RoI-aware point cloud region pooling.}~
As discussed in Sec.~\ref{sec:roi}, the proposed RoI-aware point cloud pooling module normalizes different 3D proposals to the same coordinate system to encode geometric information of proposals. It solves the ambiguous encoding by previous 3D point cloud pooling schemes as shown in Fig.~\ref{fig:roipool}. The 3D proposals are divided into regular voxels to encode the position-specific features for each 3D proposal. 

To validate the effects of the RoI-aware pooling module, we conduct the following comparison experiments. 
(a) We replace RoI-aware pooling by fixed-sized RoI pooling, \ie pooling all 3D proposals with the same fixed-size ($l=3.9, w=1.6, h=1.56$ meters for car) 3D box calculated from the mean object size of the training set with $14\times 14 \times 14$ grids. The center and orientation of the 3D grid are set as its corresponding 3D proposal's center and orientation, respectively. This is very similar to the pooling scheme used in PointRCNN, where not all geometric information is well preserved during pooling.
(b) We replace sparse convolutions of stage-II with several FC layers. As shown in Table~\ref{tab:roiaware}, removing RoI-aware pooling substantially decreases detection accuracy, while replacing sparse convolutions of stage-II with FC layers achieves similar performance, which proves the effectiveness of our proposed RoI-aware pooling but not the sparse convolution contributes to the main improvements.

\begin{table}
	\small 
	\begin{center}
		\scalebox{0.92}{
			\begin{tabular}{cc|ccc}
				\hline
				\tabincell{c}{Pooling Scheme}   & Stage-II 
				& $AP_{Easy}$ & $AP_{Mod.}$ & $AP_{Hard}$ \\
				\hline
				RoI fixed-sized pool & sparse conv & 88.78 & 78.61 & 78.05 \\
				RoI-aware pool & FCs & 89.46 & 79.32 & {\bf78.77} \\
				RoI-aware pool & sparse conv & {\bf89.47} & {\bf 79.47} &  78.54 \\
				\hline
			\end{tabular}
		}
	\end{center}
	\caption{Effects of RoI-aware point cloud pooling by replacing the RoI-aware pooling or the sparse convolution, and the pooling sizes of all the settings are $14\times14\times14$. The results are the 3D detection performance of car class on the \textit{val} split of KITTI dataset.}
	\label{tab:roiaware}
\end{table}

\medskip
\noindent
{\bf Effects of RoI pooling size.} ~
The $14\times 14$ pooling size was very commonly chosen for 2D object detection, and we follow the same setting to use $14\times 14\times 14$ as the 3D RoI-aware pooling size.
We also test different RoI pooling sizes as shown in Table \ref{tab:roisize}. The pooling size shows robust performance for different 3D objects. Similar performances can be observed if the pooling sizes are greater than $12\times 12 \times 12$.

\begin{table}
	\small 
	\begin{center}
		\begin{tabular}{c|ccc}
			\hline
			RoI Pooling Size
			& $AP_{Easy}$ & $AP_{Mod.}$ & $AP_{Hard}$ \\
			\hline
			$6\times 6\times 6$ & 89.02 & 78.85 & 78.04 \\	
			$8\times 8\times 8$  & 89.09 & 78.97 & 78.15 \\
			$10\times 10\times 10$  & 89.44 & 79.15  & 78.42 \\
			$12\times 12\times 12$  & \textbf{89.61} & 79.35 & 78.50 \\
			$14\times 14\times 14$  & 89.47 & \textbf{79.47} & 78.54 \\
			$16\times 16\times 16$  & 89.52& 79.45 & \textbf{78.56}\\
			\hline
		\end{tabular}
	\end{center}
	\caption{Effects of using different RoI-aware pooling sizes in our part-aggregation stage. The results are the 3D detection performance of car class on the \textit{val} split of KITTI dataset.}
	\label{tab:roisize}
	\vspace{-5mm}
\end{table}

\subsubsection{Sparse convolution v.s. fully-connected layers for part aggregation.}
In our Part-$A^2$ net, after applying the RoI-aware point cloud pooling module, there are several ways to implement the part-aggregation stage. The simplest strategy is to directly vectorize the pooled feature volumes to a feature vector followed by several fully-connected layers for box scoring and refinement. From the $1^{st}$ row of Table~\ref{tab:pa}, we could see that this naive way already achieved promising results, which are benefited from the effective representations of our RoI-aware point cloud pooling since each position of the feature vector encodes a specific intra-object position of the object of interest to help learn the shape of the box better. 
In the $2^{nd}$ row of Table~\ref{tab:pa}, we further investigate using sparse convolution with kernel size $3\times 3 \times 3$ to aggregate the features from local to global scales gradually, which achieves slightly better results with the same pooling size $7\times7\times7$. 
The $3^{rd}$ row shows that fully-connected layers with larger pooling size $14\times14\times14$ achieves improved performance, but this design consumes much calculations and GPU memory. 
As we mentioned in Sec.~\ref{sec:refine}, our proposed part-aggregation network adopts a large pooling size $14\times 14 \times14$ to capture details and then use sparse max-pooling to downsample the feature volumes for feature encoding, which achieves the best performance in the easy and moderate difficulty levels as shown in Table~\ref{tab:pa} with lower computation and GPU memory cost than fully-connected layers.

\begin{table}
	\small 
	\begin{center}
		\setlength\tabcolsep{3pt}
		\scalebox{0.9}{
			\begin{tabular}{ccc|ccc}
				\hline
				\tabincell{c}{Pooling\\Size} & Stage-II & \tabincell{c}{Downsampling\\in Stage-II}
				& $AP_{Easy}$ & $AP_{Mod.}$ & $AP_{Hard}$ \\
				\hline	
				$7\times 7\times 7$ & FCs  & & 89.17 & 79.11 & 78.03 \\	% 
				$7\times 7\times 7$ & sparse conv & & 89.24 & 79.21 & 78.11 \\	% 
				$14\times14\times14$& FCs &&  89.46 & 79.32 & {\bf 78.77} \\
				$14\times14\times14$& sparse conv & $\checkmark$ &{\bf89.47} & {\bf 79.47} & 78.54 \\
				\hline
			\end{tabular}
		}
	\end{center}
	\caption{Comparison of several different part-aggregation network structures. The results are the 3D detection performance of car class on the \textit{val} split of KITTI dataset.}
	\label{tab:pa}
\end{table}

\subsubsection{Ablation studies for 3D proposal generation}\label{sec:ab_bottom_up}
We investigate the two strategies for 3D proposal generation from point cloud, one is the anchor-based strategy and the other is our proposed anchor-free strategy, in Part-$A^2$-anchor and Part-$A^2$-free models. In this section, 
we first experiment and discuss two proposal generation strategies in details to provide a reference to choose better strategy for different settings of 3D proposal generation. 
Then we compare the performance of different center regression looses for the two strategies. 

\medskip
\noindent
{\bf Anchor-free v.s. anchor-based 3D proposal generation.}\quad 
We validate the effectiveness of our proposal generation strategies with state-of-the-art two-stage 3D detection methods. 
As shown in Table~\ref{tab:recall}, our preliminary PointRCNN with anchor-free proposal generation and PointNet++ backbone already achieve significantly higher recall than previous methods. With only 50 proposals, PointRCNN obtains 96.01\% recall at IoU threshold 0.5, which outperforms recall 91\% of AVOD \cite{ku2018joint} by 5.01\% at the same number of proposals, note that the latter method uses both 2D image and point cloud for proposal generation while we only use point cloud as input. 

We also report the recall of 3D bounding box at IoU threshold 0.7 by our anchor-free and anchor-based strategies in Table~\ref{tab:recall}. 
Part-$A^2$-free model (with anchor-free proposal generation strategy) achieves 77.12\% recall at IoU threshold 0.7 with only 50 proposals, which is much higher than the recall of our preliminary work PointRCNN since Part-$A^2$-free model adopts better sparse convolution based backbone. 
Our Part-$A^2$-anchor model (with anchor-based proposal generation strategy) further improves the recall to 83.71\% at IoU threshold 0.7 with 50 proposals. 
This is because the anchor-based strategy has a large number of anchors to more comprehensively cover the entire 3D space to achieve a higher recall. However, the improvement comes with sacrifices, as it needs different sets of anchors for different classes at each spatial location. For instance, the anchor size of pedestrians is $(l=0.8m, w=0.6m, h=1.7m)$ while the anchor size of cars is $(l=3.9m, w=1.6m, h=1.56m)$. They are unlikely to share the same anchor. In contrast, our anchor-free strategy still generates a single 3D proposal from each segmented foreground point even for many classes, since we only need to calculate the 3D size residual with respect to the corresponding average object size based on its semantic label. 

\begin{table}
	\small 
	\begin{center}
		\setlength\tabcolsep{4pt}
		\scalebox{0.9}{
			\begin{tabular}{c|ccc|ccc}
				\hline 
				\multirow{3}{*}{RoIs \#} & 			
				\multicolumn{3}{c|}{Recall (IoU=0.5)} & 
				\multicolumn{3}{c}{Recall (IoU=0.7)} \\
				& MV3D & AVOD & 
				\tabincell{c}{PointRCNN} &
				\tabincell{c}{PointRCNN} & \tabincell{c}{Part-$A^2$\\-free} & 
				\tabincell{c}{\textbf{}Part-$A^2$\\-anchor}\\
				\hline\textbf{}
				10 & - &  86.00 & {\bf86.66} & 29.87 & 66.31 & \textbf{80.68}\\
				20 & - & - & {\bf91.83} & 32.55  & 74.46 & \textbf{81.64}\\
				30 & - & - & {\bf93.31} & 32.76 &  76.47 & \textbf{82.90} \\
				40 & - & - & {\bf95.55} & 40.04 & 76.88 & \textbf{83.05}\\
				50 & - & 91.00 & {\bf96.01} & 40.28 & 77.12 & \textbf{83.71} \\
				100 & - & - & {\bf96.79} & 74.81 & 81.54 & \textbf{85.58} \\
				200 & - & - & {\bf98.03} & 76.29 & 84.93 & \textbf{89.32}\\ 
				300 & 91.00 & - & {\bf98.21} & 82.29 & 86.03 & \textbf{91.64} \\
				\hline 
			\end{tabular}
		}
	\end{center}
	\caption{Recall of generated proposals by compared methods with different numbers of RoIs and 3D IoU thresholds for the car class at moderate difficulty of the \emph{val} split. Note that only MV3D \cite{Chen2017CVPR} and AVOD \cite{ku2018joint} of previous methods reported the recall rates of proposals.}
	\label{tab:recall}
\end{table}

The 3D detection results of the Part-$A^2$-free and Part-$A^2$-anchor models on cars, cyclists and pedestrians are reported in Table~\ref{tab:bu_part}. We could see that the 3D detection results of cyclist and pedestrian by Part-$A^2$-free are comparable to those by Part-$A^2$-anchor model, while the results of cars by Part-$A^2$-free are lower than those by Part-$A^2$-anchor on the moderate and easy difficulties. 
Hence the bottom-up Part-$A^2$-free model has better potential on multi-class 3D detection on small-size objects (such as cyclists and pedestrians) with lower memory cost, while the anchor-based Part-$A^2$-anchor model may achieve a slightly better performance on 3D detection of large-size objects such as cars. That is because 
the predefined anchors are closer to the center locations of objects with large sizes, while the bottom-up proposal generation strategy suffers from difficulty of regressing large residuals from the object surface points to object centers. 

\begin{table}
	\small 
	\begin{center}
		\scalebox{0.9}{
			\setlength\tabcolsep{4pt}
			\begin{tabular}{c|c|c|ccc}
				\hline
				Method & Class & IoU Thresh
				& $AP_{Easy}$ & $AP_{Mod.}$  & $AP_{Hard}$ \\
				\hline 
				Part-$A^2$-free & Car & 0.7  & 88.48 & 78.96  & 78.36 \\
				Part-$A^2$-anchor & Car & 0.7 & {\bf 89.47} & {\bf79.47} & {\bf78.54} \\
				\hline
				Part-$A^2$-free & Cyclist & 0.5 & 88.18 & {\bf73.35}  & {\bf70.75} \\
				Part-$A^2$-anchor & Cyclist & 0.5 & {\bf88.31} & 73.07 & 70.20 \\
				\hline
				Part-$A^2$-free & Pedestrian & 0.5 & {\bf70.73} & {\bf64.13}  & 57.45 \\
				Part-$A^2$-anchor & Pedestrian & 0.5 & 70.37 & 63.84 & {\bf57.48}\\
				\hline
			\end{tabular}
		}
	\end{center}
	\caption{3D object detection results of Part-$A^2$-free net and Part-$A^2$-anchor net on the KITTI \emph{val} split set.}
	\label{tab:bu_part}
	\vspace{-5mm}
\end{table}

\begin{figure}[t]
	%	\small
	\vspace{-0.5cm}
	\begin{center}
		\includegraphics[width=0.99\linewidth, height=6.5cm]{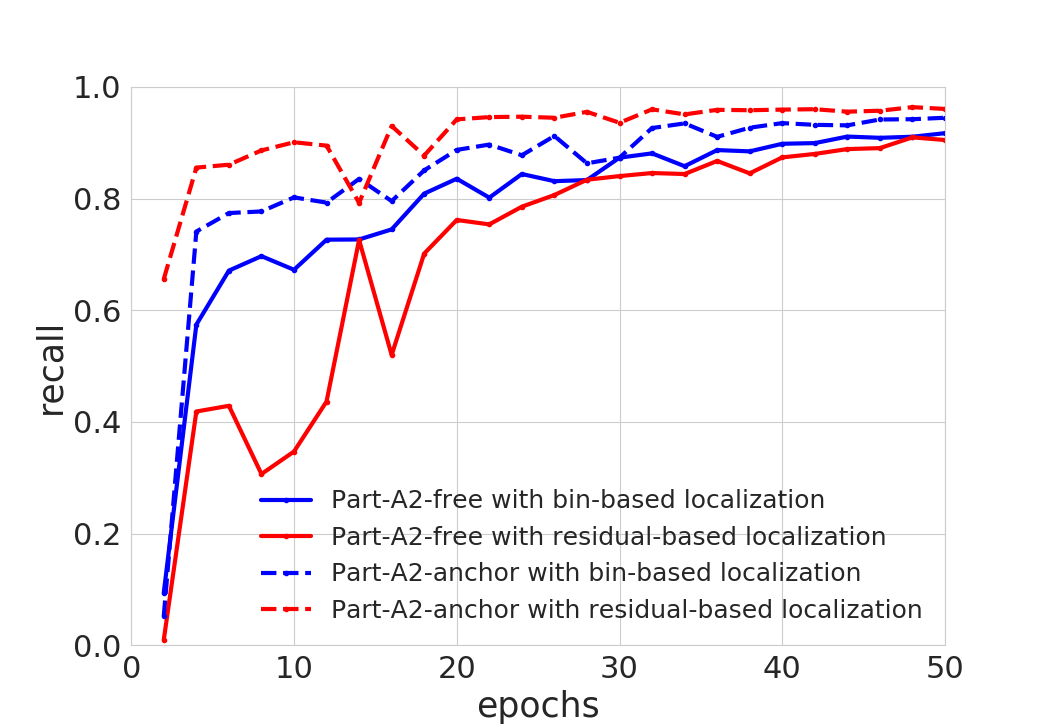}
	\end{center}
	\vspace{-3mm}
	\caption{Recall v.s. training iterations for different center regression losses of Part-$A^2$-free net and Part-$A^2$-anchor net. The results are generated according to the generated proposals of Part-$A^2$-free net and Part-$A^2$-anchor net on the car class of the \textit{val} split with IoU threshold 0.5.}
	\label{fig:loss_recall_part}
\end{figure}

\smallskip
\noindent
{\bf Center regression losses of 3D bounding box generation. }\quad 
We compare different center regression losses on our Part-$A^2$-free net and our Part-$A^2$-anchor net, including the proposed bin-based regression loss (Eq.~\eqref{eqn:bin-based}) and the residual-based regression loss (first row of Eq.~\eqref{eqn:reg_targets}). As shown in Fig.~\ref{fig:loss_recall_part}, for the Part-$A^2$-free net with anchor-free proposal generation strategy, the bin-based regression loss (solid blue line) converges faster than the residual-based regression loss (solid red line). In contrast, for the Part-$A^2$-anchor net with anchor-based proposal generation scheme, the residual-based regression loss (dashed red line) converges faster and better than the bin-based regression loss (dashed blue line). It demonstrates that the proposed bin-based center regression loss is more suitable with the anchor-free proposal generation strategy to achieve better performance, since the center regression targets of anchor-free scheme (generally from a surface point to the object center) vary a lot and bin-based localization could better constrain the regression targets and make the convergence faster and more stable. Fig.~\ref{fig:loss_recall_part} shows that the Part-$A^2$-anchor net achieves better recall with the residual-based center regression loss 
and we also adopt residual-based center localization loss for the Part-$A^2$-anchor net as mentioned in Sec.~\ref{sec:proposal_gen}. 

\subsubsection{Benefits of intra-object part location prediction for 3D object detection}
To validate the effectiveness of utilizing the free-of-charge intra-object part locations for 3D detection, we test removing the part location supervisions from our Part-$A^2$-anchor model. In the backbone network, we only remove the branch for predicting intra-object part locations and keep other modules unchanged. The point-wise part locations for RoI-aware pooling are replaced with the canonical coordinates of each point. 

As shown in Table~\ref{tab:exp_part}, compared with the model trained without intra-object part location supervisions (the $3^{rd}$ vs the $4^{th}$ rows), the models with part location supervisions achieves better recall and average precision on all difficulty levels of the \textit{val} split of the car class. The remarkable improvements on recall and precision indicate that the network learned better 3D features for scoring box and refining locations for  detection with detailed and accurate supervisions of the intra-object part locations.

\subsubsection{One-stage v.s. two-stage 3D object detection}

Table~\ref{tab:exp_part} shows that
without the stage-II for box scoring and refinement, the proposal recalls of our first proposal stage are comparable ($80.90$ v.s. $80.99$). However, the performance improves significantly ($82.92$ v.s. $84.33$) after the 100 proposals are refined by the part-aggregation stage. It demonstrates that the predicted intra-object part locations are beneficial for stage-II, and our part-aggregation stage-II could effectively aggregate the predicted intra-object part locations to improve the quality of the predicted 3D boxes. The performance gaps between stage-I and stage-II ($1^{st}$ row v.s. $3^{rd}$ row, $2^{nd}$ row v.s. $4^{th}$ row) also demonstrate that our stage-II improves the 3D detection performance significantly by re-scoring the box proposals and refining their box locations.

\begin{table}
	\small 
	\begin{center}
		\scalebox{0.9}{
			\setlength\tabcolsep{4pt}
			\begin{tabular}{ccc|c|ccc}
				\hline
				Stage-I & 
				Stage-II & 
				\tabincell{c}{Part \\Prediction} & 
				\tabincell{c}{Recall \\ box\#100}	
				& $AP_{Easy}$ & $AP_{Mod.}$ & $AP_{Hard}$ \\
				\hline
				\checkmark &  &  & 80.90 & 88.48  & 77.97 & 75.84\\
				\checkmark & & \checkmark & 80.99 & 88.90  & 78.54 & 76.44\\
				\hline
				\checkmark & \checkmark & & 82.92 & 89.23 & 79.00 & 77.66 \\
				\checkmark & \checkmark & \checkmark & {\bf 84.33} & {\bf89.47} & {\bf 79.47} & {\bf 78.54} \\
				\hline
			\end{tabular}
		}
	\end{center}
	\caption{Effects of intra-object part location supervisions and stage-II refinement module, and the evaluation metrics are the recall and average precision with 3D rotated IoU threshold 0.7. 
	The results are the 3D detection performance of car class on the \textit{val} split of KITTI dataset, 
	and the detection results of stage-I are generated by directly applying NMS to the box proposals from stage-I. 
	}
	\label{tab:exp_part}
\end{table}

\subsubsection{Effects of IoU guided box scoring}\label{sec:ab_iou}
As mentioned in Sec.~\ref{sec:refine}, we apply the normalized 3D IoU to estimate the quality of the predicted 3D boxes, which is used as the ranking score in the final NMS (non-maximum-suppression) operation to remove the redundant boxes. Table~\ref{tab:nms_score} shows that comparing with the traditional classification score for NMS, our 3D IoU guided scoring method increases the performance marginally in all difficulty levels, which validates the effectiveness of using normalized 3D IoU to indicate the quality of predicted 3D boxes. 

\begin{table}
	\small 
	\begin{center}
		\begin{tabular}{c|ccc}
			\hline
			NMS Ranking Score
			& $AP_{Easy}$ & $AP_{Mod.}$ & $AP_{Hard}$ \\
			\hline
			 classification & 89.13 & 78.81 & 77.95 \\
			 3D IoU guided scoring & {\bf89.47} & {\bf 79.47} & {\bf 78.54} \\
			 \hline 
		\end{tabular}
	\end{center}
	\caption{Effects of 3D IoU guided box scoring for ranking the quality of the predicted 3D boxes. 	The results are the 3D detection performance of car class on the \textit{val} split of KITTI dataset.}
	\label{tab:nms_score}
\end{table}

\subsubsection{Memory cost of anchor-free and anchor-based proposal generation}
As shown in Table~\ref{tab:exp_params}, we compare the model complexity of the anchor-free and anchor-based proposal generation strategies by calculating the number of parameters and the number of generated boxes with different number of object classes. Part-$A^2$-free model (with anchor-free proposal generation strategy) consistently generates $\sim$16k proposals (\ie, the number of points of the point cloud), which is independent with the number of classes, while the  number of generated boxes (\ie, the predefined anchors) of Part-$A^2$-anchor model (with anchor-based proposal generation), increases linearly with the number of classes since each class has its own anchors with specified object sizes for each class. 
The number of anchors of Part-$A^2$-anchor model achieves 211.2k for detecting objects of 3 classes, which shows that our anchor-free proposal generation strategy is a relatively light-weight and memory efficient strategy especially for multiple classes. 

We also report the inference GPU memory cost for three classes detection (car, pedestrian and cyclist) on KITTI \cite{Geiger2012CVPR} dataset. 
The inference is conducted by PyTorch framework on a single NVIDIA TITAN Xp GPU card.
For the inference of a single scene, Part-$A^2$-free model consumes about 1.16GB GPU memory while Part-$A^2$-anchor model consumes 1.63GB GPU memory. For the inference with six scenes simultaneously, Part-$A^2$-free model consumes about 3.42GB GPU memory while Part-$A^2$-anchor model consumes 5.46GB GPU memory. 
It demonstrates that the Part-$A^2$-free model (with anchor-free proposal generation strategy) is more memory efficient than Part-$A^2$-anchor model (with anchor-based proposal generation). 

\begin{table}
	\begin{center}
		\scalebox{0.98}{
		\begin{tabular}{c|cc|cc}
			\hline
			\multirow{3}{*}{\tabincell{c}{Number of \\classes}} & 			
			\multicolumn{2}{c|}{Part-$A^2$-free} & 
			\multicolumn{2}{c}{Part-$A^2$-anchor} 
			\\
			 &
			\tabincell{c}{Number of \\parameters} & 
			\tabincell{c}{Number of \\generated boxes} &
			\tabincell{c}{Number of \\parameters} & 
			\tabincell{c}{Number of \\ anchors}\\
			\hline 
			1 & 1775269 & $\sim$16k & 4648588 & 70.4k \\
			3 & 1775527 & $\sim$16k & 4662952 & 211.2k \\
			\hline 
		\end{tabular}}
	\end{center}
	\caption{The number of parameters on proposal generation head, and the number of generated boxes with different number of classes for Part-$A^2$-free model and Part-$A^2$-anchor model. The parameters of (5, 10, 100) are counted by setting faked number of classes and the number of generated boxes are for the KITTI scene. }
	\label{tab:exp_params}
\end{table}

\subsubsection{Analysis of false positive samples}

\begin{figure}[t]
	%	\small
	\begin{center}
		\includegraphics[width=1.0\linewidth]{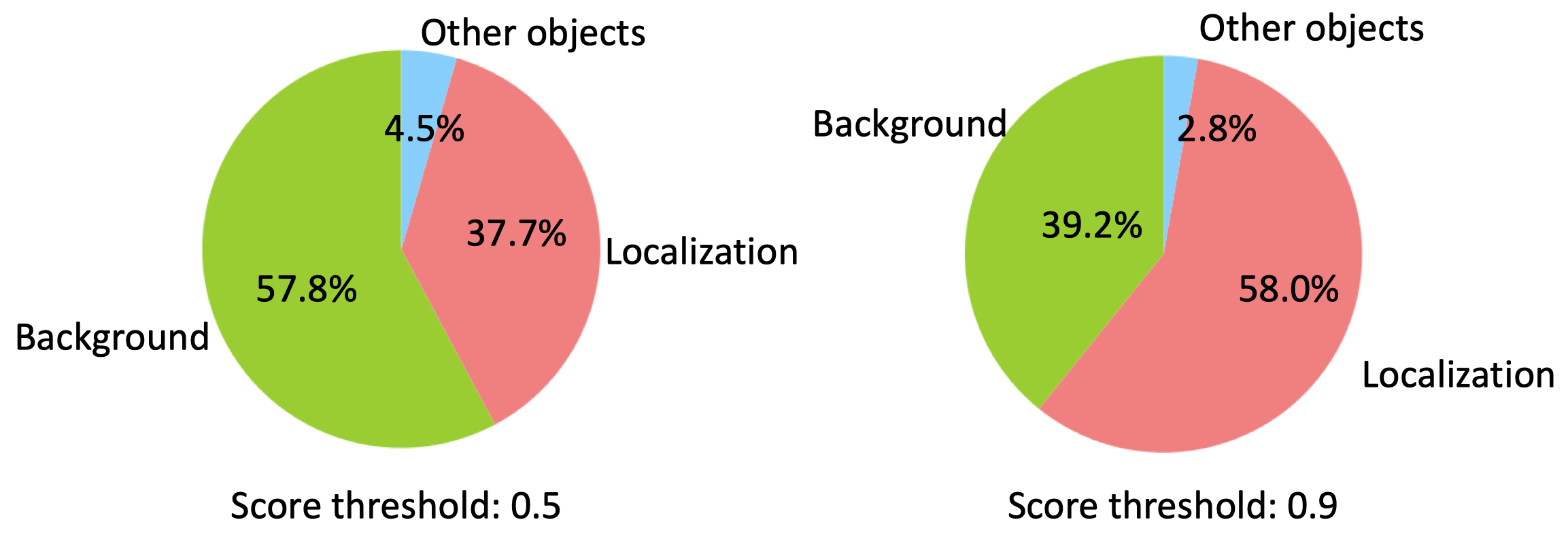}
	\end{center}
	\vspace{-3mm}
	\caption{Ratios of high-scored false positives for the car class on the \textit{val} split of KITTI dataset that are due to poor localization, confusion with other objects, or confusion with background or unlabeled objects. }
	\label{fig:fp_statistics}
\end{figure}

Fig.~\ref{fig:fp_statistics} shows the ratios of false positives of our best performance Part-$A^2$-anchor model on the KITTI validation dataset with different score thresholds, which are caused by confusion with background, poor localization, and confusion with objects from other categories. It can be seen that, the majority of false positives are from background and poor localization. The confusion of background mainly comes from the fact that the sparse point cloud could not provide enough semantic information for some background like the flower terrace. The LiDAR-only 3D detection methods may mistakenly recognize them as foreground objects like car since they have similar geometry shape in the point cloud. The ratio of false positives from poor localization increases significantly with the increasing score threshold. This is because the evaluation requirement of 3D rotated IoU constraint for 3D detection is more strict than the evaluation metric of 2D detection.

\begin{table*}
	\small 
	\begin{center}
	\scalebox{0.78}{
		\setlength\tabcolsep{3pt}
		\begin{tabular}{c|c||ccc|ccc|ccc|ccc|ccl|ccl}
			\hline
			\multirow{2}{*}{Method} & 
			\multirow{2}{*}{Modality} & 			
			\multicolumn{3}{c|}{~~3D Detection (Car)~~} & \multicolumn{3}{c|}{~BEV Detection (Car)~} &
			\multicolumn{3}{c|}{~3D Detection (Ped.)} & \multicolumn{3}{c|}{BEV Detection (Ped.)} &  \multicolumn{3}{c|}{~3D Detection (Cyc.)} & \multicolumn{3}{c}{BEV Detection (Cyc.)}\\
			&&\underline{Mod.} & Easy & Hard &
			\underline{Mod.} & Easy & Hard & 
			\underline{Mod.} & Easy & Hard & 
			\underline{Mod.} & Easy & Hard & 
			\underline{Mod.} & Easy & Hard & 
			\underline{Mod.} & Easy & Hard\\
			\hline
			MV3D \cite{Chen2017CVPR} & RGB + LiDAR & 62.35 & 71.09 & 55.12& 76.90  & 86.02 & 68.49 & 
			- & - & - & - & - & - 
			& - & - & - & - & - & -\\
			ContFuse \cite{Liang2018ECCV} & RGB + LiDAR & 66.22 & 82.54 & 64.04 & 85.83 & 88.81 & 77.33 & 
			- & - & - & - & - & - &
			- & - & - & - & - & -\\ 
			AVOD-FPN \cite{ku2018joint} & RGB + LiDAR & 71.88 & 81.94 & 66.38 & 83.79 & 88.53 & 77.90 & 
			42.81 & 50.80 & 40.88 & 51.05 & 58.75 & 47.54 &
			52.18 & 64.00  & 46.61 & 57.48 & 68.09 & 50.77 \\			
			F-PointNet \cite{qi2017frustum} & RGB + LiDAR & 70.39 & 81.20  & 62.19 & 84.00 & 88.70 & 75.33 &
			44.89 & 51.21 & 40.23 & 50.22 & 58.09 & 47.20 &
			56.77 & 71.96 & 50.39 & 61.96 & 75.38 & 54.68 \\
			PC-CNN-V2 \cite{8461232} & RGB + LiDAR & 73.80 & 84.33 & 64.83 & 86.10 & 88.49 & 77.26 & 
			- & - & - & - & - & - &
			- & - & - & - & - & -\\
			UberATG-MMF \cite{Liang2019CVPR} & RGB + LiDAR & 76.75 & 86.81 & 68.41 & 87.47 & 89.49 & 79.10 & 
		- & - & - & - & - & - &
			- & - & - & - & - & -\\	
			\hline 		
			VoxelNet \cite{zhou2018voxelnet}& LiDAR only & 65.11 & 77.47 & 57.73 & 79.26 & 89.35 & 77.39 &
			33.69 & 39.48 & 31.51 & 40.74 & 46.13 & 38.11 &
			48.36 & 61.22 & 44.37 & 54.76 & 66.70 & 50.55 \\
			SECOND \cite{yan2018second} & LiDAR only & 73.66& 83.13  & 66.20 & 79.37 & 88.07  & 77.95 & 
			42.56 & 51.07 & 37.29 & 46.27 & 55.10 & 44.76 &
			53.85 & 70.51  & 40.90 & 56.04 & 73.67 & 48.78\\
			PointPillars \cite{lang2018pointpillars} & LiDAR only & 74.99  & 79.05 & 68.30 & {\bf86.10} & 88.35 & 79.83 & 
			43.53 & 52.08 & 41.49 & 50.23 & 58.66 & 47.19 &
			59.07 & 75.78 & 52.92 & 62.25 & 79.14 & 56.00 \\
			%				\hline 
			PointRCNN (Ours) & LiDAR only & 75.76 & {\bf85.94} & 68.32  & 85.68 & 89.47 & 79.10 & 
			41.78 & 49.43 & 38.63 & 47.53 & 55.92 & 44.67  &
			59.60 & 73.93  & 53.59 & 66.77 & 81.52 & 60.78 \\
			Part-$A^2$-anchor (Ours) & LiDAR only & {\bf77.86} & {\bf85.94} & {\bf72.00} & 84.76 & {\bf89.52}  & {\bf81.47} &  
			{\bf44.50} & {\bf54.49} & {\bf42.36} & {\bf51.12} & {\bf59.72} & {\bf48.04} &  
			{\bf62.73} & {\bf78.58}  & {\bf57.74} & {\bf68.12} & {\bf81.91} & {\bf61.92} \\
			\hline
		\end{tabular}
	}
\end{center}
	\caption{Performance evaluation on KITTI official test server (\textit{test} split). The 3D object detection and bird's eye view detection are evaluated by mean average precision with 11 recall positions. The rotated IoU threshold is 0.7 for car and 0.5 for pedestrian/cyclist. 
	}
	\label{tab:test}
	\vspace{-0.3cm}
\end{table*}

\begin{table}
	\vspace{-0.3cm}
	\small 
	\begin{center}
		\scalebox{0.92}{
			\setlength\tabcolsep{2pt}
			\begin{tabular}{c|c|c|ccc}
				\hline
				\multirow{2}{*}{Method} & 
				\multirow{2}{*}{Reference} & 
				\multirow{2}{*}{Modality} &
				\multicolumn{3}{c}{AP (IoU=0.7)}\\
				&&& \underline{Mod.} & Easy  & Hard \\
				
				\hline
				MV3D \cite{Chen2017CVPR} & CVPR 2017 & RGB \& LiDAR& 62.68  & 71.29 & 56.56 \\
				ContFuse\cite{Liang2018ECCV} & ECCV 2018  & RGB \& LiDAR& 73.25 & 86.32  & 67.81 \\
				AVOD-FPN \cite{ku2018joint} & IROS 2018 & RGB \& LiDAR & 74.44 & 84.41  & 68.65 \\			
				F-PointNet \cite{qi2017frustum} & CVPR 2018  & RGB \& LiDAR& 70.92& 83.76  & 63.65 \\ 
				\hline 	
				VoxelNet \cite{zhou2018voxelnet} & CVPR 2018 & LiDAR only & 65.46& 81.98  & 62.85 \\
				SECOND \cite{yan2018second} & Sensors 2018 & LiDAR only & 76.48 & 87.43  & 69.10 \\
				%				\hline 
				PointRCNN (Ours) & CVPR 2019 & LiDAR only & 78.63 & 88.88  & 77.38 \\
				Part-$A^2$-free (Ours) & - & LiDAR only & 78.96 & 88.48 & 78.36 \\
				Part-$A^2$-anchor (Ours) & - & LiDAR only & {\bf 79.47} &{\bf89.47} & {\bf 78.54} \\
				\hline
			\end{tabular}
		}
	\end{center}
	\caption{Performance comparison of 3D object detection on the car class of the KITTI \emph{val} split set.}
	\label{tab:val}
\end{table}

\begin{table}
	\small 
	\vspace{-0.3cm}
	\begin{center}
		\scalebox{0.92}{
			\setlength\tabcolsep{1.9pt}
			\begin{tabular}{c|c|c|ccc}
				\hline
				\multirow{2}{*}{Method} & 
				\multirow{2}{*}{Reference} & 
				\multirow{2}{*}{Modality} &
				\multicolumn{3}{c}{AP (IoU=0.7)}\\
				& & & \underline{Mod.} & Easy & Hard \\
				\hline
				MV3D \cite{Chen2017CVPR} & CVPR 2017 & RGB \& LiDAR & 78.10 & 86.55 & 76.67 \\			
				F-PointNet \cite{qi2017frustum} & CVPR 2018& RGB \& LiDAR  & 84.02 & 88.16 & 76.44 \\
				ContFuse \cite{Liang2018ECCV} & ECCV 2018 & RGB \& LiDAR & 87.34 & 95.44  & 82.43 \\
				\hline
				VoxelNet \cite{zhou2018voxelnet} & CVPR 2018 & LiDAR only & 84.81 & 89.60  & 78.57 \\
				SECOND \cite{yan2018second} & Sensors 2018 & LiDAR only & 87.07 & 89.96 & 79.66 \\
				%				\hline 
				PointRCNN (Ours) & CVPR 2019 & LiDAR only & 87.89 & 90.21 & 85.51 \\
				Part-$A^2$-free (Ours) & - & LiDAR only & 88.05 & 90.23 & 85.85 \\ 
				Part-$A^2$-anchor (Ours) & - & LiDAR only & {\bf 88.61} & {\bf 90.42}  & {\bf 87.31} \\
				\hline
			\end{tabular}
		}
	\end{center}
	\caption{Performance comparison of bird-view object detection on the car class of the KITTI \emph{val} split set.}
	\label{tab:val_bev}
\end{table}

\begin{table}
	\small 
	\begin{center}
		\scalebox{0.99}{
			\begin{tabular}{c|c|ccc}
				\hline
				\multirow{2}{*}{Method} & \multirow{2}{*}{Class} &
				\multicolumn{3}{c}{AP (IoU=0.5)}\\
				& & \underline{Mod.} & Easy & Hard \\
				\hline 
				PointRCNN & Cyclist & 69.70 & 86.13 & 65.40 \\
				Part-$A^2$-free & Cyclist & {\bf73.35} & 88.18  & {\bf70.75} \\
				Part-$A^2$-anchor & Cyclist & 73.07 & {\bf88.31}  & 70.20 \\
				\hline
				PointRCNN & Pedestrian & 63.70 & 69.43 & {\bf 58.13} \\
				Part-$A^2$-free & Pedestrian & {\bf64.13} & {\bf70.73}  & 57.45 \\
				Part-$A^2$-anchor & Pedestrian & 63.84 & 70.37  & 57.48 \\
				\hline
			\end{tabular}
		}
	\end{center}
	\caption{3D object detection results of cyclist and pedestrian of different models on the KITTI \emph{val} split set.}
	\label{tab:people}
\end{table}

\begin{table}
	\small 
	\begin{center}
		\scalebox{0.95}{
			\setlength\tabcolsep{2pt}
			\begin{tabular}{c|ccc|c}
				\hline 
				& mAbsError$_x$ & mAbsError$_y$ & mAbsError$_z$ & mAbsError \\
				\hline 
				Overall & 7.24\% & 6.42\% & 5.17\% & 6.28\% \\
				False Positives & 12.97\% & 12.09\% & 7.71\% & 10.92\% \\
				\hline 
			\end{tabular}
		}
	\end{center}
	\caption{Mean distance error of predicted intra-object part locations by part-aware stage for the car class of KITTI \textit{val} split. As shown in Fig.~4, here $x, y, z$ are along the direction of width, length and height of the object, respectively. \textcolor{black}{Note that here the false positives denotes the false positives samples caused by inaccurate localizations.}}
	\label{tab:merror}
\end{table}

\begin{table}
	\small 
	\begin{center}
		\scalebox{1.0}{
			\begin{tabular}{cccc}
				\hline 
				x-axis & y-axis & z-axis & Overall \\
				\hline 
				0.468 & 0.442 & 0.552 & 0.531 \\
				\hline 
			\end{tabular}
		}
	\end{center}
	\caption{\textcolor{black}{Pearson correlation coefficient between the errors of the  predicted intra-object part locations and the errors of the predicted 3D bounding boxes.}}
	\label{tab:correlation}
	\vspace{-3mm}
\end{table}

\subsection{Main results and comparison with state-of-the-arts on KITTI benchmark}\label{sec:exp_kitti}
In this section, we report the comparison results with state-of-the-art 3D detection methods on the KITTI benchmark. We mainly report the performance of our Part-$A^2$-anchor model as it is able to reach higher accuracy in our ablation studies.

\medskip
\noindent
{\bf Comparison with state-of-the-art 3D detection methods.}  ~
We evaluate our methods on the 3D detection benchmark and the bird's eye view detection benchmark of the KITTI test split, whose results are evaluated on KITTI's offical test server. The results are shown in Table~\ref{tab:test}. 

\textcolor{black}{For the 3D object detection benchmark, by only using LiDAR point clouds, our proposed Part-$A^2$ net outperforms all previous peer-reviewed LiDAR only methods on all difficulty levels for all the three classes, and outperforms all previous multi-sensor methods on the most important ``moderate`` difficulty level for both car and cyclist classes. 
For the bird's view detection of car, pedestrian and cyclist, our method outperforms previous state-of-the-art methods by large margins on almost all the difficulty levels.} 
As of August 15, 2019, our proposed Part-$A^2$-anchor net ranks $1^{st}$ place among all methods on the most important car class of 3D object detection leaderboard of KITTI 3D Object Detection Benchmark~\cite{kitti_leaderboard}, while our method also ranks $1^{st}$ among all LiDAR-only methods on the cyclist class.

\begin{figure}
	\begin{center}
		\includegraphics[width=0.95\linewidth]{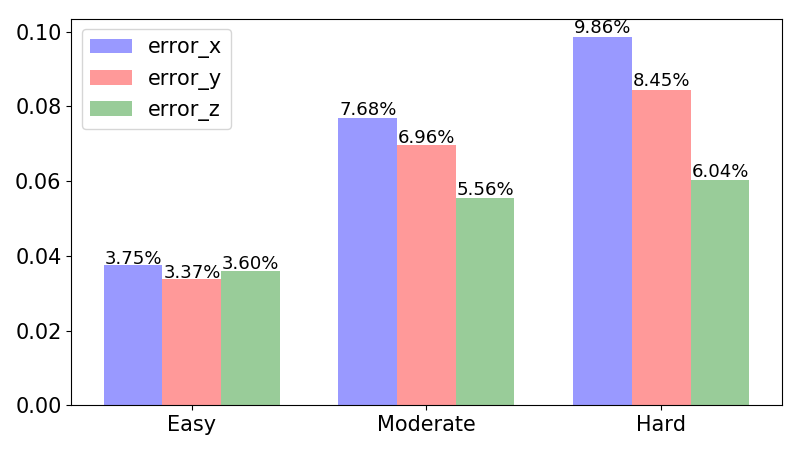}
	\end{center}
	\caption{Statistics of predicted intra-object part location errors for the car class on the \textit{val} split of KITTI dataset.}
	\label{fig:partstat}
\end{figure}

\begin{figure*}
	%	\vspace{-5mm}
	\centering
	\small
	\scalebox{0.975}{
		\begin{tabular}{@{\hspace{0.0mm}}c@{\hspace{1.0mm}}c@{\hspace{1.0mm}}c@{\hspace{1.0mm}}c}
			\includegraphics[width=0.25\linewidth]{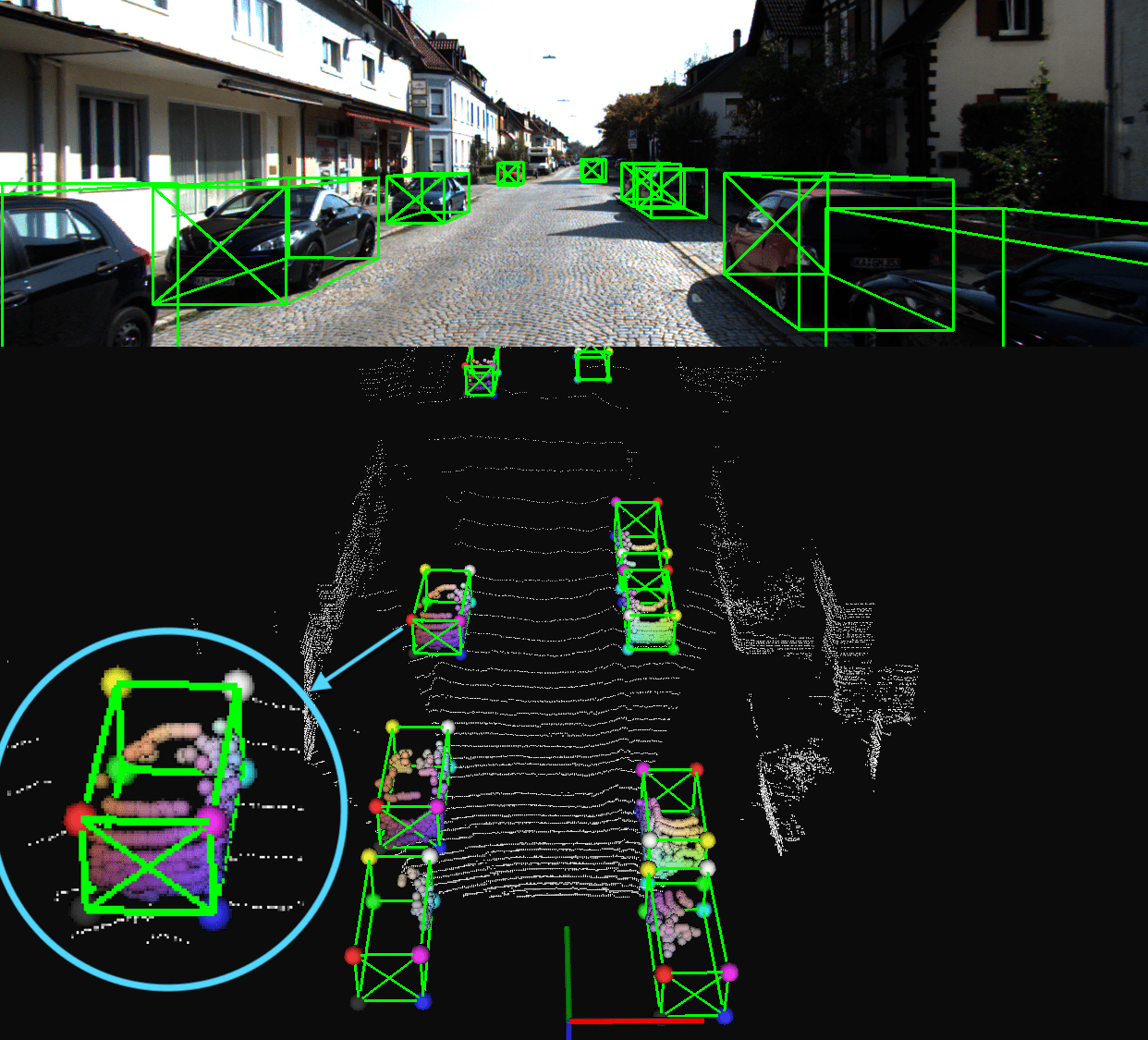}&
			\includegraphics[width=0.25\linewidth]{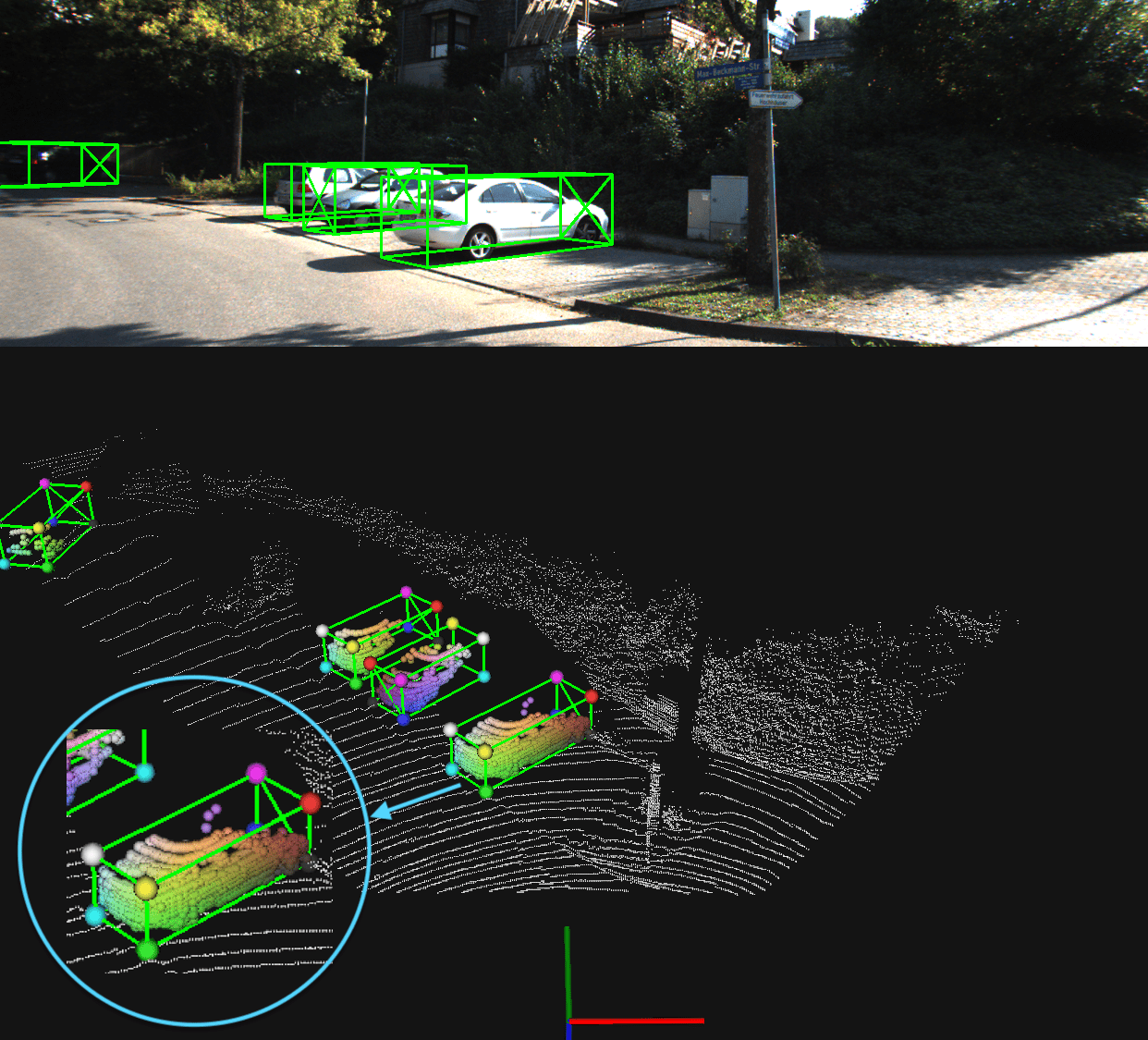}&
			\includegraphics[width=0.25\linewidth]{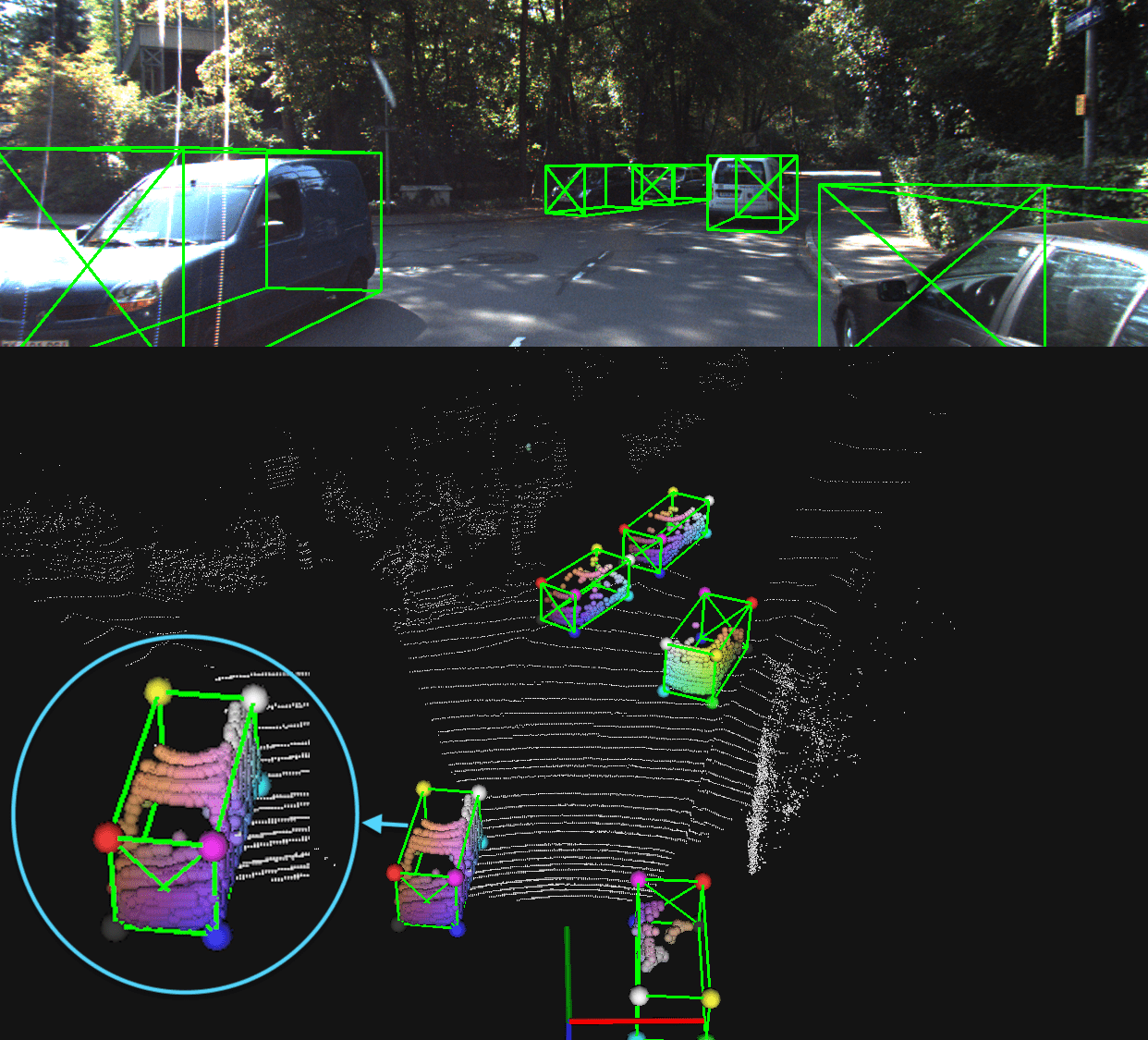}&
			\includegraphics[width=0.25\linewidth]{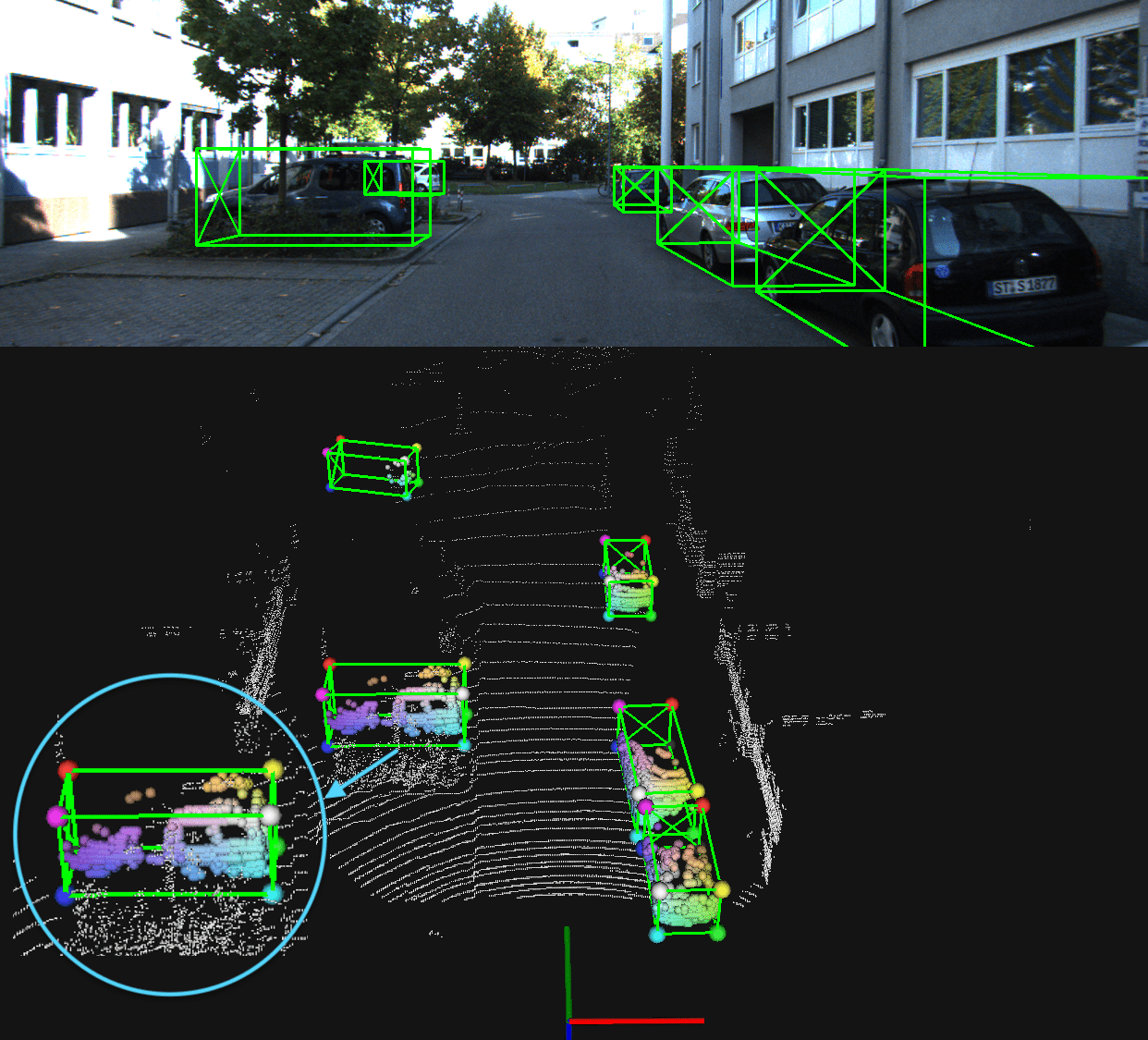}\\
			\includegraphics[width=0.25\linewidth]{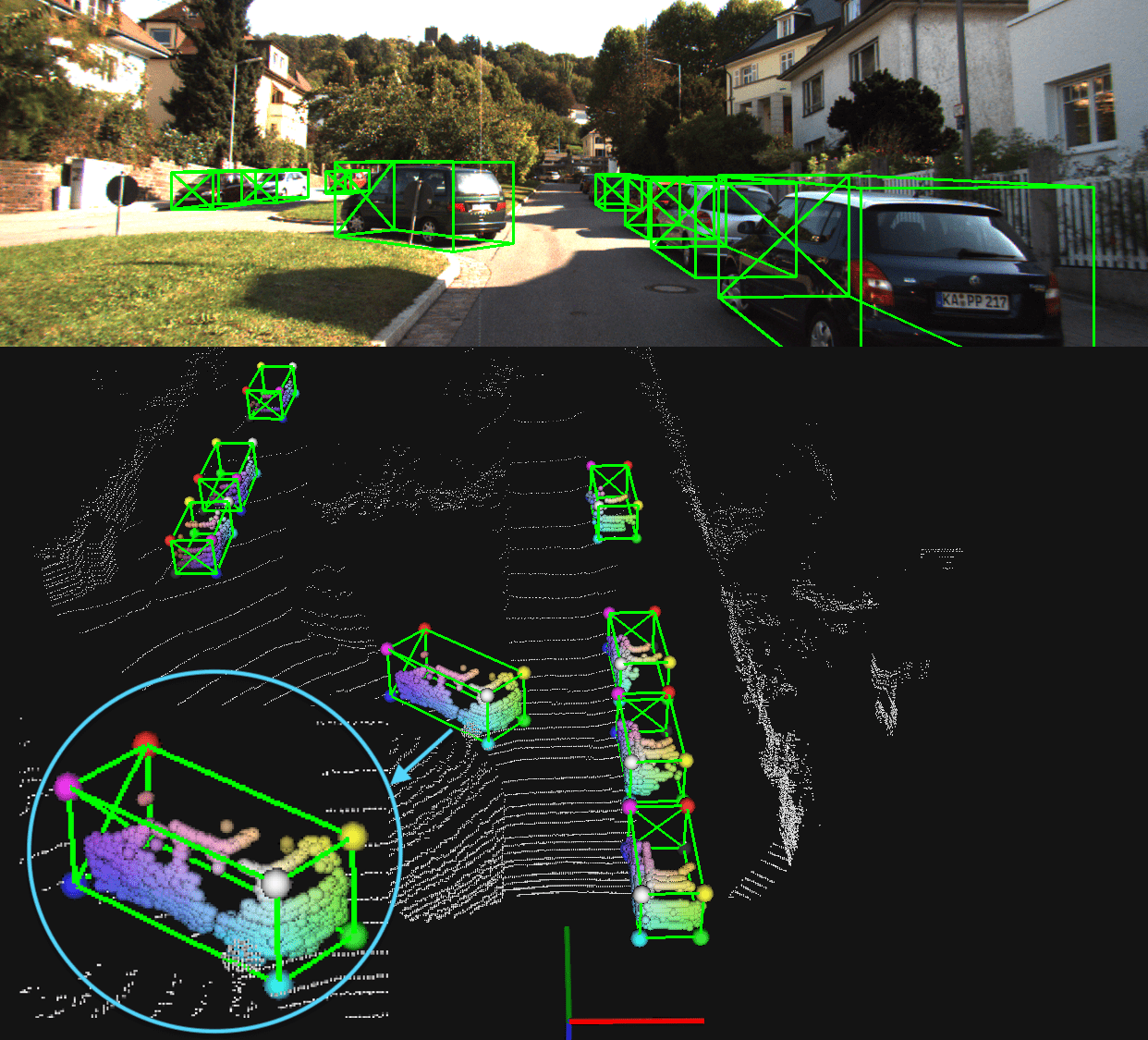}&
			\includegraphics[width=0.25\linewidth]{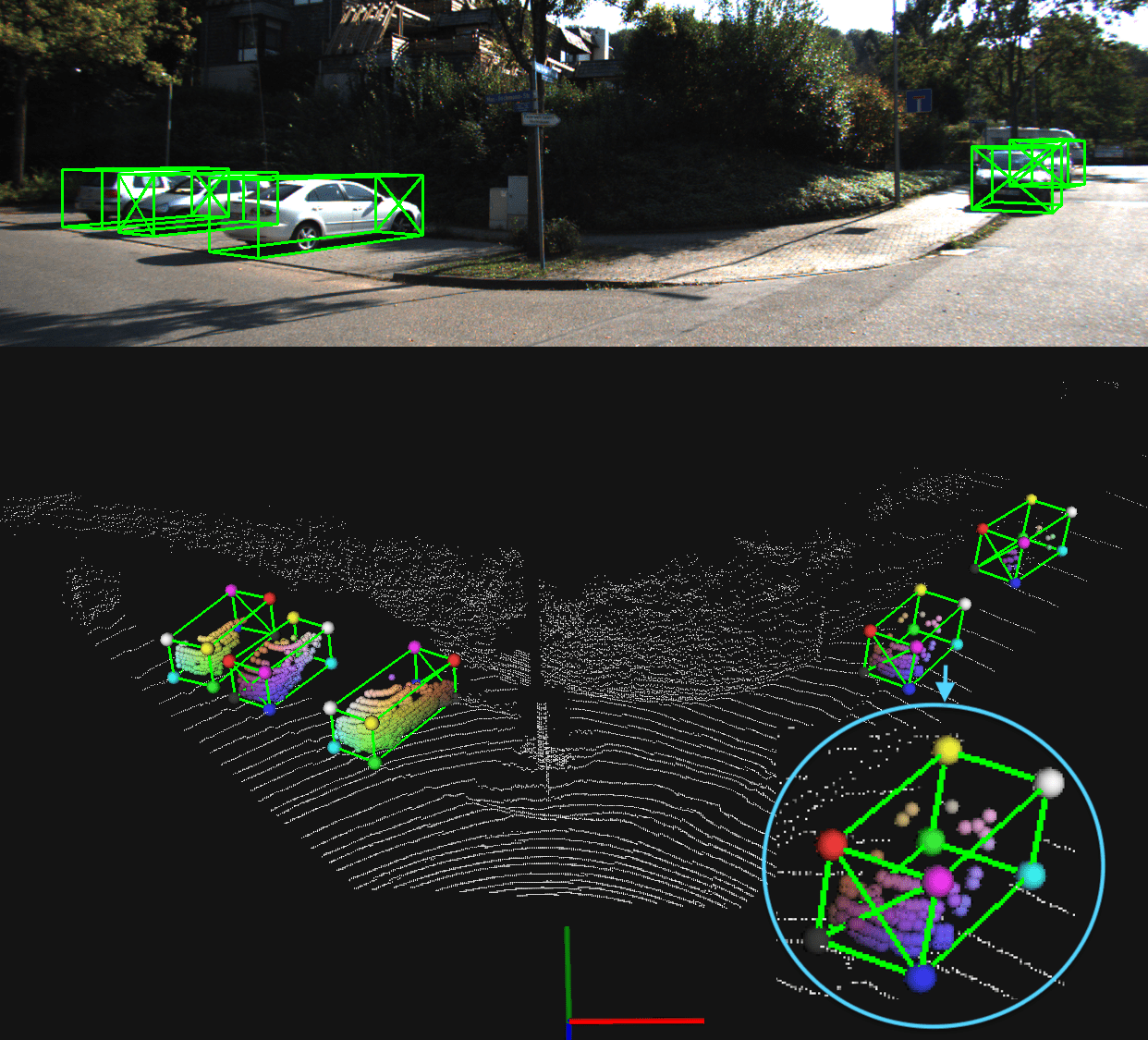}&
			\includegraphics[width=0.25\linewidth]{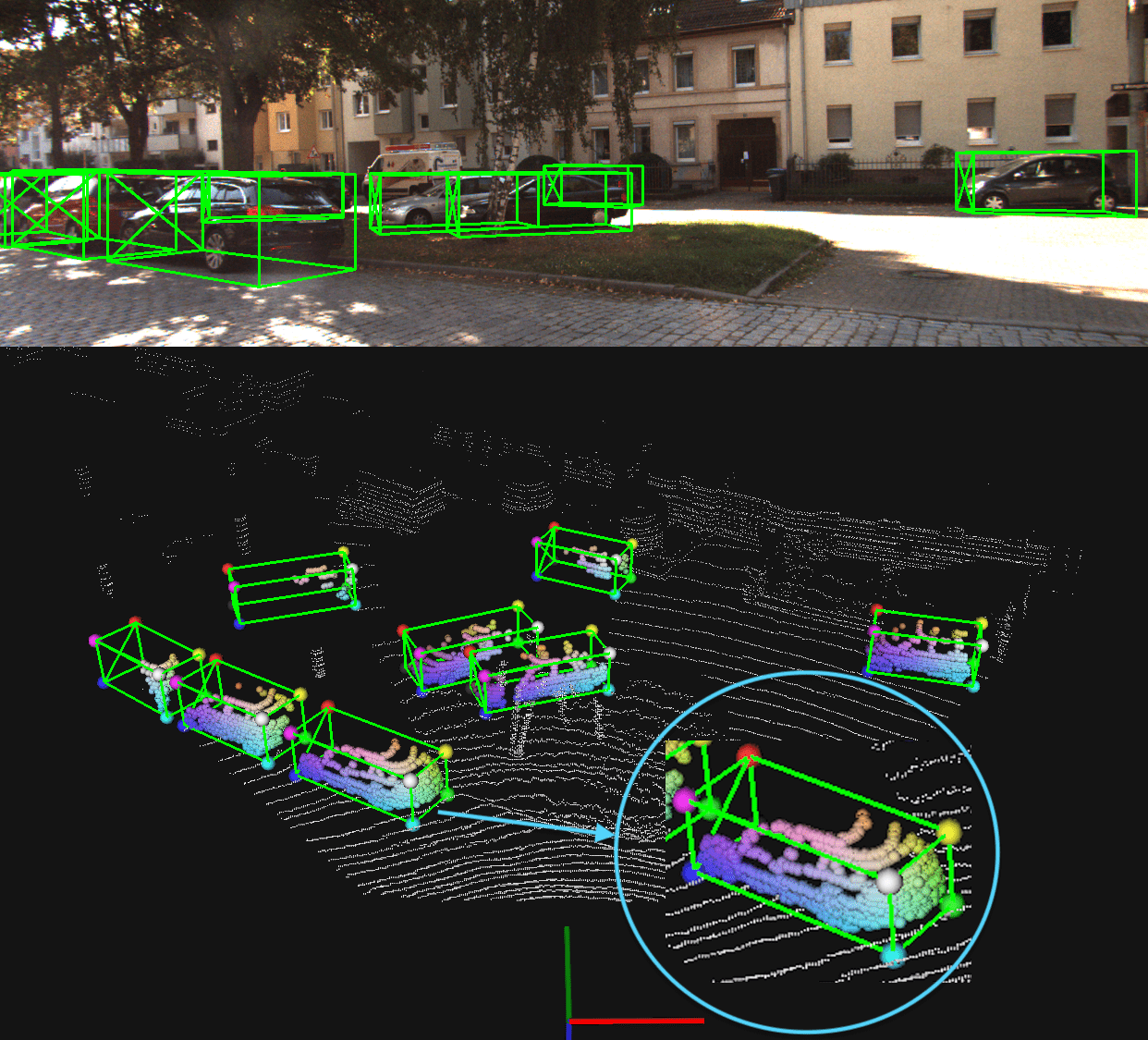}&
			\includegraphics[width=0.25\linewidth]{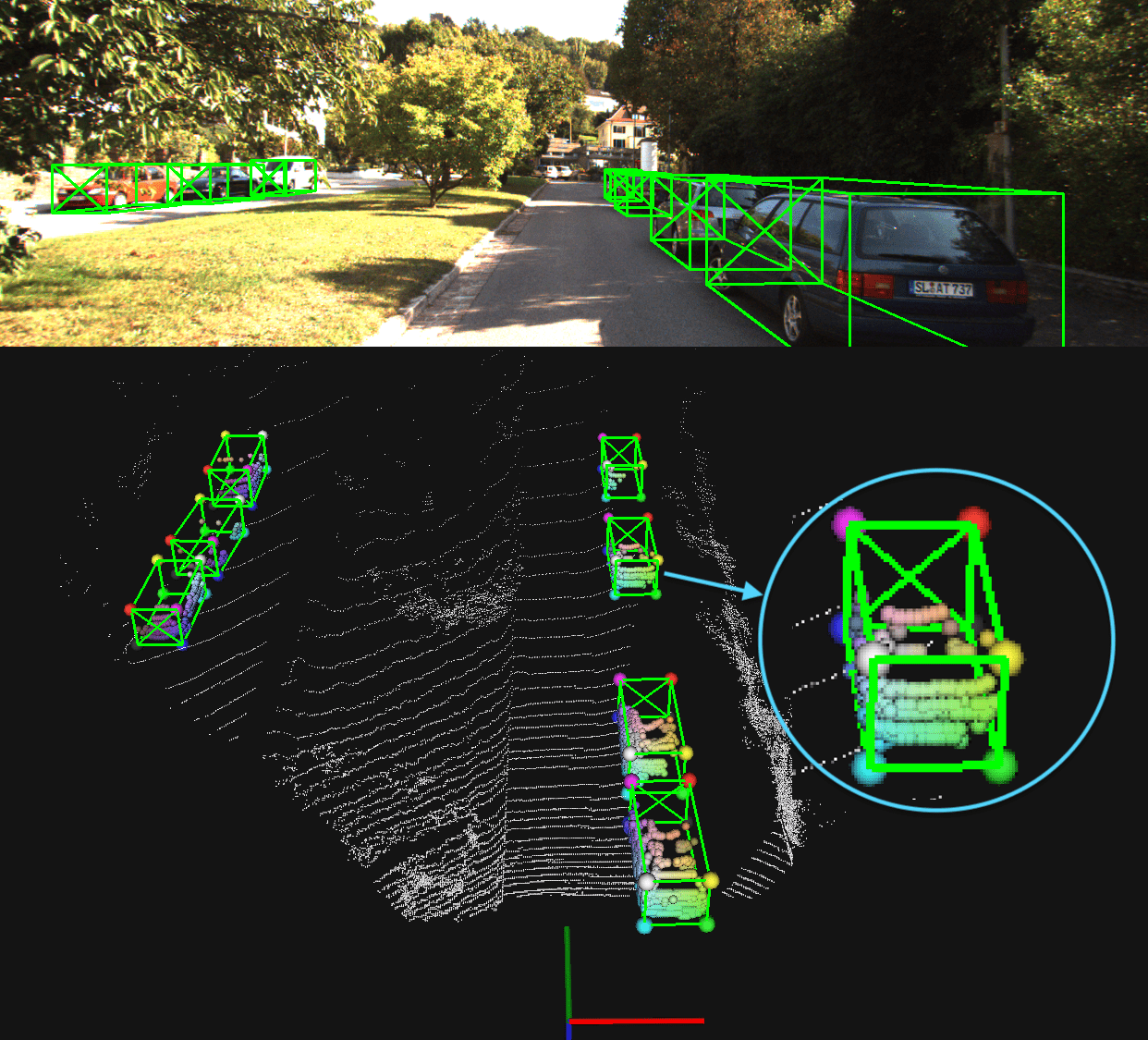}\\
			\includegraphics[width=0.25\linewidth]{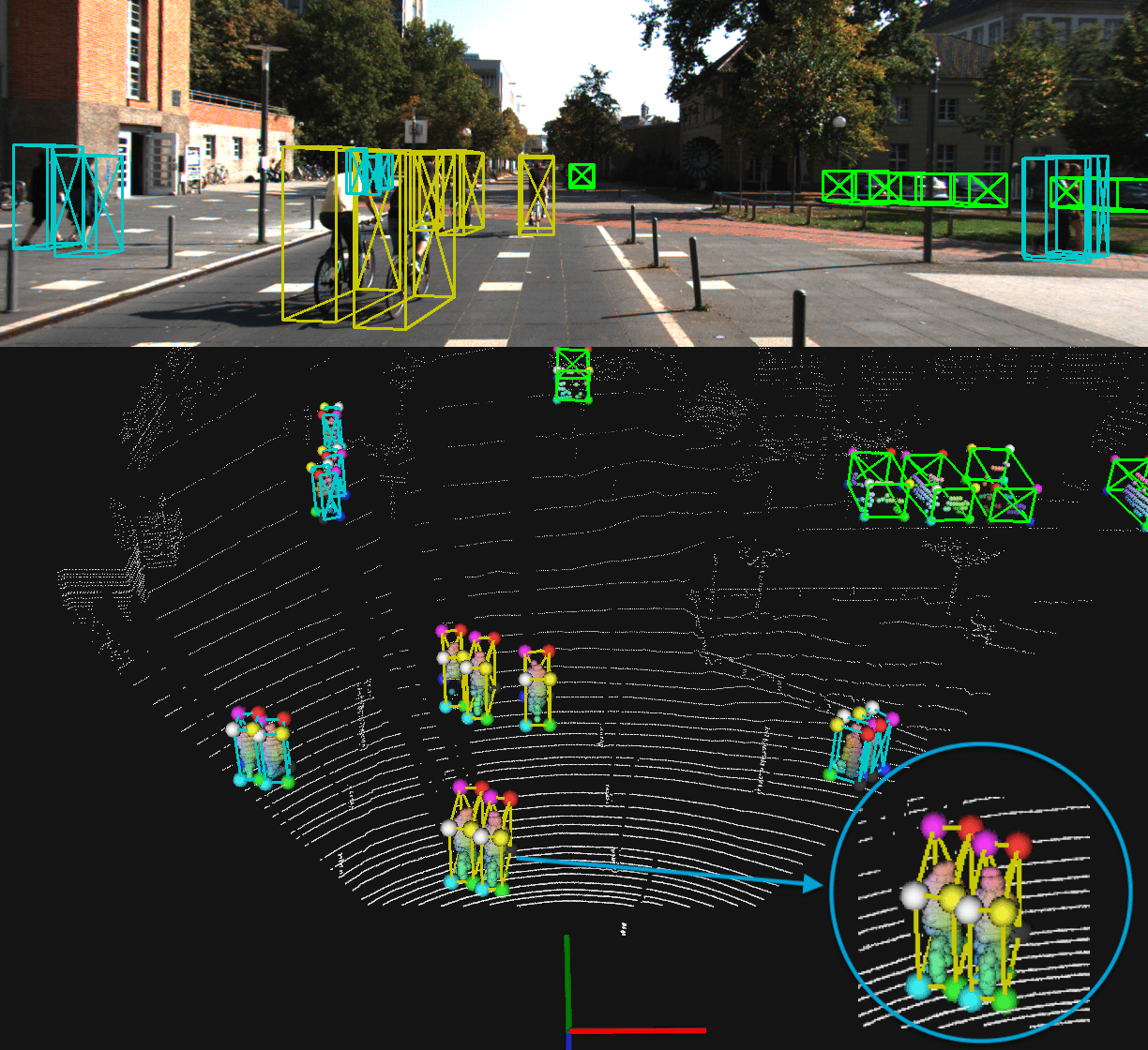}&
			\includegraphics[width=0.25\linewidth]{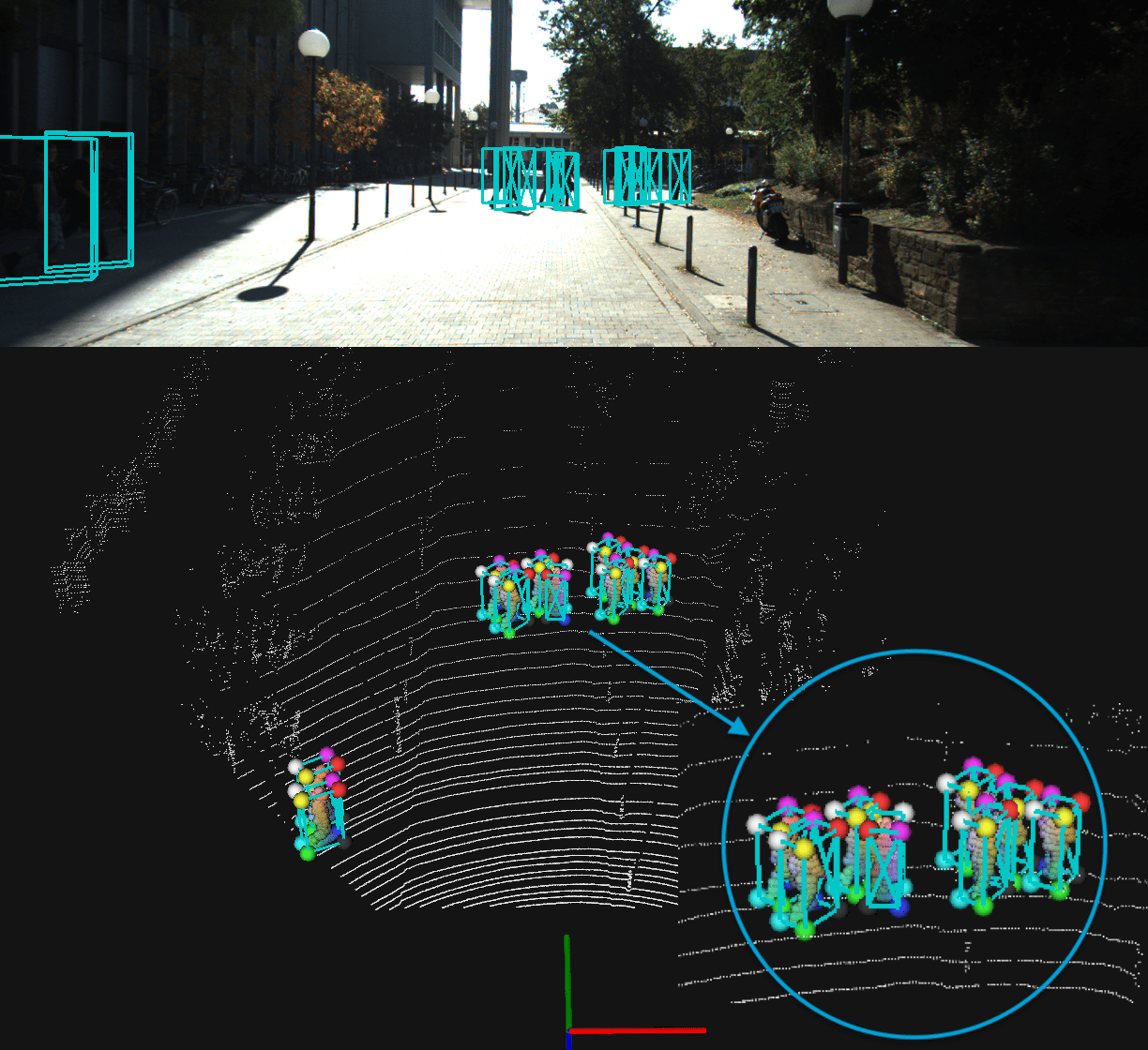}&
			\includegraphics[width=0.25\linewidth]{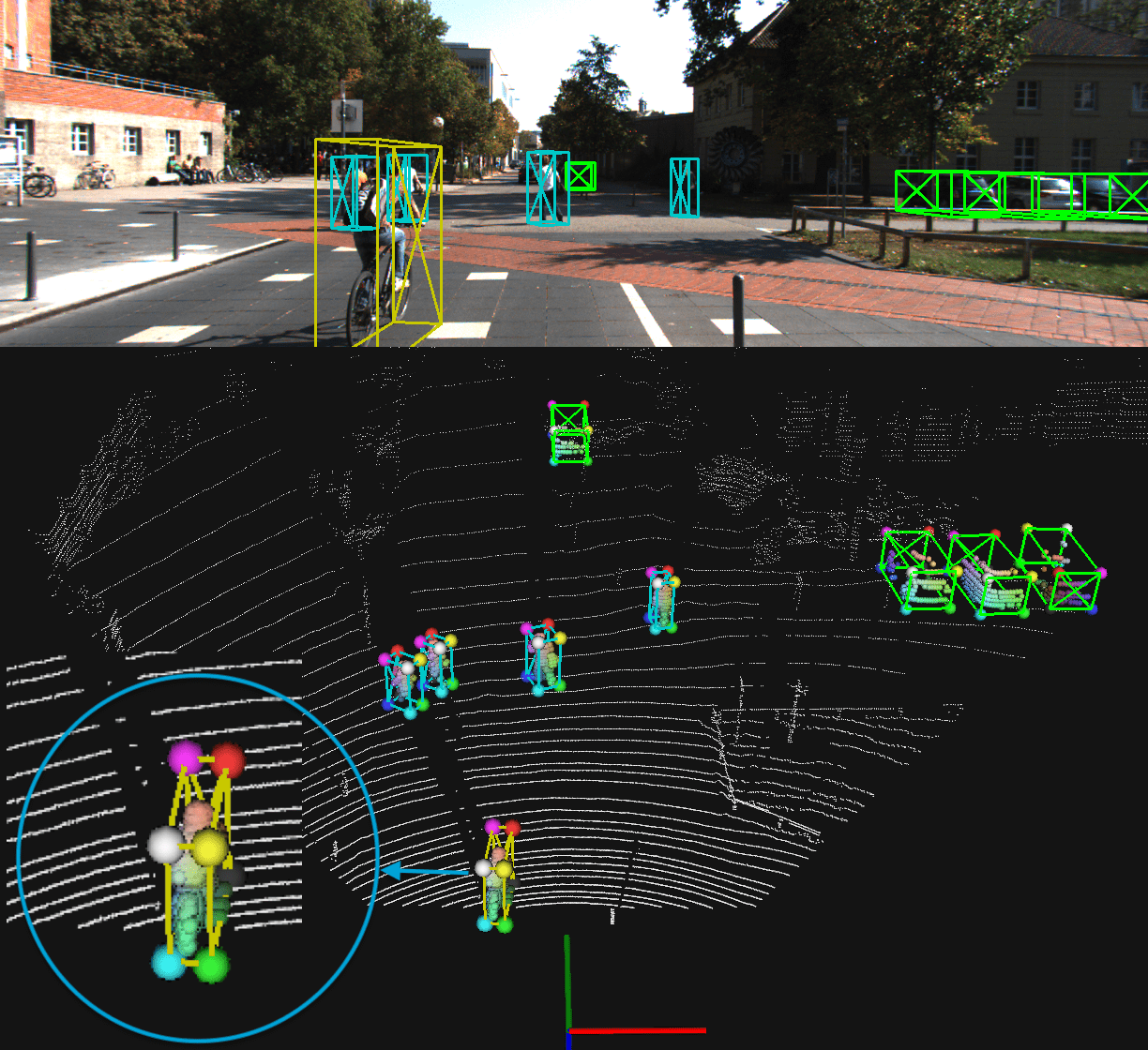}&
			\includegraphics[width=0.25\linewidth]{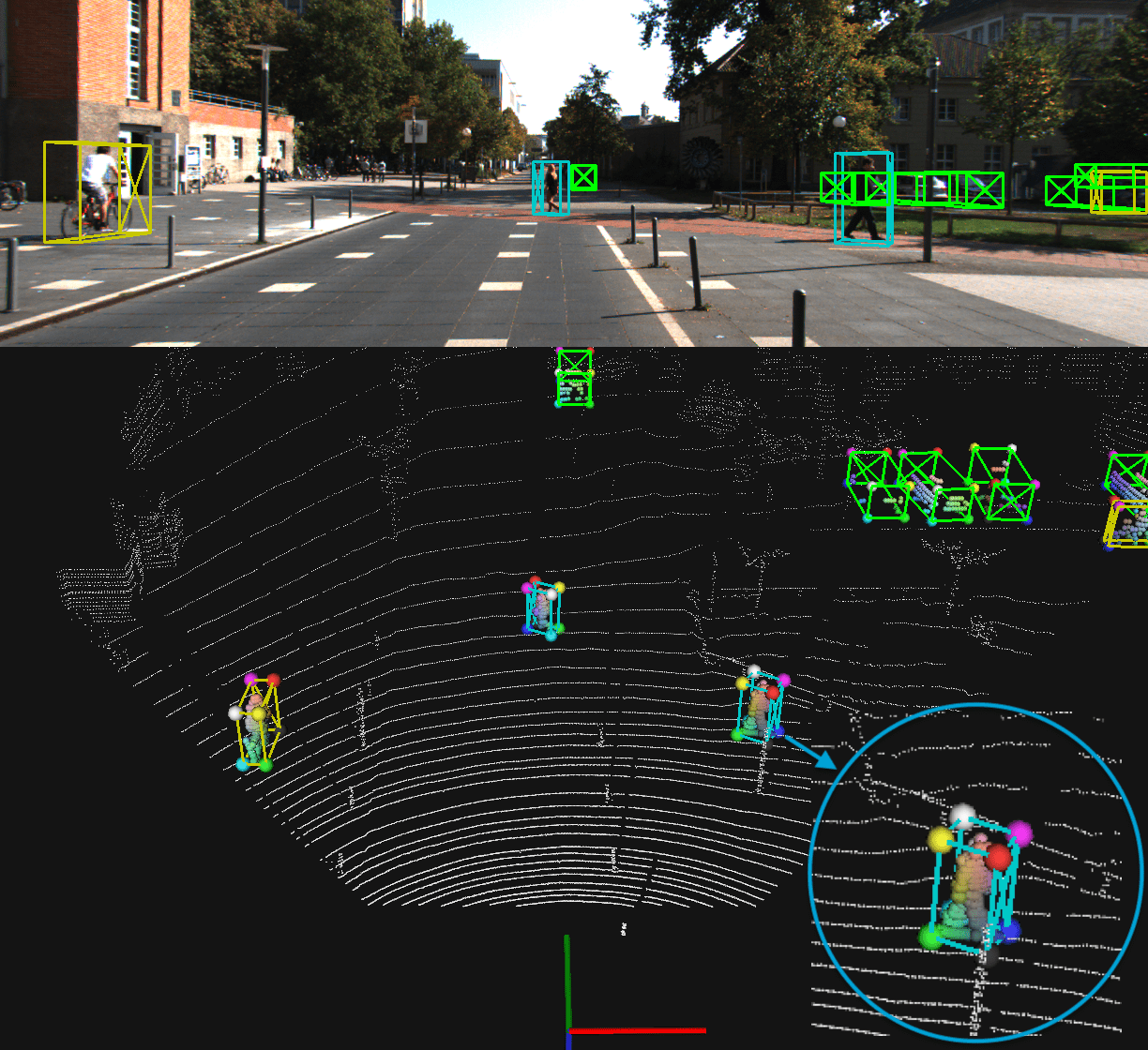}\\
	\end{tabular}}
	\caption{Qualitative results of Part-$A^2$-anchor Net on the KITTI \emph{test} split. The predicted 3D boxes are drawn with green 3D bounding boxes, and the estimated intra-object part locations are visualized with interpolated colors as shown in Fig.~\ref{fig:part_coords}. Best viewed in colors.}
	\label{fig:test_vis}
	\vspace{-0.35cm}
\end{figure*}

\medskip
\noindent
{\bf Results on validation set.}  ~
For the most important car category, 
our methods are compared with state-of-the-art methods on KITTI {\it val} split including both 3D object detection (shown in Table~\ref{tab:val}) and 3D object localization (shown in Table~\ref{tab:val_bev}). We could see that on 
the most important ``moderate'' difficulty level, 
our Part-$A^2$ net outperforms state-of-the-art methods on both two tasks with large margins by using only the point clouds as inputs. 
In addition, our Part-$A^2$ net achieves new state-of-the-art performance on all difficulty levels of the KITTI 3D object detection {\it val} split, which demonstrates the effectiveness of our proposed methods for 3D object detection. 

As shown in Table~\ref{tab:people}, we also report the performance of our methods for cyclist and pedestrian on the validation set for reference. 
Note that compared with PointRCNN, our latest method Part-$A^2$-anchor net improves the performance of cyclist significantly while achieves comparable results on pedestrian. 
The reason for slightly inferior performance on pedestrians might be that the orientation of pedestrian is hard to be recognized from the sparse point cloud, which is harmful for the prediction of part locations in our Part-$A^2$-anchor net. Multi-sensor methods that integrate RGB images would have advantages for detecting small objects like pedestrians.

\medskip
\noindent
{\bf Evaluation of Part-$A^2$-anchor net for predicting intra-object part locations.} ~ 
The intra-object part locations predicted by our part-aware stage-I are crucial for the part-aggregation stage-II to accurately score the box and refine the box location. 
Here we evaluate the accuracy of predicted intra-object part locations by the following metric:
\begin{align}
\text{AbsError}_u = \frac{1}{|| \mathcal{G} ||} \sum_{i \in \mathcal{G}} |\tilde{u}^{(part)}_i - u^{(part)}_i|, u \in \{x, y, z\},
\end{align}
where  $\tilde{u}^{(part)}_i$ is the predicted part location, $u^{(part)}_i$ is the ground truth part location, and $\mathcal{G}$ is the set of foreground points for each sample. The final mAbsError$_u$ is the mean value of AbsError$_u$ for all samples.

As shown in Table~\ref{tab:merror}, for the most important car category, the mean error of our predicted intra-object part locations is 6.28\%, which shows that the part-aware network accurately predicts the intra-object part locations since the average error is only $\pm 6.28$cm per meter for the cars. Based on this accurate intra-object part locations, our part-aggregation stage-II could better score the boxes and refine the box locations by utilizing the predicted geometric information. Here we also report the detailed error statistics of predicted intra-object part locations on different difficulty levels of the KITTI \textit{val} split in Fig.~\ref{fig:partstat} for reference.

\textcolor{black}{
We further analyze the correlations between the errors of the predicted intra-object part locations and the errors of the predicted 3D bounding boxes by calculating the Pearson correlation coefficient, which is $[-1, 1]$ with $1$ denotes fully positive linear correlation and $-1$ is fully negative linear correlation.
Here we utilize $1-\text{IoU}$ to indicate the errors of the predicted 3D bounding boxes, 
where $\text{IoU}$ is the 3D Intersection-over-Union (IoU) between the predicted 3D bounding box and its best matched ground-truth box. 
As shown in Table~\ref{tab:correlation}, we could see that the errors of intra-object part locations have obviously positive correlation with the errors of the predicted 3D bounding boxes. The overall correlation coefficient is 0.531 and the most correlated axis is the $z$-axis in the height direction where the correlation coefficient achieves 0.552, which demonstrates that accurate intra-object part locations are beneficial for predicting more accurate 3D bounding boxes. 
}

\textcolor{black}{
We also report the errors of intra-object part locations on the false positive samples which are caused by inaccurate localization (see row 2 of Table.~\ref{tab:merror}), and we could see that the predicted part location errors increase significantly in all three axes, which indicate that inaccurately predicted intra-object part locations may lead to unsatisfactory 3D object localization and decrease the performance of 3D object detection.
}

\subsection{Qualitative results}\label{sec:exp4}
We present some representative results generated by our proposed Part-$A^2$-anchor net on the \textit{test} split of KITTI dataset in Fig.~\ref{fig:test_vis}. From the figure we could see that our proposed part-aware network could estimate accurate intra-object part locations by using only point cloud as inputs, which are aggregated by our designed part-aggregation network to generate accurate 3D bounding boxes.

\section{Conclusion}
In this paper, we extend our preliminary work PointRCNN to a novel 3D detection framework, the part-aware and aggregation neural network (Part-$A^2$ net), for detecting 3D objects from raw point clouds. Our part-aware stage-I learns to estimate the accurate intra-object part locations by using the free-of-charge intra-object location labels and foreground labels from the ground-truth 3D box annotations. Meanwhile, the 3D proposals are generated by two alternative strategies, anchor-free scheme and anchor-based scheme. 
The predicted intra-object part locations of each object are pooled by the novel RoI-aware point cloud pooling scheme. The following part-aggregation stage-II can better capture the geometric information of object parts to accurately score the boxes and refine their locations. 

Our approach significantly outperforms existing 3D detection methods and achieves new state-of-the-art performance on the challenging KITTI 3D detection benchmark. Extensive experiments were carefully designed and conducted to investigate the individual components of our proposed framework.

% or


%\appendices
%\section{Pending}
%%\section{Proof of the First Zonklar Equation}
%%Appendix one text goes here.
%%
%%% you can choose not to have a title for an appendix
%%% if you want by leaving the argument blank
%%\section{}
%%Appendix two text goes here.
%
%
%% use section* for acknowledgment
%\ifCLASSOPTIONcompsoc
%  % The Computer Society usually uses the plural form
%\else
%  % regular IEEE prefers the singular form
%\fi

\ifCLASSOPTIONcaptionsoff
  \newpage
\fi



%\bibliographystyle{IEEEtran}
%\begin{thebibliography}{1}
%\end{thebibliography}

%\bibliographystyle{ieee}
%\bibliography{egbib}

\bibliographystyle{IEEEtran}
\bibliography{IEEEabrv,egbib}

\begin{thebibliography}{10}
\providecommand{\url}[1]{#1}
\csname url@samestyle\endcsname
\providecommand{\newblock}{\relax}
\providecommand{\bibinfo}[2]{#2}
\providecommand{\BIBentrySTDinterwordspacing}{\spaceskip=0pt\relax}
\providecommand{\BIBentryALTinterwordstretchfactor}{4}
\providecommand{\BIBentryALTinterwordspacing}{\spaceskip=\fontdimen2\font plus
\BIBentryALTinterwordstretchfactor\fontdimen3\font minus
  \fontdimen4\font\relax}
\providecommand{\BIBforeignlanguage}[2]{{%
\expandafter\ifx\csname l@#1\endcsname\relax
\typeout{** WARNING: IEEEtran.bst: No hyphenation pattern has been}%
\typeout{** loaded for the language `#1'. Using the pattern for}%
\typeout{** the default language instead.}%
\else
\language=\csname l@#1\endcsname
\fi
#2}}
\providecommand{\BIBdecl}{\relax}
\BIBdecl

\bibitem{Chen2017CVPR}
X.~Chen, H.~Ma, J.~Wan, B.~Li, and T.~Xia, ``Multi-view 3d object detection
  network for autonomous driving,'' in \emph{CVPR}, 2017.

\bibitem{song2014sliding}
S.~Song and J.~Xiao, ``Sliding shapes for 3d object detection in depth
  images,'' in \emph{European conference on computer vision}.\hskip 1em plus
  0.5em minus 0.4em\relax Springer, 2014, pp. 634--651.

\bibitem{song2016deep}
------, ``Deep sliding shapes for amodal 3d object detection in rgb-d images,''
  in \emph{Proceedings of the IEEE Conference on Computer Vision and Pattern
  Recognition}, 2016, pp. 808--816.

\bibitem{ku2018joint}
J.~Ku, M.~Mozifian, J.~Lee, A.~Harakeh, and S.~Waslander, ``Joint 3d proposal
  generation and object detection from view aggregation,'' \emph{IROS}, 2018.

\bibitem{Liang2018ECCV}
M.~Liang, B.~Yang, S.~Wang, and R.~Urtasun, ``Deep continuous fusion for
  multi-sensor 3d object detection,'' in \emph{ECCV}, 2018.

\bibitem{qi2017frustum}
C.~R. Qi, W.~Liu, C.~Wu, H.~Su, and L.~J. Guibas, ``Frustum pointnets for 3d
  object detection from rgb-d data,'' \emph{arXiv preprint arXiv:1711.08488},
  2017.

\bibitem{yan2018second}
Y.~Yan, Y.~Mao, and B.~Li, ``Second: Sparsely embedded convolutional
  detection,'' \emph{Sensors}, vol.~18, no.~10, p. 3337, 2018.

\bibitem{8461232}
X.~Du, M.~H. Ang, S.~Karaman, and D.~Rus, ``A general pipeline for 3d detection
  of vehicles,'' in \emph{2018 IEEE International Conference on Robotics and
  Automation (ICRA)}, May 2018, pp. 3194--3200.

\bibitem{lang2018pointpillars}
A.~H. Lang, S.~Vora, H.~Caesar, L.~Zhou, J.~Yang, and O.~Beijbom,
  ``Pointpillars: Fast encoders for object detection from point clouds,''
  \emph{CVPR}, 2019.

\bibitem{Yang2018CoRL}
B.~Yang, M.~Liang, and R.~Urtasun, ``Hdnet: Exploiting hd maps for 3d object
  detection,'' in \emph{2nd Conference on Robot Learning (CoRL)}, 2018.

\bibitem{yang2018pixor}
B.~Yang, W.~Luo, and R.~Urtasun, ``Pixor: Real-time 3d object detection from
  point clouds,'' in \emph{Proceedings of the IEEE Conference on Computer
  Vision and Pattern Recognition}, 2018, pp. 7652--7660.

\bibitem{luo2018fast}
W.~Luo, B.~Yang, and R.~Urtasun, ``Fast and furious: Real time end-to-end 3d
  detection, tracking and motion forecasting with a single convolutional net,''
  in \emph{Proceedings of the IEEE Conference on Computer Vision and Pattern
  Recognition}, 2018, pp. 3569--3577.

\bibitem{simony2018complex}
M.~Simony, S.~Milzy, K.~Amendey, and H.-M. Gross, ``Complex-yolo: an
  euler-region-proposal for real-time 3d object detection on point clouds,'' in
  \emph{Proceedings of the European Conference on Computer Vision (ECCV)},
  2018, pp. 0--0.

\bibitem{girshick2015fast}
R.~Girshick, ``Fast r-cnn,'' in \emph{Proceedings of the IEEE international
  conference on computer vision}, 2015, pp. 1440--1448.

\bibitem{ren2015faster}
S.~Ren, K.~He, R.~Girshick, and J.~Sun, ``Faster r-cnn: Towards real-time
  object detection with region proposal networks,'' in \emph{Advances in neural
  information processing systems}, 2015, pp. 91--99.

\bibitem{liu2016ssd}
W.~Liu, D.~Anguelov, D.~Erhan, C.~Szegedy, S.~Reed, C.-Y. Fu, and A.~C. Berg,
  ``Ssd: Single shot multibox detector,'' in \emph{European conference on
  computer vision}.\hskip 1em plus 0.5em minus 0.4em\relax Springer, 2016, pp.
  21--37.

\bibitem{redmon2016you}
J.~Redmon, S.~Divvala, R.~Girshick, and A.~Farhadi, ``You only look once:
  Unified, real-time object detection,'' in \emph{Proceedings of the IEEE
  conference on computer vision and pattern recognition}, 2016, pp. 779--788.

\bibitem{redmon2017yolo9000}
J.~Redmon and A.~Farhadi, ``Yolo9000: better, faster, stronger,'' in
  \emph{Proceedings of the IEEE conference on computer vision and pattern
  recognition}, 2017, pp. 7263--7271.

\bibitem{lin2017feature}
T.-Y. Lin, P.~Doll{\'a}r, R.~Girshick, K.~He, B.~Hariharan, and S.~Belongie,
  ``Feature pyramid networks for object detection,'' in \emph{Proceedings of
  the IEEE Conference on Computer Vision and Pattern Recognition}, 2017, pp.
  2117--2125.

\bibitem{dai2017deformable}
J.~Dai, H.~Qi, Y.~Xiong, Y.~Li, G.~Zhang, H.~Hu, and Y.~Wei, ``Deformable
  convolutional networks,'' in \emph{Proceedings of the IEEE international
  conference on computer vision}, 2017, pp. 764--773.

\bibitem{lin2018focal}
T.-Y. Lin, P.~Goyal, R.~Girshick, K.~He, and P.~Doll{\'a}r, ``Focal loss for
  dense object detection,'' \emph{IEEE transactions on pattern analysis and
  machine intelligence}, 2018.

\bibitem{law2018cornernet}
H.~Law and J.~Deng, ``Cornernet: Detecting objects as paired keypoints,'' in
  \emph{Proceedings of the European Conference on Computer Vision (ECCV)},
  2018, pp. 734--750.

\bibitem{he2017mask}
K.~He, G.~Gkioxari, P.~Doll{\'a}r, and R.~Girshick, ``Mask r-cnn,'' in
  \emph{Proceedings of the IEEE international conference on computer vision},
  2017, pp. 2961--2969.

\bibitem{cai2018cascade}
Z.~Cai and N.~Vasconcelos, ``Cascade r-cnn: Delving into high quality object
  detection,'' in \emph{Proceedings of the IEEE Conference on Computer Vision
  and Pattern Recognition}, 2018, pp. 6154--6162.

\bibitem{xu2018pointfusion}
D.~Xu, D.~Anguelov, and A.~Jain, ``Pointfusion: Deep sensor fusion for 3d
  bounding box estimation,'' in \emph{Proceedings of the IEEE Conference on
  Computer Vision and Pattern Recognition}, 2018, pp. 244--253.

\bibitem{wang2019frustum}
Z.~Wang and K.~Jia, ``Frustum convnet: Sliding frustums to aggregate local
  point-wise features for amodal 3d object detection,'' in \emph{IROS}.\hskip
  1em plus 0.5em minus 0.4em\relax IEEE, 2019.

\bibitem{qi2017pointnet}
C.~R. Qi, H.~Su, K.~Mo, and L.~J. Guibas, ``Pointnet: Deep learning on point
  sets for 3d classification and segmentation,'' in \emph{Proceedings of the
  IEEE Conference on Computer Vision and Pattern Recognition}, 2017, pp.
  652--660.

\bibitem{qi2017pointnet++}
C.~R. Qi, L.~Yi, H.~Su, and L.~J. Guibas, ``Pointnet++: Deep hierarchical
  feature learning on point sets in a metric space,'' in \emph{Advances in
  Neural Information Processing Systems}, 2017, pp. 5099--5108.

\bibitem{zhou2018voxelnet}
Y.~Zhou and O.~Tuzel, ``Voxelnet: End-to-end learning for point cloud based 3d
  object detection,'' in \emph{Proceedings of the IEEE Conference on Computer
  Vision and Pattern Recognition}, 2018, pp. 4490--4499.

\bibitem{SubmanifoldSparseConvNet}
B.~Graham and L.~van~der Maaten, ``Submanifold sparse convolutional networks,''
  \emph{arXiv preprint arXiv:1706.01307}, 2017.

\bibitem{3DSemanticSegmentationWithSubmanifoldSparseConvNet}
B.~Graham, M.~Engelcke, and L.~van~der Maaten, ``3d semantic segmentation with
  submanifold sparse convolutional networks,'' \emph{CVPR}, 2018.

\bibitem{shi2019pointrcnn}
S.~Shi, X.~Wang, and H.~Li, ``Pointrcnn: 3d object proposal generation and
  detection from point cloud,'' in \emph{Proceedings of the IEEE Conference on
  Computer Vision and Pattern Recognition}, 2019, pp. 770--779.

\bibitem{Geiger2012CVPR}
A.~Geiger, P.~Lenz, and R.~Urtasun, ``Are we ready for autonomous driving? the
  kitti vision benchmark suite,'' in \emph{Conference on Computer Vision and
  Pattern Recognition (CVPR)}, 2012.

\bibitem{mousavian20173d}
A.~Mousavian, D.~Anguelov, J.~Flynn, and J.~Ko{\v{s}}eck{\'a}, ``3d bounding
  box estimation using deep learning and geometry,'' in \emph{Computer Vision
  and Pattern Recognition (CVPR), 2017 IEEE Conference on}.\hskip 1em plus
  0.5em minus 0.4em\relax IEEE, 2017, pp. 5632--5640.

\bibitem{li2019gs3d}
B.~Li, W.~Ouyang, L.~Sheng, X.~Zeng, and X.~Wang, ``Gs3d: An efficient 3d
  object detection framework for autonomous driving,'' in \emph{Proceedings of
  the IEEE Conference on Computer Vision and Pattern Recognition}, 2019, pp.
  1019--1028.

\bibitem{chabot2017deep}
F.~Chabot, M.~Chaouch, J.~Rabarisoa, C.~Teuli{\`e}re, and T.~Chateau, ``Deep
  manta: A coarse-to-fine many-task network for joint 2d and 3d vehicle
  analysis from monocular image,'' in \emph{Proc. IEEE Conf. Comput. Vis.
  Pattern Recognit.(CVPR)}, 2017, pp. 2040--2049.

\bibitem{zhu2014single}
M.~Zhu, K.~G. Derpanis, Y.~Yang, S.~Brahmbhatt, M.~Zhang, C.~Phillips,
  M.~Lecce, and K.~Daniilidis, ``Single image 3d object detection and pose
  estimation for grasping,'' in \emph{Robotics and Automation (ICRA), 2014 IEEE
  International Conference on}.\hskip 1em plus 0.5em minus 0.4em\relax IEEE,
  2014, pp. 3936--3943.

\bibitem{mottaghi2015coarse}
R.~Mottaghi, Y.~Xiang, and S.~Savarese, ``A coarse-to-fine model for 3d pose
  estimation and sub-category recognition,'' in \emph{Proceedings of the IEEE
  Conference on Computer Vision and Pattern Recognition}, 2015, pp. 418--426.

\bibitem{manhardt2018roi10d}
F.~Manhardt, W.~Kehl, and A.~Gaidon, ``Roi-10d: Monocular lifting of 2d
  detection to 6d pose and metric shape,'' in \emph{Computer Vision and Pattern
  Recognition (CVPR)}.\hskip 1em plus 0.5em minus 0.4em\relax IEEE, 2019.

\bibitem{chen2016monocular}
X.~Chen, K.~Kundu, Z.~Zhang, H.~Ma, S.~Fidler, and R.~Urtasun, ``Monocular 3d
  object detection for autonomous driving,'' in \emph{Proceedings of the IEEE
  Conference on Computer Vision and Pattern Recognition}, 2016, pp. 2147--2156.

\bibitem{chen20153d}
X.~Chen, K.~Kundu, Y.~Zhu, A.~G. Berneshawi, H.~Ma, S.~Fidler, and R.~Urtasun,
  ``3d object proposals for accurate object class detection,'' in
  \emph{Advances in Neural Information Processing Systems}, 2015, pp. 424--432.

\bibitem{ku2019monopsr}
J.~Ku*, A.~D. Pon*, and S.~L. Waslander, ``Monocular 3d object detection
  leveraging accurate proposals and shape reconstruction,'' in \emph{CVPR},
  2019.

\bibitem{licvpr2019}
P.~Li, X.~Chen, and S.~Shen, ``Stereo r-cnn based 3d object detection for
  autonomous driving,'' in \emph{CVPR}, 2019.

\bibitem{wangcvpr2019}
Y.~Wang, W.-L. Chao, D.~Garg, B.~Hariharan, M.~Campbell, and K.~Weinberger,
  ``Pseudo-lidar from visual depth estimation: Bridging the gap in 3d object
  detection for autonomous driving,'' in \emph{CVPR}, 2019.

\bibitem{ronneberger2015u}
O.~Ronneberger, P.~Fischer, and T.~Brox, ``U-net: Convolutional networks for
  biomedical image segmentation,'' in \emph{International Conference on Medical
  image computing and computer-assisted intervention}.\hskip 1em plus 0.5em
  minus 0.4em\relax Springer, 2015, pp. 234--241.

\bibitem{yi2019gspn}
L.~Yi, W.~Zhao, H.~Wang, M.~Sung, and L.~J. Guibas, ``Gspn: Generative shape
  proposal network for 3d instance segmentation in point cloud,'' in
  \emph{Proceedings of the IEEE Conference on Computer Vision and Pattern
  Recognition}, 2019, pp. 3947--3956.

\bibitem{hou20193d}
J.~Hou, A.~Dai, and M.~Nie{\ss}ner, ``3d-sis: 3d semantic instance segmentation
  of rgb-d scans,'' in \emph{Proceedings of the IEEE Conference on Computer
  Vision and Pattern Recognition}, 2019, pp. 4421--4430.

\bibitem{wang2018sgpn}
W.~Wang, R.~Yu, Q.~Huang, and U.~Neumann, ``Sgpn: Similarity group proposal
  network for 3d point cloud instance segmentation,'' in \emph{Proceedings of
  the IEEE Conference on Computer Vision and Pattern Recognition}, 2018, pp.
  2569--2578.

\bibitem{wang2019associatively}
X.~Wang, S.~Liu, X.~Shen, C.~Shen, and J.~Jia, ``Associatively segmenting
  instances and semantics in point clouds,'' in \emph{Proceedings of the IEEE
  Conference on Computer Vision and Pattern Recognition}, 2019, pp. 4096--4105.

\bibitem{lahoud20193d}
J.~Lahoud, B.~Ghanem, M.~Pollefeys, and M.~R. Oswald, ``3d instance
  segmentation via multi-task metric learning,'' \emph{arXiv preprint
  arXiv:1906.08650}, 2019.

\bibitem{bert2017}
\BIBentryALTinterwordspacing
B.~D. Brabandere, D.~Neven, and L.~V. Gool, ``Semantic instance segmentation
  with a discriminative loss function,'' \emph{CoRR}, vol. abs/1708.02551,
  2017. [Online]. Available: \url{http://arxiv.org/abs/1708.02551}
\BIBentrySTDinterwordspacing

\bibitem{bai2017deep}
M.~Bai and R.~Urtasun, ``Deep watershed transform for instance segmentation,''
  in \emph{Proceedings of the IEEE Conference on Computer Vision and Pattern
  Recognition}, 2017, pp. 5221--5229.

\bibitem{felzenszwalb2009object}
P.~F. Felzenszwalb, R.~B. Girshick, D.~McAllester, and D.~Ramanan, ``Object
  detection with discriminatively trained part-based models,'' \emph{IEEE
  transactions on pattern analysis and machine intelligence}, vol.~32, no.~9,
  pp. 1627--1645, 2009.

\bibitem{fidler20123d}
S.~Fidler, S.~Dickinson, and R.~Urtasun, ``3d object detection and viewpoint
  estimation with a deformable 3d cuboid model,'' in \emph{Advances in neural
  information processing systems}, 2012, pp. 611--619.

\bibitem{pepik20123d}
B.~Pepik, P.~Gehler, M.~Stark, and B.~Schiele, ``3d 2 pm--3d deformable part
  models,'' in \emph{European Conference on Computer Vision}.\hskip 1em plus
  0.5em minus 0.4em\relax Springer, 2012, pp. 356--370.

\bibitem{lim2014fpm}
J.~J. Lim, A.~Khosla, and A.~Torralba, ``Fpm: Fine pose parts-based model with
  3d cad models,'' in \emph{European conference on computer vision}.\hskip 1em
  plus 0.5em minus 0.4em\relax Springer, 2014, pp. 478--493.

\bibitem{huang2018recurrent}
Q.~Huang, W.~Wang, and U.~Neumann, ``Recurrent slice networks for 3d
  segmentation of point clouds,'' in \emph{Proceedings of the IEEE Conference
  on Computer Vision and Pattern Recognition}, 2018, pp. 2626--2635.

\bibitem{li2018pointcnn}
Y.~Li, R.~Bu, M.~Sun, W.~Wu, X.~Di, and B.~Chen, ``Pointcnn: Convolution on
  x-transformed points,'' in \emph{Advances in Neural Information Processing
  Systems}, 2018, pp. 820--830.

\bibitem{wang2018deep}
S.~Wang, S.~Suo, W.-C. Ma, A.~Pokrovsky, and R.~Urtasun, ``Deep parametric
  continuous convolutional neural networks,'' in \emph{Proceedings of the IEEE
  Conference on Computer Vision and Pattern Recognition}, 2018, pp. 2589--2597.

\bibitem{zhao2019pointweb}
H.~Zhao, L.~Jiang, C.-W. Fu, and J.~Jia, ``Pointweb: Enhancing local
  neighborhood features for point cloud processing,'' in \emph{Proceedings of
  the IEEE Conference on Computer Vision and Pattern Recognition}, 2019, pp.
  5565--5573.

\bibitem{wu2019pointconv}
W.~Wu, Z.~Qi, and L.~Fuxin, ``Pointconv: Deep convolutional networks on 3d
  point clouds,'' in \emph{Proceedings of the IEEE Conference on Computer
  Vision and Pattern Recognition}, 2019, pp. 9621--9630.

\bibitem{jiang2018acquisition}
B.~Jiang, R.~Luo, J.~Mao, T.~Xiao, and Y.~Jiang, ``Acquisition of localization
  confidence for accurate object detection,'' in \emph{Proceedings of the
  European Conference on Computer Vision (ECCV)}, 2018, pp. 784--799.

\bibitem{sun2018fishnet}
S.~Sun, J.~Pang, J.~Shi, S.~Yi, and W.~Ouyang, ``Fishnet: A versatile backbone
  for image, region, and pixel level prediction,'' in \emph{Advances in Neural
  Information Processing Systems}, 2018, pp. 762--772.

\bibitem{Liang2019CVPR}
M.~Liang*, B.~Yang*, Y.~Chen, R.~Hu, and R.~Urtasun, ``Multi-task multi-sensor
  fusion for 3d object detection,'' in \emph{CVPR}, 2019.

\bibitem{kitti_leaderboard}
{KITTI 3D object detection benchmark leader board},
  \url{http://www.cvlibs.net/datasets/kitti/eval_object.php?obj_benchmark=3d},
  Accessed on 2019-08-15.

\end{thebibliography}

\begin{IEEEbiography}[{\includegraphics[width=1in,height=1.25in]{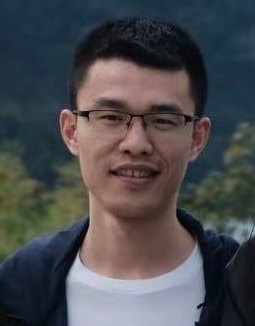}}]{Shaoshuai Shi}
received his B.S. degree in Computer Science and Technology from Harbin Institute of Technology, China in 2017. He is currently a Ph.D. student in the Department of Electronic Engineering at The Chinese University of Hong Kong. His research interests include computer vision, deep learning and 3D scene understanding.
\end{IEEEbiography}

\begin{IEEEbiography}[{\includegraphics[width=1in,height=1.25in]{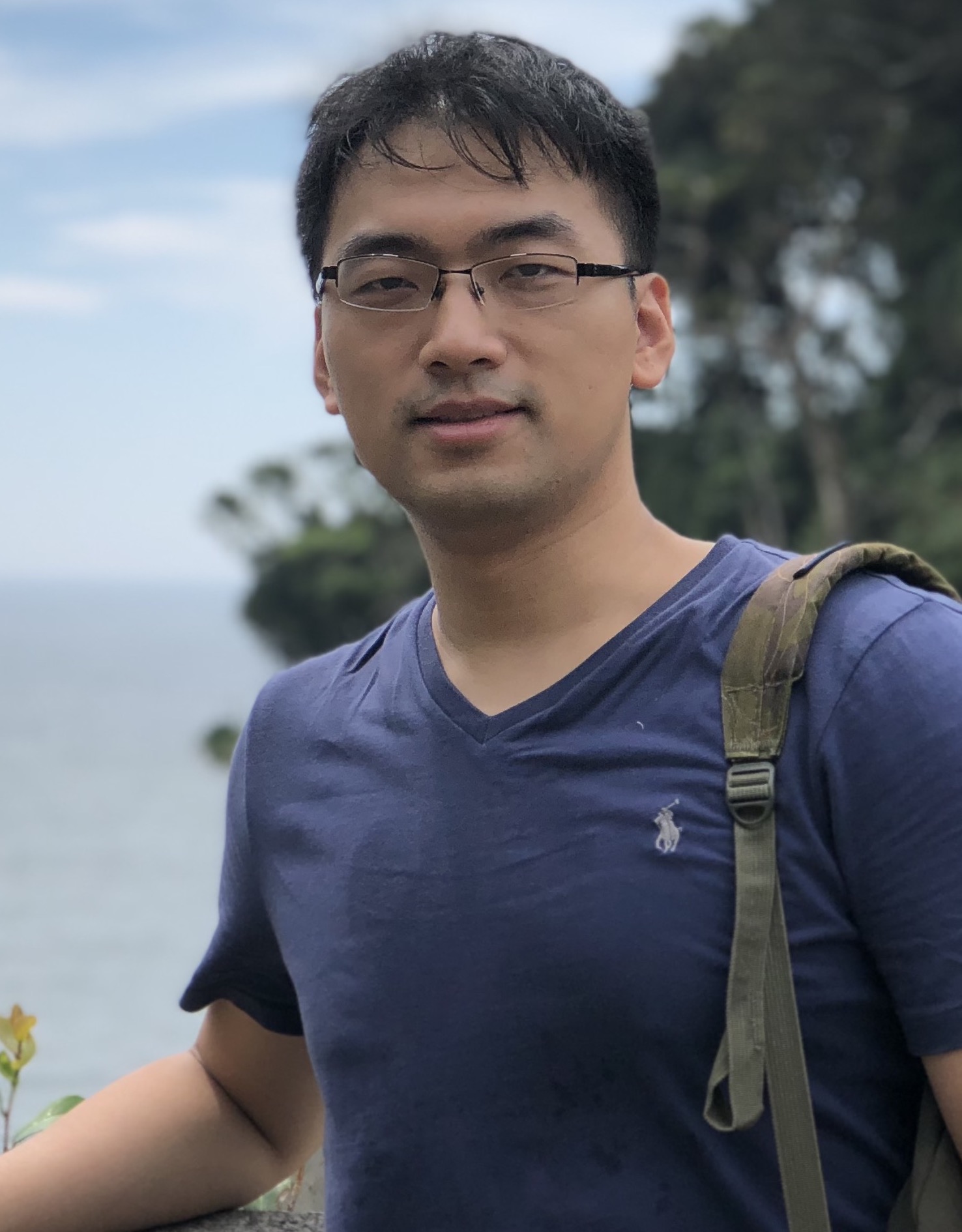}}]{Zhe Wang}
received his B.S. degree in Optical Engineering of Zhejiang University in 2012, and the Ph.D. degree in the Department of Electronic Engineering at The Chinese University of Hong Kong. He is current a Research Vice Director of SenseTime. His research interests include computer vision and deep learning.
\end{IEEEbiography}

\begin{IEEEbiography}[{\includegraphics[width=1in,height=1.25in]{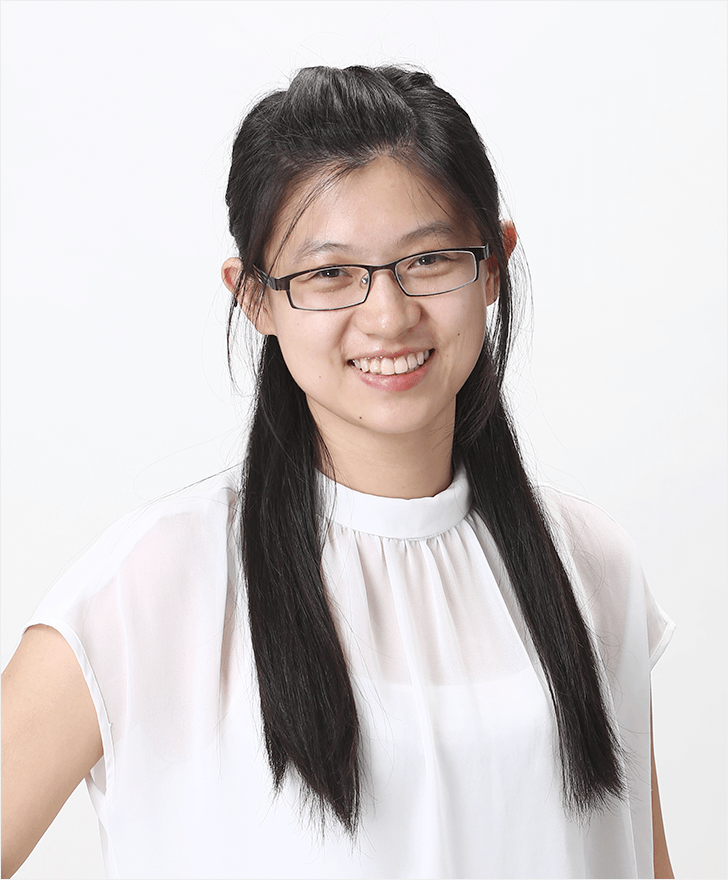}}]{Jianping Shi}
received her B.S. degree in Computer Science and Engineering from Zhejiang University, China in 2011, and the Ph.D. degree from the Computer Science and Engineering Department at The Chinese University of Hong Kong in 2015. She is currently the Executive Research Director of SenseTime. Her research interests include computer vision and deep learning.
\end{IEEEbiography}

\begin{IEEEbiography}[{\includegraphics[width=1in,height=1.25in]{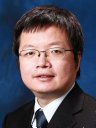}}]{Xiaogang Wang}
received the B.S. degree from the University of Science and Technology of China in 2001, the MS degree from The Chinese University of Hong Kong in 2003, and the PhD degree from the Computer Science and Artificial Intelligence Laboratory, Massachusetts Institute of Technology in 2009. He is currently an associate professor in the Department of Electronic Engineering at The Chinese University of Hong Kong. His research interests include computer vision and machine learning. 
%He is a member of the IEEE.	
\end{IEEEbiography}

\begin{IEEEbiography}[{\includegraphics[width=1in,height=1.25in]{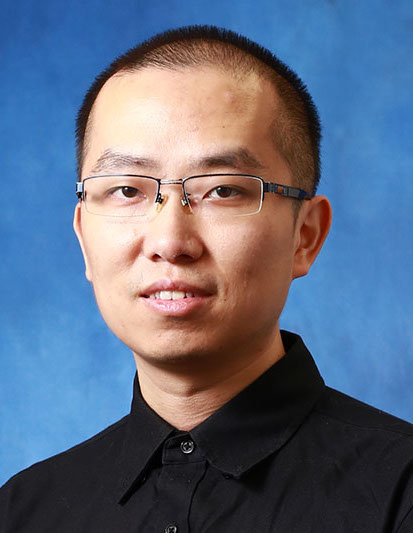}}]{Hongsheng Li}
received the bachelor’s degree in automation from the East China University of Science and Technology, and the master’s and doctorate degrees in computer science from Lehigh University, Pennsylvania, in 2006, 2010, and 2012, respectively. He is currently an assistant professor in the Department of Electronic Engineering at The Chinese University of Hong Kong. His research interests include computer vision, medical image analysis, and machine learning.
\end{IEEEbiography}

\end{document}